%% file: arXiv.tex
\documentclass[10pt,twocolumn,letterpaper]{article}

\PassOptionsToPackage{dvipsnames,svgnames,table}{xcolor}

\usepackage[pagenumbers]{iccv} % To force page numbers, e.g. for an arXiv version

\usepackage{graphicx}
\usepackage{amsmath}
\usepackage{amssymb}
\usepackage{booktabs}

\usepackage{graphbox}
\usepackage{enumitem}

\usepackage{makecell}
\usepackage{tabularx}
\usepackage{multirow}
\usepackage{multicol}
\usepackage{diagbox}  % Import diagbox for diagonal cell division
\usepackage{booktabs} % For better table styling

\usepackage[T1]{fontenc}    % use 8-bit T1 fonts
\usepackage{amsfonts}       % blackboard math symbols
\usepackage{pifont}
\usepackage{dsfont}

\usepackage{algorithm}
\usepackage{algorithmic}

\usepackage[utf8]{inputenc} % allow utf-8 input
\usepackage{url}            % simple URL typesetting
\usepackage{mathtools}
\usepackage{nicefrac}       % compact symbols for 1/2, etc.
\usepackage{tikz}
\usepackage{placeins}
\usepackage[accsupp]{axessibility}  % Improves PDF readability for those with disabilities.

\usepackage[normalem]{ulem}

\include{macro}

\definecolor{iccvblue}{rgb}{0.21,0.49,0.74}
\usepackage[pagebackref,breaklinks,colorlinks,allcolors=iccvblue]{hyperref}

\def\paperID{2803} % *** Enter the Paper ID here
\def\confName{ICCV}
\def\confYear{2803}

\title{Exploiting Diffusion Prior for Task-driven Image Restoration}

\author{$\text{Jaeha Kim}^{1}$\quad $\text{Junghun Oh}^{1}$\quad $\text{Kyoung Mu Lee}^{1,2}$\\
$^{1}$Dept. of ECE\&ASRI, $^{2}$IPAI, Seoul National University, Korea\\
{\tt\small jhkim97s2@gmail.com, \{dh6dh, kyoungmu\}@snu.ac.kr}
}

\begin{document}
\maketitle
\input{Sections/submission_arXiv/1_abstract}
\input{Sections/submission_arXiv/2_introduction}
\input{Sections/submission_arXiv/3_related_works}
\input{Sections/submission_arXiv/4_proposed_method}
\input{Sections/submission_arXiv/5_experiments}
\input{Sections/submission_arXiv/6_conclusion}
{
    \small
    \bibliographystyle{ieeenat_fullname}
    \bibliography{main}
}

\clearpage % Starts a new page

\twocolumn[
    \begin{center}
        \vspace{0.5cm}
        \Large \emph{\textbf{Supplementary Material for}} \\
        \Large \textbf{Exploiting Diffusion Prior for Task-driven Image Restoration}\\
        \vspace{1.0cm}
        \fontsize{12}{14}\selectfont
        %%%%%%%%% AUTHORS - PLEASE UPDATE
        \author{$\text{Jaeha Kim}^{1}$\quad $\text{Junghun Oh}^{1}$\quad $\text{Kyoung Mu Lee}^{1,2}$\\
        $^{1}$Dept. of ECE\&ASRI, $^{2}$IPAI, Seoul National University, Korea\\
        {\tt\small jhkim97s2@gmail.com, \{dh6dh, kyoungmu\}@snu.ac.kr}
        }
        \vspace{1.0cm}
    \end{center}
]

%%%%%%%%%%%%%%%%%%%%%% Supple Section/Figure/Table Number - with S %%%%%%%%%%%%%%%%%%%%%
% Supple 할때는 S로, Rebuttal 할 때에는 R로 바꿔서 사용
\appendix
\setcounter{section}{0}
\setcounter{figure}{0}
\setcounter{table}{0}
\setcounter{equation}{0}
\setcounter{algorithm}{0}

\renewcommand{\thetable}{S\arabic{table}}
\renewcommand{\thesection}{S\arabic{section}}
\renewcommand{\thefigure}{S\arabic{figure}}
\renewcommand{\theequation}{S\arabic{equation}}
\renewcommand{\thealgorithm}{S\arabic{algorithm}}

%%%%%%%%% BODY
In this supplementary document, we show the additional details, results, and ablation studies omitted from the main manuscript due to the lack of space:
% 
\iffalse
\begin{compactitem}[$\bullet$]
    \item \ref{sec:degradation}. Degradation details
    % 
    \item \ref{sec:training-details}. Training details
    % 
    \item \ref{sec:impact-pixel-error}. Impact of pre-restoration network
    % 
    \item \ref{sec:training-algorithm}. Training algorithm
    % 
    \item \ref{sec:complementary-HLF}. Benefit of using two feature spaces in HLF
    % 
    \item \ref{sec:computational-cost}. Computational cost of the EDTR
    % 
    \item \ref{sec:output-stochasticity}. Output stochasticity of the EDTR
    % 
    \item \ref{sec:comparison-detection}. Comparison with DiffBIR for detection
    % 
    \item \ref{sec:sr4ir-combined-with-sd}. SR4IR combined with SD
    % 
    \item \ref{sec:details-previous}. Details for the previous works
    % 
    \item \ref{sec:additional-ablation}. Additional ablation studies
    % 
    \item \ref{sec:further-visualization}. Further visualization results
    % 
\end{compactitem}
\fi
% 
\def\myvspace{0mm}
\begin{itemize}
    \item \ref{sec:degradation}. Degradation details
    \vspace{\myvspace}
    \item \ref{sec:training-details}. Training details
    \vspace{\myvspace}
    \item \ref{sec:impact-pixel-error}. Impact of pre-restoration network
    \vspace{\myvspace}
    \item \ref{sec:training-algorithm}. Training algorithm
    \vspace{\myvspace}
    \item \ref{sec:complementary-HLF}. Benefit of using two feature spaces in HLF
    \vspace{\myvspace}
    \item \ref{sec:computational-cost}. Computational cost of the EDTR
    \vspace{\myvspace}
    \item \ref{sec:output-stochasticity}. Output stochasticity of the EDTR
    \vspace{\myvspace}
    \item \ref{sec:comparison-detection}. Comparison with DiffBIR for detection
    \vspace{\myvspace}
    \item \ref{sec:sr4ir-combined-with-sd}. SR4IR combined with SD
    \vspace{\myvspace}
    \item \ref{sec:details-previous}. Details for the previous works
    \vspace{\myvspace}
    \item \ref{sec:additional-ablation}. Additional ablation studies
    \vspace{\myvspace}
    \item \ref{sec:further-visualization}. Further visualization results
    \vspace{\myvspace}
\end{itemize}

\input{Sections/supple_arXiv/Degradation_details}
\input{Sections/supple_arXiv/Training_details}
\input{Sections/supple_arXiv/Impact_pre-restoration}
\input{Sections/supple_arXiv/Training_algorithm}
\input{Sections/supple_arXiv/Complementary_HLF}
\input{Sections/supple_arXiv/Computational_cost}
\input{Sections/supple_arXiv/Stochasticity}
\input{Sections/supple_arXiv/Comparison_detection}
\input{Sections/supple_arXiv/SR4IR-combined-with-SD}
\input{Sections/supple_arXiv/Details_for_previous}
\input{Sections/supple_arXiv/Additional_ablations}

\input{Sections/supple_arXiv/Further_visualization}

\end{document}

%% file: macro.tex
\long\def\ignorethis#1{}

\newcommand{\normtwo}[1]{\lVert #1 \rVert_2^2}
\newcommand{\normone}[1]{\left\lVert #1 \right\rVert_1}

\newcommand{\Paragraph}[1]{\vspace{1mm}\noindent\textbf{#1}}

\newcommand{\ignore}[1]{}   % ignore this

\newcommand{\figspace}{\vspace{-2mm}}
\newcommand{\figxspace}{\vspace{-3mm}}

\definecolor{BurntOrange}{rgb}{0.8,0.33,0.0}
\definecolor{DarkBlue}{rgb}{0.0,0.0,0.75}

\newcolumntype{L}[1]{>{\raggedright\let\newline\\\arraybackslash\hspace{0pt}}m{#1}}
\newcolumntype{C}[1]{>{\centering\let\newline\\\arraybackslash\hspace{0pt}}m{#1}}
\newcolumntype{R}[1]{>{\raggedleft\let\newline\\\arraybackslash\hspace{0pt}}m{#1}}

\definecolor{green4mark}{RGB}{21, 152, 56}
\definecolor{red4mark}{RGB}{252, 54, 65}
\newcommand{\cmark}{\textcolor{green4mark}{\ding{51}}}
\newcommand{\xmark}{\textcolor{red4mark}{\ding{55}}}

\def\cyanP{20}

%% file: Sections/submission_arXiv/1_abstract.tex
\begin{abstract}
Task-driven image restoration (TDIR) has recently emerged to address performance drops in high-level vision tasks caused by low-quality (LQ) inputs.
% 
% The goal of TDIR is to improve both visual quality and task performance.
% 
Previous TDIR methods struggle to handle practical scenarios in which images are degraded by multiple complex factors, leaving minimal clues for restoration.
This motivates us to leverage the diffusion prior, one of the most powerful natural image priors.
However, while the diffusion prior can help generate visually plausible results, using it to restore task-relevant details remains challenging, even when combined with recent TDIR methods.
To address this, we propose EDTR, which effectively harnesses the power of diffusion prior to restore task-relevant details.
Specifically, we propose directly leveraging useful clues from LQ images in the diffusion process by generating from pixel-error-based pre-restored LQ images with mild noise added.
Moreover, we employ a small number of denoising steps to prevent the generation of redundant details that dilute crucial task-related information.
We demonstrate that our method effectively utilizes diffusion prior for TDIR, significantly enhancing task performance and visual quality across diverse tasks with multiple complex degradations. 
\end{abstract}

%% file: Sections/submission_arXiv/2_introduction.tex
\vspace{-1mm}
\section{Introduction}
\label{sec:introduction}
\vspace{-1mm}
Image degradation often occurs in the real world due to various factors, such as transmission loss, limited camera performance, or poor shooting conditions.
In such cases, key high-frequency details can be lost, resulting in significant performance drops in high-level vision tasks, such as image classification, semantic segmentation, and object detection.

Image restoration (IR) can be a promising solution to improve the performance of high-level vision tasks. It involves recovering missing details in low-quality (LQ) images using a learned natural image prior.
However, recent studies~\cite{kim2024beyond, wu2024unsupervised, pei2018does, dai2016image} show that the straightforward application of IR methods as a pre-processing step is not very effective in restoring essential information for high-level vision tasks.
This has led to the emergence of Task-driven Image Restoration (TDIR), which aims to improve visual quality in ways that benefit high-level vision tasks.

\begin{figure}[t]
    \centering
    \begin{minipage}{0.255 \linewidth}
        \subfloat[Low-quality]{\includegraphics[width=\linewidth]{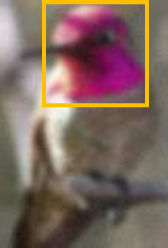}}
        %\vspace{1mm}
    \end{minipage}
    \hfill
    \begin{minipage}{0.7 \linewidth}
        \subfloat[\label{fig:main_figure_b}DiffBIR~\cite{lin2023diffbir}]{\includegraphics[width=0.48\linewidth]{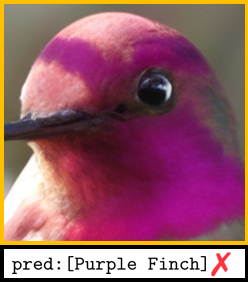}}
        \hfill
        \subfloat[\label{fig:main_figure_c}SR4IR~\cite{kim2024beyond}+SD~\cite{rombach2022high}]{\includegraphics[width=0.48\linewidth]{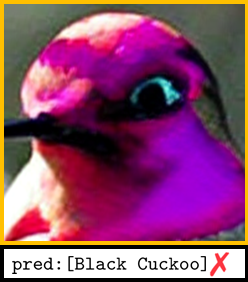}}
        \\
        % \vspace{1mm}
        \\
        \subfloat[\label{fig:main_figure_d}\textbf{EDTR (Ours)}]{\includegraphics[width=0.48\linewidth]{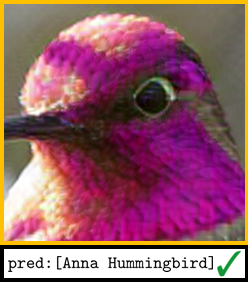}}
        \hfill
        \subfloat[High-quality]{\includegraphics[width=0.48\linewidth]{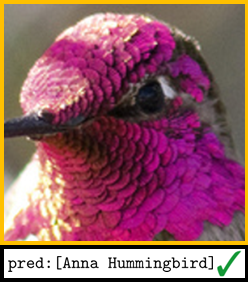}}
    \end{minipage}
    \vspace{-1mm}
    \caption{
        \textbf{Effect of diffusion prior on image classification performance.}
        % 
        %, such as incorrect eye shapes or missing plumage.
        While incorporating the StableDiffusion~\cite{rombach2022high} produces visually plausible results as (b), it lacks task-relevant details (\eg, eye shape), even when combined with the previous TDIR method as (c), leading to incorrect classification.
        In contrast, EDTR successfully restores details as (d), resulting in correct classification.
    }
    \label{fig:main_figure}
    \figxspace
    % \vspace{-2mm}
\end{figure}

Previous TDIR methods address each type of degradation separately, \eg, super-resolution~\cite{kim2024beyond}, denoising~\cite{liu2017image}.
This restricts their practical applicability, as real-world images often exhibit multiple degradations simultaneously.
To address this, we tackle a more realistic scenario by jointly handling multiple complex degradations.
However, we observe that previous TDIR methods achieve limited performance gains in such scenarios, as they struggle to utilize the few available clues in degraded images to restore task-relevant details.
This highlights the need for a strong natural image prior for the TDIR problem.

In this paper, we leverage the prior knowledge from the pre-trained StableDiffusion (SD)~\cite{rombach2022high}, which provides one of the most powerful natural image priors.
However, we observe that while the conventional SD-based IR method~\cite{lin2023diffbir} can generate visually pleasing results, its straightforward adoption fails to restore task-relevant details, leading to incorrect predictions, as shown in Figure~\ref{fig:main_figure_b}.
More importantly, we observe that SD still struggles to restore task-relevant details even when combined with a recent state-of-the-art TDIR method~\cite{kim2024beyond}, as shown in Figure~\ref{fig:main_figure_c}.
In other words, \textit{while diffusion prior is powerful, it must be properly adapted}, as its standard form fails to improve TDIR.
This issue arises because restoring task-relevant details is inherently challenging in the standard diffusion process used in conventional SD-based IR methods~\cite{lin2023diffbir, wang2024exploiting, yang2023pasd, yu2024scaling, wu2024seesr}.
We identify two key reasons:
(1) Although the LQ image is used as a condition for SD, generating from pure noise hinders the direct utilization of useful clues in LQ images, making it difficult to restore task-relevant details.
(2) Long-step denoising, around 50, introduces redundant details that dilute crucial task-related information for TDIR.

To address this, we propose EDTR, \textbf{E}xploiting \textbf{D}iffusion prior for \textbf{T}ask-driven Image \textbf{R}estoration, the TDIR method adopting the diffusion prior in ways that directly leverage its power to restore task-relevant details in complex degradation scenarios.
The primary component of EDTR is three-fold:
(1) We directly incoporate LQ images into the diffusion process.
Since SD was not originally trained to handle complex degradations, directly feeding LQ images into the diffusion process is ineffective.
Therefore, we first pre-restore the LQ images using a pixel-error-based IR method, then generate from these images with mild noise added, rather than generating from pure noise, a process referred to as pre-restoration and partial diffusion.
(2) We apply a short-step denoising (\eg, 1, 4) during inference.
We claim that increasing the number of steps (\eg, 50) reduces task performance by introducing redundant details that dilute crucial details for TDIR.
(3) We introduce tailored training loss terms to guide diffusion prior to restore task-relevant details.
These loss terms are essential for achieving visually appealing results, even in a one-step denoising.

We validate our method in three high-level vision tasks: image classification, semantic segmentation, and object detection.
These tasks involve complex degradation scenarios, including a mix of downsampling, blur, noise, and JPEG artifacts.
Experimental results demonstrate that EDTR effectively harnesses the power of diffusion prior, significantly improving task performance (\eg, about +50\% improvement in detection) while producing visually pleasing results, as shown in Figure~\ref{fig:main_figure_d}, and generalizes well to real-world images.
We summarize our contributions as:
% 
% \vspace{1mm}
\begin{itemize}
    \item We propose EDTR, the TDIR method that directly leverages a powerful pre-trained diffusion prior.
    \vspace{1mm}
    % 
    % \item We propose tailored designs for exploiting the diffusion prior in TDIR, including pre-restoration, partial diffusion, short-step denoising, and specialized training loss terms.
    \item We propose tailored designs for exploiting the diffusion prior to effectively restore task-relevant details.
    \vspace{1mm}
    \item We significantly improve task performance while producing visually pleasing results under complex degradation.
\end{itemize}

%% file: Sections/submission_arXiv/3_related_works.tex
\vspace{-1mm}
\section{Related Work}
\vspace{-3mm}
\Paragraph{General image restoration.}
Since the pioneering SRCNN~\cite{sr_srcnn}, numerous deep neural network-based methods~\cite{sr_vdsr, sr_srgan, sr_edsr, sr_dbpn, sr_esrgan, sr_rdn, sr_rcan, sr_san, sr_han, db_nafnet} have emerged in the field of image restoration~(IR).
Vision transformers~\cite{misc_vit}-based IR methods~\cite{db_restormer, rs_ipt, db_uformer, chen2023hat}, such as SwinIR~\cite{sr_swinir}, have also demonstrated strong performance.
Recently, IR studies~\cite{wang2024exploiting, lin2023diffbir, yang2023pasd, yu2024scaling, liu2024diff} have focused on leveraging the large pre-trained diffusion models, primarily StableDiffusion~(SD)~\cite{rombach2022high}, which provides a powerful natural image prior essential for addressing the ill-posedness of the IR problem.
StableSR~\cite{wang2024exploiting} and DiffBIR~\cite{lin2023diffbir} introduce trainable modules alongside a frozen SD backbone, similar to ControlNet~\cite{zhang2023adding}.
More recent works~\cite{wu2024seesr, yu2024scaling, sun2024coser, mei2025power} have focused on fully harnessing SD's generative capabilities by designing carefully crafted input conditions.

\begin{figure*}
    \centering
    % \captionsetup[subfigure]{labelfont=scriptsize, textfont=scriptsize}
    \subfloat[The architecture of the proposed EDTR with high-level vision task network.\label{fig:edtr-overview}]{\includegraphics[width=0.98\linewidth]{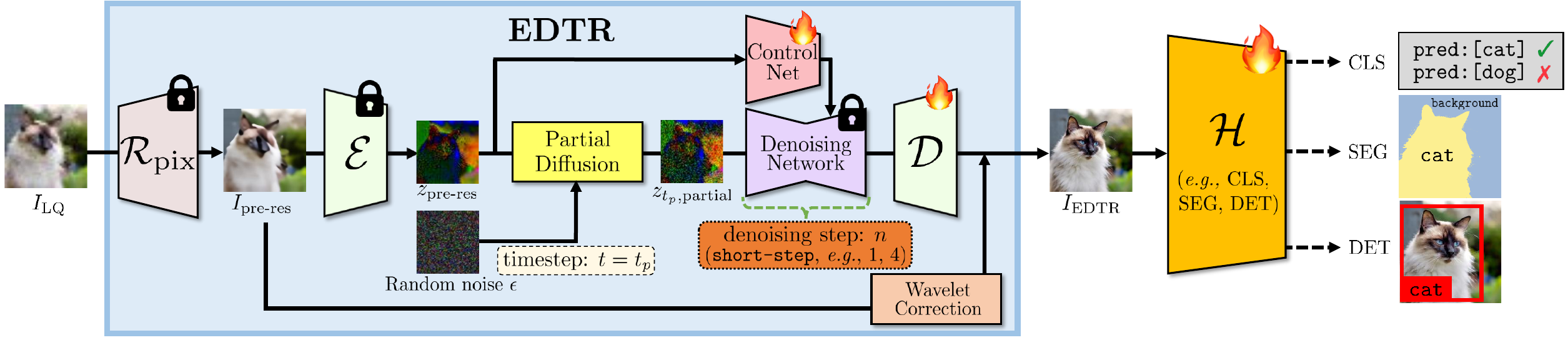}}
    \\
    \vspace{1mm}
    \subfloat[The joint training procedure of EDTR and high-level vision task network.\label{fig:edtr-training}]{\includegraphics[width=0.92\linewidth]{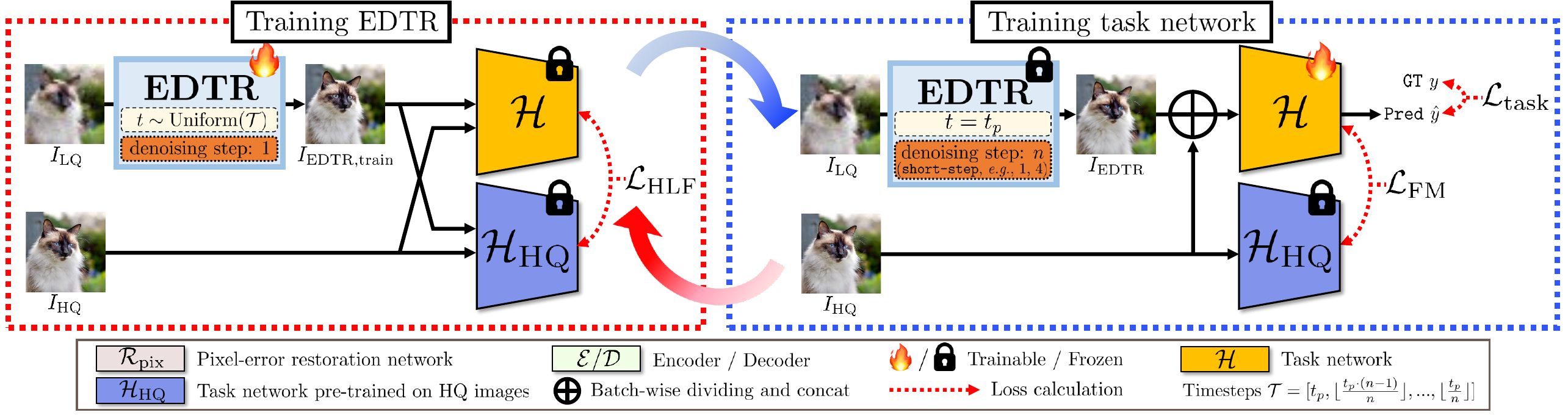}}
    \vspace{-3mm}
    \caption{\textbf{The overall framework of the proposed method.}
    $I_{\text{EDTR},\text{train}}$ is the EDTR restored image when EDTR is in training mode.
    }
    \label{fig:edtr-architecture}
    \figxspace
    \vspace{-2mm}
\end{figure*}

\Paragraph{Task-driven image restoration.}
Taking a different perspective, recent studies~\cite{shermeyer2019effects, sun2022rethinking, bai2018finding, bai2018sod, liu2022image, wu2024unsupervised} have focused on the emerging field of Task-driven Image Restoration~(TDIR), which aims to improve high-level vision task performance when handling LQ images.
AOD-Net~\cite{li2017aod} suggests that using dehazing as a preprocessing step and applying object detection can enhance detection performance in foggy conditions.
However, Pei~\etal~\cite{pei2018does} show that applying IR methods as a simple preprocessing step does not significantly improve detection performance, with similar findings reported in the super-resolution field~\cite{dai2016image}.

To address this, recent TDIR methods~\cite{sr_tdsr, liu2017image, zhao2018residual, liu2020connecting, huang2020dsnet, li2023detection, son2020urie} have started training the IR network in combination with the task networks.
% 
% URIE~\cite{son2020urie} proposes an effective network architecture for TDIR and trains the IR network using classification loss from a pre-trained classification network.
% 
TDSR~\cite{sr_tdsr} trains the IR network using both pixel-error loss and detection loss from a pre-trained object detection network, with a dedicated training schedule.
RSRSSN~\cite{zhao2018residual} proposes an end-to-end trainable IR and detection networks with Feature Maps Multibox loss.
SR4IR~\cite{kim2024beyond} introduces a joint training framework for TDIR with a task-driven perceptual loss to guide the IR network in learning task-relevant features.
More recently, UniRestore~\cite{chen2025unirestore} introduces the diffusion prior into TDIR; however, its diffusion-based training is independent of task-driven training, performed in two stages, limiting the ability to fully leverage the diffusion prior for TDIR.
We instead propose integrating the diffusion prior directly within task-driven training to better restore task-relevant details.
% 
%Most recently, UniRestore~\cite{chen2025unirestore} introduces the diffusion prior into TDIR; however, the diffusion-based restoration and task-driven training are performed in two separate stages, limiting the ability to fully harness diffusion prior for restoring task-relevant details.
% Most recently, UniRestore~\cite{chen2025unirestore} introduces the diffusion prior into TDIR; however, it fine-tunes diffusion models independently of task-driven training, limiting the ability to fully harness diffusion prior for restoring task-relevant details.
% 
%Instead, we propose directly leveraging the power of SD with tailored design choices for TDIR.
% However, while SD has been proven to possess a powerful natural image prior, leveraging SD for TDIR remains unexplored and is the focus of our paper.
% 
% BAD-Net~\cite{li2023detection} proposes an attention fusion module to utilize both dehazed and hazy features, with an interval iterative training strategy for detection-friendly IR.
% 
% On the other hand, VaT~\cite{wu2024unsupervised} proposes training only a lightweight network to bridge existing IR and HT networks, optimized through cycle-consistency training.

%% file: Sections/submission_arXiv/4_proposed_method.tex
\vspace{-2mm}
\section{Proposed Method}
\label{sec:proposed_method}

\vspace{-2mm}
\paragraph{Problem definition.}
Let $I_\text{HQ}$ and $I_\text{LQ}$ denote the high-quality~(HQ) and low-quality~(LQ) images, respectively, with paired labels $y$ from corresponding high-level vision task, \eg, class label in the image classification.
The goal of Task-driven Image Restoration~(TDIR) is to restore LQ images in a way that enhances the subsequent task performance while also generating visually appealing results.
Formally, given a trainable image restoration~(IR) model~$\mathcal{R}$ and a trainable task network $\mathcal{H}$, our aim is to maximize $\mathbf{M}_\text{H}(\mathcal{H}(\mathcal{R}(I_\text{LQ})), y)$, where $\mathbf{M}_\text{H}$ is the evaluation metric for the task, \eg, the accuracy for classification, while also achieving visually pleasing restored results $\mathcal{R}(I_\text{LQ})$.

\vspace{-1mm}
\subsection{Preliminary: Conventional SD-based IR}
\label{ssec:preliminary}
Unlike pure generation, IR requires generating images that are faithful to the given LQ images.
To achieve this, previous StableDiffusion~(SD)-based IR methods~\cite{wang2024exploiting, lin2023diffbir, yang2023pasd, yu2024scaling, sun2024coser} commonly use LQ images as a \textit{condition} for SD through ControlNet~\cite{zhang2023adding}.
Note that, for better efficiency and more stable training, SD applies the diffusion process in the latent space rather than the image space, following the encoder and decoder structure of the VAE~\cite{kingma2014auto}.

During training, given an image pair ($I_\text{HQ}$, $I_\text{LQ}$), a VAE encoder~$\mathcal{E}$, and Gaussian noise with variance $\beta_t \in (0,1)$, $I_\text{HQ}$ is first diffused according to DDPM~\cite{ho2020denoising} as follows:
\vspace{-2mm}
\begin{equation}\label{eq:conventional-diffusion}
    \begin{split}
        z_{t} =&\, q(z_{t} | z_0 = \mathcal{E}(I_\text{HQ})) \\
        =&\,\sqrt{\bar{\alpha}_{t}}\,\mathcal{E}(I_\text{HQ}) + \sqrt{1-\bar{\alpha}_{t}}\,\epsilon,
    \end{split}
\end{equation}
where $q(z_t | z_{t-1}) \coloneqq\,\mathcal{N}(\sqrt{1-\beta_t}\,z_{t-1}, \beta_t\,\mathbf{I})$, $\alpha_t = 1 - \beta_t$, $\bar{\alpha}_t = \prod_{i=1}^{t} \alpha_{i}$, and $\epsilon \sim \mathcal{N}(0,\mathbf{I})$. The timestep $t$ is randomly selected from $[1, T]$, where $T$ is the total number of timesteps.
Then the trainable module, usually ControlNet, is trained to predict the injected noise as follows:
\vspace{-2mm}
\begin{equation}\label{eq:conventional-diffusion-training-loss}
    \begin{split}
        \mathcal{L}_\epsilon =&\, \normtwo{\epsilon - \epsilon_\theta(z_\text{t}, t, \mathcal{E}(I_\text{LQ}))}, \\
    \end{split}
\end{equation}
where $\epsilon_\theta$ is a denoising network combined with ControlNet.
Note that $I_\text{LQ}$ is used only as condition input for ControlNet.

During inference, denoising starts from a \textit{pure noise}.
It gradually moves toward less noisy images by iteratively alternating between denoising and adding noise, using sampling methods~\cite{nichol2021improved, song2020denoising, ho2020denoising}.
% 
% It is well-known that while multiple steps increase inference time, they generally yield better visual quality.
% 
After approximately 50 steps, the process obtains the predicted latent $\hat{z}_0$, which is then mapped back to the image space using the VAE decoder.

\vspace{-1mm}
\subsection{EDTR Architecture}
\label{ssec:arch-edtr}

\vspace{-1mm}
\paragraph{Pre-restoration and partial diffusion.}
As noted earlier, conventional SD-based IR methods begin the denoising process from pure noise during the inference.
Although $I_\text{LQ}$ are used as a condition through ControlNet, its behavior heavily depends on SD's generative capacity.
This poses a challenge for TDIR, as it hinders the use of valuable clues in LQ images, making it difficult to restore task-relevant details.

To address this, we directly incorporate $I_\text{LQ}$ into the diffusion process, \ie, Equation~\eqref{eq:conventional-diffusion}.
However, this approach poses two key challenges:
(1) Since SD was not originally trained on $I_\text{LQ}$, which include complex degradations such as JPEG artifacts, directly incorporating $I_\text{LQ}$ into the diffusion process is ineffective at leveraging the power of SD.
(2) When $I_\text{LQ}$ is fed directly into SD, degradation artifacts can be interpreted as generation cues, producing tasks-irrelevant details.
Instead, we first pre-restore $I_\text{LQ}$ using a pixel-error-based IR method~$\mathcal{R}_\text{pix}$, such as SwinIR~\cite{sr_swinir}:
\vspace{-1.5mm}
\begin{equation}\label{eq:partial-diffusion}
    \begin{split}
        z_\text{pre-res} =&\, \mathcal{E}(\mathcal{R}_\text{pix}(I_\text{LQ})).
    \end{split}
\end{equation}

\begin{figure}[!t]
    \centering
    \subfloat{\includegraphics[width=1.0\linewidth]{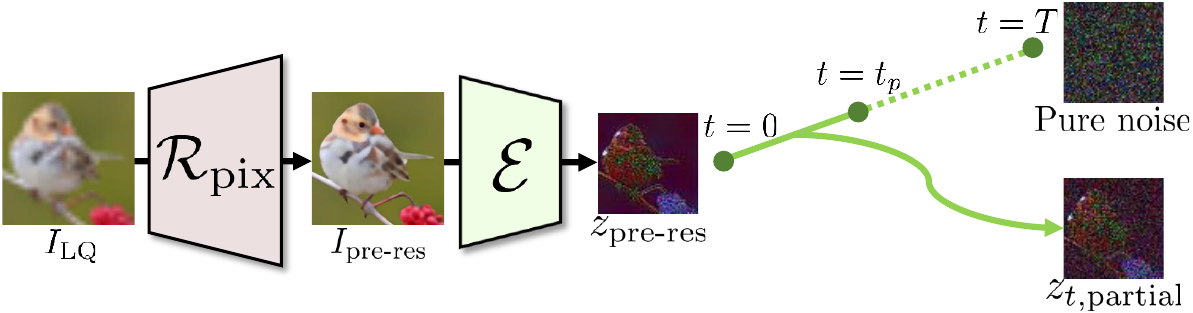}}
    \vspace{-3mm}
    \caption{\textbf{Illustration of pre-restoration and partial diffusion.}
        %
        % We omit the encoding, decoding, and conditioning for simplicity in visualization.
    }
    \label{fig:PR&PD}
    \figxspace
    \vspace{-2mm}
\end{figure}

\vspace{-1.3mm}
Although introducing strong noise in the diffusion process enables the generation of additional information by heavily relying on the diffusion prior, preserving valuable clues in the LQ image is more crucial for the TDIR problem.
Thus, we propose adding \textit{mild} noise to the pre-restored images and starting generation from it.
We denote this technique as partial diffusion, which can be formally defined as:
\vspace{-9.5mm}

\begin{equation}\label{eq:partial-diffusion}
    \begin{split}
        z_{t,\text{partial}} =&\, q(z_{t \leq t_p} | z_0 = z_\text{pre-res}) \\
        =&\,\sqrt{\bar{\alpha}_{t}}\,z_\text{pre-res} + \sqrt{1-\bar{\alpha}_{t}}\,\epsilon,
    \end{split}
\end{equation}
% \vspace{-0.2mm}
% 
where $z_{t,\text{partial}}$ is the partial diffusion result and $t_p$ is a fixed hyper-parameter with a relatively smaller value than $T$.
We empirically set $(t_{p},T)=(200,1000)$.
Figure~\ref{fig:PR&PD} illustrates the proposed diffusion process.
% For the choice of $t_{p}$, please refer to our supplementary materials.
% 

% In addition, it should be noted that, unlike conventional SD-based IR methods, where training and inference differ, \ie, using $z_t$ as input during training but pure noise during inference, we consistently use $z_\text{PD}$.
% % 
% This consistency critically affects the accurate restoration of task-relevant details essential for TDIR.

\vspace{-4mm}
\paragraph{Short-step denoising.}
As noted in the preliminary section, conventional SD-based IR methods employ long-step denoising (\eg, 50).
However, in the context of the TDIR problem, we find that a large number of denoising steps negatively impact on task performance.
Although each step introduces details through the diffusion prior, repeatedly adding such details can generate redundant content, thereby diluting crucial task-relevant information.
Hence, we adopt a short-step denoising (\eg, 1, 4) during inference.
Specifically, for a given number of denoising steps~$n$, we denoise $z_{t_p,\text{partial}}$ by passing it through $n$ timesteps: $\mathcal{T} = [t_p, \lfloor t_p/n \cdot (n-1)\rfloor,..., \lfloor t_p/n \rfloor]$, following \cite{nichol2021improved}, and obtain the final denoising result $z_\text{diff-res}$.
In the one-step denosing case (\ie, $n=1$), this process simplifies to:
\vspace{-1mm}
\begin{equation}\label{eq:one-step-denoising}
    \begin{split}
        z_{\text{diff-res}} = \frac{z_{t_p,\text{partial}} - \sqrt{1-\bar{\alpha}_{t_p}}{\epsilon_\theta}(z_{t_p,\text{partial}}, t_p, z_\text{pre-res})}{\sqrt{\bar{\alpha}_{t_p}}}.
    \end{split}
\end{equation}

\vspace{-3mm}
\noindent Note that conventional SD-based IR methods~\cite{lin2023diffbir, wang2024exploiting} struggle to generate visually pleasing results in short-step denoising scenarios~\cite{wang2024sinsr, wu2024one}.
However, we find that optimizing loss functions, specifically designed to restore task-relevant details as discussed in Section~\ref{sssec:leveraging-generative-prior-for-HT}, enables the achievement of visually pleasing results, even in one-step denoising.

\begin{figure}[!t]
    \centering
    \subfloat[Conventional SD-based IR methods.]{\includegraphics[width=0.9\linewidth]{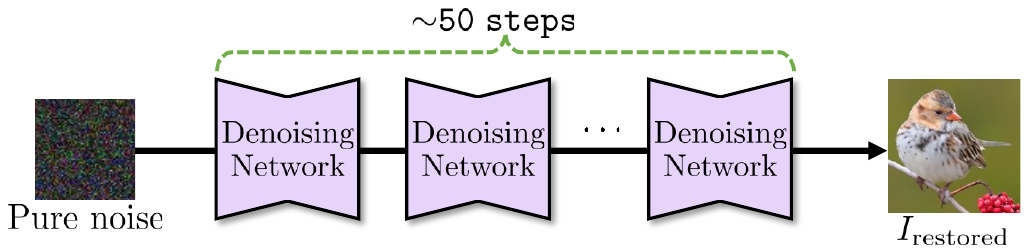}\vspace{-1.8mm}}
    \\
    \subfloat[Pre-restoration, partial diffusion, and short-step denosing.]{\includegraphics[width=0.9\linewidth]{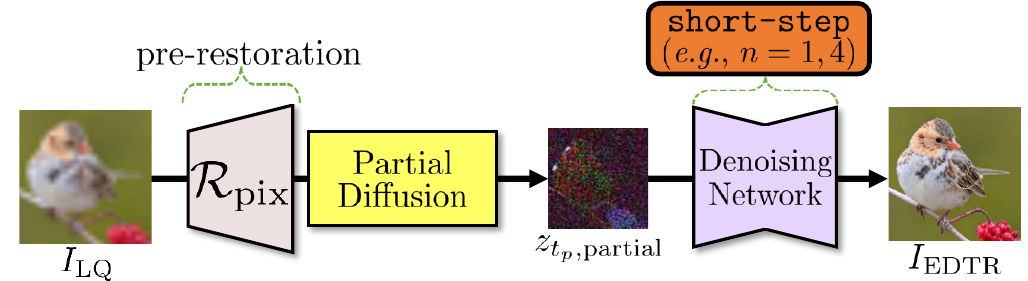}}
    \vspace{-2mm}
    \caption{\textbf{The comparison of IR procedure.}
        We omit the encoding, decoding, and conditioning for simplicity in visualization.
    }
    \label{fig:partial-diffusion}
    \figxspace
    \vspace{-2mm}
\end{figure}

Figure~\ref{fig:partial-diffusion} compares the overall IR procedure.
Unlike conventional methods, which start from pure noise and require around 50 denoising steps, we use the input LQ image with pre-restoration and partial diffusion, followed by short-step denoising.
This approach directly leverages useful clues in LQ images to restore task-relevant details.

% 
% Unlike conventional SD-based IR methods that optimize Equation~\eqref{eq:conventional-diffusion-training-loss}, we optimize loss functions designed to restore task-relevant details, as discussed in Section~\ref{sssec:leveraging-generative-prior-for-HT}.
% %
% Moreover, utilizing these loss terms is key to achieving visually pleasing results, even in the one-step denoising setting.

% Figure~\ref{fig:partial-diffusion} compares the overall IR procedure.
% %
% Unlike conventional SD-based IR methods~\cite{lin2023diffbir, wang2024exploiting}, which start generation from pure noise and require approximately 50 denoising steps, we directly use the input LQ image with pre-restoration and partial diffusion, followed by short-step denoising.
% %
% This approach allows us to leverage useful clues in LQ images to restore task-relevant details.
% 
% 
% Additionally, conventional SD-based IR methods face a discrepancy between training and inference, as they are trained on noisy HQ images but tested on pure noise.
% 
% In contrast, our method eliminates this discrepancy between training and inference.
% 

% 
\vspace{-4mm}
\paragraph{Decoding into RGB space.}
After the denoising process, we transform the $z_\text{diff-res}$ back into the image space using the VAE decoder~$\mathcal{D}$.
We further apply Wavelet color correction~\cite{wang2024exploiting} to achieve a faithful reconstruction as follows:
\vspace{-6mm}
\begin{equation}\label{eq:decoding}
    \begin{split}
        I_{\text{EDTR}} = \mathbf{H}(\mathcal{D}(z_\text{diff-res})) + \mathbf{L}(\mathcal{R}_\text{pix}(I_\text{LQ})),
    \end{split}
\end{equation}
where $I_{\text{EDTR}}$ is the final restored result from our EDTR. The $\mathbf{H}(\cdot)$ and $\mathbf{L}(\cdot)$ correspond to the high-frequency and low-frequency components extracted by Wavelet transform.
Figure~\ref{fig:edtr-overview} illustrates the overall architecture of our EDTR.

\subsection{Training Framework}
\label{ssec:training-edtr-HT}
%
% \subsubsection{Leveraging diffusion prior for high-level vision}
\vspace{-1mm}
\label{sssec:leveraging-generative-prior-for-HT}
\paragraph{High-level feature loss.}
% 
% ARCHIVE
% Moderate diffusion with one-step denoising allows us to limit the excessive generative prior of SD, which could negatively impact HT performance.
% % 
% However, this approach does not ensure that the generative prior is effectively used to enhance HT-relevant details. 
% 
% Although partial diffusion with one-step denoising limits the excessive generative power of SD, it does not guarantee that the diffusion prior is effectively harnessed to restore task-relevant details.
% Although partial diffusion with one-step denoising limits the excessive generation, it does not guarantee that the diffusion prior is effectively harnessed to restore task-relevant details.
% 
The goal of TDIR is to restore task-relevant details to enhance task performance.
However, the noise prediction loss in Equation~\eqref{eq:conventional-diffusion-training-loss}, which is conventionally used for training diffusion models, is not designed for this purpose.
Hence, we introduce high-level feature (HLF) loss, which can guide the model to harness the diffusion prior for restoring task-relevant details.

HLF loss measures the \textit{feature distance} between EDTR restored results and HQ images, following the previous TDIR method~\cite{kim2024beyond}.
Specifically, we compute the HLF loss using the feature space of two task networks: the task network~$\mathcal{H}$, which is optimized during training, and another task network $\mathcal{H}_\text{HQ}$, pre-trained on HQ images, to capture complementary task-relevant information (see Supplementary Section~\ref{sec:complementary-HLF}).
Formally, the HLF loss is defined as:
\vspace{-2mm}
\begin{equation}\label{eq:HLF-loss}
    \begin{split}
        \mathcal{L}_\text{HLF} = \frac{1}{2} ( &\normone{\mathcal{H}^f(I_{\text{EDTR},\text{train}}) - \mathcal{H}^f(I_\text{HQ})} \\
        + &\normone{\mathcal{H}^f_\text{HQ}(I_{\text{EDTR},\text{train}}) - \mathcal{H}^f_\text{HQ}(I_\text{HQ})} ),
    \end{split}
\end{equation}
 
\vspace{-2.2mm}
\noindent where $\mathcal{H}^f$ and $\mathcal{H}^f_\text{HQ}$ are intermediate feature spaces of $\mathcal{H}$ and $\mathcal{H}_\text{HQ}$, respectively.
Note that $I_{\text{EDTR},\text{train}}$ in Equation~\eqref{eq:HLF-loss} represents the restored image from the one-step denoising result of $z_{t,\text{partial}}$, where $t$ is randomly sampled from $\mathcal{T}$ during training, following the conventional diffusion loss in Equation~\eqref{eq:conventional-diffusion-training-loss}, regardless of denoising steps $n$ during inference (see Supplementary Algorithm~\ref{alg:edtr} for details).
The HLF loss updates only EDTR and does not affect $\mathcal{H}$.
% Note that $I_\text{EDTR}$ in Equation~\eqref{eq:HLF-loss} represents the one-step denoising results that enables efficient backpropagation, regardless of the number of denoising steps during the inference phase (see Algorithm~\fakeref{S1} in the supplementary material for details).

\vspace{-3.8mm}
\paragraph{Task loss.}
While training the proposed EDTR to restore task-relevant details, we also train the task networks using the restored outputs.
We utilize both $I_\text{EDTR}$ and HQ images to stabilize training and introduce additional regularization.
Formally, task loss~$\mathcal{L}_\text{task}$ for the task network is defined as:
\vspace{-1mm}
\begin{equation}\label{eq:task-loss}
    \begin{split}
        \mathcal{L}_\text{task} &= f_\text{task} (\mathcal{H}(I_{\text{EDTR}~\oplus~\text{HQ}}), y),
    \end{split}
\end{equation}
where $f_\text{task}$ is the standard loss function for each task, \eg, cross-entropy loss for classification, $I_{\text{EDTR}~\oplus~\text{HQ}}$ represents batch-wise concatenated images, consisting of a half-batch from $I_\text{EDTR}$ and a half-batch from $I_\text{HQ}$, and $y$ are corresponding task labels.
Note that $\mathcal{L_\text{task}}$ updates only the task network~$\mathcal{H}$ and does not update EDTR.

\vspace{-4.0mm}
\paragraph{Feature matching loss.}
To align the feature space of task network $\mathcal{H}$ with $\mathcal{H}_\text{HQ}$, which is well-trained to utilize high-frequency details present in HQ images, we further introduce feature matching~(FM) loss, defined as:
\vspace{-1mm}
\begin{equation}\label{eq:feature-mathcing-loss}
    \begin{split}
        \mathcal{L}_\text{FM} = \normone{\mathcal{H}^f(I_{\text{EDTR}~\oplus~\text{HQ}}) - \mathcal{H}^f_\text{HQ}(I_\text{HQ})}.
    \end{split}
\end{equation}
% 
\iffalse
% Note that the FM loss can be interpreted as a form of cross-quality knowledge distillation~\cite{hinton2015distilling, su2016adapting} at the feature level.
% 
% Similar to the task loss, the FM loss only updates the $\mathcal{H}$.
\fi
For clarity, note that $I_\text{EDTR}$ in Equations~\eqref{eq:task-loss} and \eqref{eq:feature-mathcing-loss} is the $n$ step denoising result, different from the one-step result $I_{\text{EDTR},\text{train}}$ in Equation~\eqref{eq:HLF-loss}.
The $\mathcal{L}_\text{FM}$ can be interpreted as a form of cross-quality knowledge distillation~\cite{hinton2015distilling, su2016adapting} at the feature level. 
Similar to the task loss, $\mathcal{L}_\text{FM}$ only updates $\mathcal{H}$.

\vspace{-4.0mm}
\paragraph{Joint training in an alternating manner.}
Since SD was originally trained to operate in the image space, backpropagating the task loss, such as cross-entropy loss that is not image-based, causes instability to the EDTR.
To avoid this, we train both the EDTR and the task network~$\mathcal{H}$ using each corresponding training loss in an alternating manner~\cite{kim2024beyond}, as illustrated in Figure~\ref{fig:edtr-training}.
Specifically, in each training iteration, we first update the EDTR using the HLF loss, then update $\mathcal{H}$ using a combination of the task loss and the FM loss, which can be formally written as follows:
\vspace{-1mm}
\begin{equation}\label{eq:training-loss}
    \begin{split}
        \min\limits_{\boldsymbol{\theta}_{\text{EDTR}}}\;\mathcal{L}_\text{HLF},\;\;\textit{then}\;\;\min\limits_{\boldsymbol{\theta}_{\mathcal{H}}}\;\mathcal{L}_\text{task} + \alpha\,\mathcal{L}_\text{FM},
    \end{split}
\end{equation}
where $\boldsymbol{\theta}_\text{EDTR}$ and $\boldsymbol{\theta}_{\mathcal{H}}$ are the trainable parameters of the EDTR and task network~$\mathcal{H}$, and $\alpha$ is the hyper-parameter.

%% file: Sections/submission_arXiv/5_experiments.tex
\vspace{-0.1mm}
\section{Experiments}
\label{sec:experiments}
\vspace{-1mm}

We validate our EDTR on three representative high-level vision tasks: image classification, semantic segmentation, and object detection, under scenarios involving image degradation.
Note that the final prediction $\hat{y}$ for each task is obtained as: $\hat{y} = \mathcal{H}(\text{EDTR}(I_\text{LQ}))$, where $\mathcal{H}$ is the trained task network and $I_\text{LQ}$ is the input low-quality~(LQ) images.

\vspace{-4mm}
\paragraph{Datasets.}
We use three high-level vision tasks with representative datasets: CUB200~\cite{WahCUB_200_2011} for classification, and PASCAL VOC2012~\cite{everingham2010pascal} for segmentation and detection.
We construct two degraded LQ image sets:\,1) Mixture-\textit{A}: downsampling combined with JPEG compression, and 2) Mixture-\textit{B}: a combination of downsampling, blur, noise, and compression, following the CodeFormer~\cite{zhou2022towards} degradation.
Details on degradation types are in the second column of Table~\ref{table:main-table} and Section~\ref{sec:degradation} in the supplementary materials.

\vspace{-4.0mm}
\paragraph{Evaluation metrics.}
We use representative evaluation metrics for each task: top-1 accuracy~(Acc) for classification, mean Intersection over Union~(mIoU) for segmentation, and mean Average Precision~(mAP) for detection.
The mAP score is measured following the standard COCO~\cite{data_coco} evaluation setting.
We use two no-reference metrics to assess visual appeal: NIQE~\cite{measure_niqe} and Q-Align~\cite{wu2023qalign}, where the latter leveraging large vision-language models.
We also report the standard PSNR for comparison.

\vspace{-4.0mm}
\paragraph{Network architecture.}
We use StableDiffusion-2.1~\cite{rombach2022high} (SD) and ControlNet~\cite{zhang2023adding} for EDTR.
ControlNet and the VAE decoder of SD are set as trainable modules.
We also incorporate the pixel-error-based SwinIR~\cite{sr_swinir} model as the pre-restoration network. 
For the task network, we employ representative architecture for each task: ResNet~\cite{cls_resnet} for classification, DeepLabV3~\cite{seg_deeplab} for segmentation, and FasterRCNN~\cite{det_fasterrcnn} for detection.
The task network $\mathcal{H}$ shares the same architecture as $\mathcal{H}_\text{HQ}$ (the network pre-trained at HQ), and the weights of $\mathcal{H}$ are initialized from $\mathcal{H}_\text{HQ}$.

% \vspace{-1mm}
\subsection{Experimental Results} \label{subsec:experimental-results}
%
% \vspace{-1.0mm}
Table~\ref{table:main-table} and Figure~\ref{fig:qualitative-results} present quantitative and qualitative results, respectively, across various types of degradation in image classification, semantic segmentation, and object detection.
Note that in Table~\ref{table:main-table}, `Oracle', `\textit{No} restoration', and `SwinIR' refer to the performance when the task network is trained and tested directly on HQ, LQ, and pixel-error-based SwinIR restored images, respectively.
EDTR-$n$\,step refers to the EDTR model with $n$ denoising steps.

\begin{table*}[t!]
\small
\centering
\setlength\tabcolsep{1.0pt}
\def\arraystretch{1.1}
\resizebox{1.00\linewidth}{!}{
    \begin{tabular}{L{3.3cm}|C{2.9cm}|C{1.5cm}|C{1.2cm}C{1.4cm}C{1.2cm}|C{1.5cm}|C{1.2cm}C{1.4cm}C{1.2cm}|C{1.5cm}|C{1.2cm}C{1.4cm}C{1.2cm}}
    \toprule
    \,~\multirow{3}{*}{IR Methods} & \multirow{3}{*}{Degradation type} & \multicolumn{4}{c|}{Image classification} & \multicolumn{4}{c|}{Semantic segmentation} & \multicolumn{4}{c}{Object detection} \\
    \cline{3-14}
    & & Task & \multicolumn{3}{c|}{Visual quality} & Task & \multicolumn{3}{c|}{Visual quality} & Task & \multicolumn{3}{c}{Visual quality} \\
    % \cline{3-14}
    & & Acc$_\uparrow$ (\%) & NIQE$_\downarrow$ & Q-Align$_\uparrow$ & PSNR$_\uparrow$ & mIoU$_\uparrow$ (\%) & NIQE$_\downarrow$ & Q-Align$_\uparrow$ & PSNR$_\uparrow$ &  mAP$_\uparrow$ (\%) & NIQE$_\downarrow$ & Q-Align$_\uparrow$ & PSNR$_\uparrow$ \\
    \hline
    \;~Oracle & None & \cellcolor{cyan!\cyanP}{82.5} & 5.37 & 3.69 & +$\inf$ & \cellcolor{cyan!\cyanP}{67.0} & 4.71 & 4.02 & +$\inf$ & \cellcolor{cyan!\cyanP}{36.9} & 3.70 & 3.95 & +$\inf$ \\
    \hline
    \;~\textit{No} restoration & \multirow{7}{*}{\makecell{Mixture-\textit{A}:\\SR$(s = 8)$~+\\JPEG$(q = 75)$}} & \cellcolor{cyan!\cyanP}{60.8} & 14.08 & 1.03 & 23.72 & \cellcolor{cyan!\cyanP}{47.4} & 13.67 & 1.16 & 22.43 & \cellcolor{cyan!\cyanP}{14.5} & 14.54 & 1.04 & 21.08 \\
    % \;~TDSR~\cite{sr_tdsr} & & 65.8 & 7.72 & 2.47 & 25.10 & 51.1 & 8.00 & 2.32 & 23.08 & 15.9 & 7.02 & 2.19 & 21.65 \\
    \;~SwinIR~\cite{sr_swinir} & & \cellcolor{cyan!\cyanP}{70.0} & 9.97 & 2.55 & \textbf{25.21} & \cellcolor{cyan!\cyanP}{56.2} & 8.66 & 2.54 & \textbf{23.47} & \cellcolor{cyan!\cyanP}{21.0} & 8.96 & 2.22 & \textbf{21.70} \\
    % \;~URIE~\cite{son2020urie} & & \cellcolor{cyan!\cyanP}{} & & & & \cellcolor{cyan!\cyanP}{} & & & & \cellcolor{cyan!\cyanP}{} & & & \\
    \;~TDSR~\cite{sr_tdsr} & & \cellcolor{cyan!\cyanP}{66.1} & 7.53 & 2.51 & \underline{24.99} & \cellcolor{cyan!\cyanP}{51.3} & 7.67 & 2.37 & \underline{23.41} & \cellcolor{cyan!\cyanP}{16.2} & 6.43 & 2.13 & \underline{21.62} \\
    \;~RSRSSN~\cite{zhao2018residual} & & \cellcolor{cyan!\cyanP}{69.2} & 7.32 & 2.75 & 24.04 & \cellcolor{cyan!\cyanP}{53.0} & 6.97 & 2.58 & 23.06 & \cellcolor{cyan!\cyanP}{22.3} & 6.59 & 2.10 & 21.40 \\
    \;~SR4IR~\cite{kim2024beyond} & & \cellcolor{cyan!\cyanP}{71.5} & 5.41 & 3.37 & 24.07 & \cellcolor{cyan!\cyanP}{56.2} & \textbf{4.80} & 3.16 & 22.85 & \cellcolor{cyan!\cyanP}{22.3} & 5.38 & 3.00 & 20.95 \\
    \;~\textbf{EDTR}-1\,step \textbf{(Ours)} & & \cellcolor{cyan!\cyanP}{\textbf{74.4}} & \underline{4.68} & \underline{3.53} & 23.63 & \cellcolor{cyan!\cyanP}{\underline{64.1}} & 5.30 & \underline{3.59} & 22.33 & \cellcolor{cyan!\cyanP}{\underline{31.2}} & \underline{4.62} & \underline{3.68} & 20.60 \\
    \;~\textbf{EDTR}-4\,step \textbf{(Ours)} & & \cellcolor{cyan!\cyanP}{\underline{74.1}} & \textbf{4.26} & \textbf{3.81} & 22.46 & \cellcolor{cyan!\cyanP}{\textbf{65.3}} & \underline{5.17} & \textbf{3.91} & 21.48 & \cellcolor{cyan!\cyanP}{\textbf{33.4}} & \textbf{4.19} & \textbf{4.01} & 19.89 \\
    \hline
    \;~\textit{No} restoration & \multirow{7}{*}{\makecell{Mixture-\textit{B}:\\SR$(s \in [1, 16])$~+\\Blur$(\sigma \in [0, 8])$~+\\Noise$(\sigma \in [0, 10])$~+\\JPEG$(q \in [50, 100])$}} & \cellcolor{cyan!\cyanP}{47.6} & 15.11 & 1.08 & 23.05 & \cellcolor{cyan!\cyanP}{40.2} & 14.84 & 1.19 & 22.10 & \cellcolor{cyan!\cyanP}{0.0} & 15.24 & 1.07 & 20.86 \\
    % \;~TDSR~\cite{sr_tdsr} & & 57.2 & 7.72 & 2.20 & 24.84 & 45.3 & 8.43 & 2.18 & 23.61 & 14.6 & 6.61 & 1.90 & 22.07 \\
    \;~SwinIR~\cite{sr_swinir} & & \cellcolor{cyan!\cyanP}{60.7} & 10.26 & 2.07 & \textbf{25.04} & \cellcolor{cyan!\cyanP}{50.4} & 9.47 & 2.30 & \textbf{23.67} & \cellcolor{cyan!\cyanP}{22.1} & 9.60 & 1.92 & \textbf{22.20} \\
    % \;~URIE~\cite{son2020urie} & & \cellcolor{cyan!\cyanP}{} & & & & \cellcolor{cyan!\cyanP}{} & & & & \cellcolor{cyan!\cyanP}{} & & & \\
    \;~TDSR~\cite{sr_tdsr} & & \cellcolor{cyan!\cyanP}{56.4} & 7.29 & 2.20 & \underline{24.68} & \cellcolor{cyan!\cyanP}{46.0} & 8.07 & 2.13 & \underline{23.55} & \cellcolor{cyan!\cyanP}{14.6} & 6.36 & 1.94 & \underline{21.97} \\
    \;~RSRSSN~\cite{zhao2018residual} & & \cellcolor{cyan!\cyanP}{61.4} & 7.35 & 2.35 & 23.35 & \cellcolor{cyan!\cyanP}{48.2} & 7.72 & 2.36 & 23.01 & \cellcolor{cyan!\cyanP}{23.1} & 6.81 & 1.86 & 21.68 \\
    \;~SR4IR~\cite{kim2024beyond} & & \cellcolor{cyan!\cyanP}{63.4} & 6.08 & 3.11 & 23.62 & \cellcolor{cyan!\cyanP}{51.0} & \textbf{4.72} & 3.15 & 22.71 & \cellcolor{cyan!\cyanP}{22.6} & 4.52 & 3.14 & 20.97 \\
    \;~\textbf{EDTR}-1\,step \textbf{(Ours)} & & \cellcolor{cyan!\cyanP}{\textbf{68.8}} & \underline{4.75} & \underline{3.48} & 23.03 & \cellcolor{cyan!\cyanP}{\underline{60.4}} & 6.36 & \underline{3.55} & 22.04 & \cellcolor{cyan!\cyanP}{\underline{30.6}} & \underline{4.48} & \underline{3.64} & 20.70 \\
    \;~\textbf{EDTR}-4\,step \textbf{(Ours)} & & \cellcolor{cyan!\cyanP}{\underline{68.4}} & \textbf{4.46} & \textbf{3.49} & 23.22 & \cellcolor{cyan!\cyanP}{\textbf{62.9}} & \underline{5.80} & \textbf{3.74} & 22.24 & \cellcolor{cyan!\cyanP}{\textbf{31.9}} & \textbf{4.12} & \textbf{4.02} & 20.29 \\
    \bottomrule
    \end{tabular}
}
\vspace{-0.3cm}
\caption{
    \textbf{Performance of high-level vision tasks and quality of restored image results under various degradation types.}
    We indicate the best and second-best results with \textbf{bold} and \underline{underlined} formatting, respectively.
}
% \vspace{-0.5cm}
% \vspace{-0.2cm}
\label{table:main-table}
\end{table*}
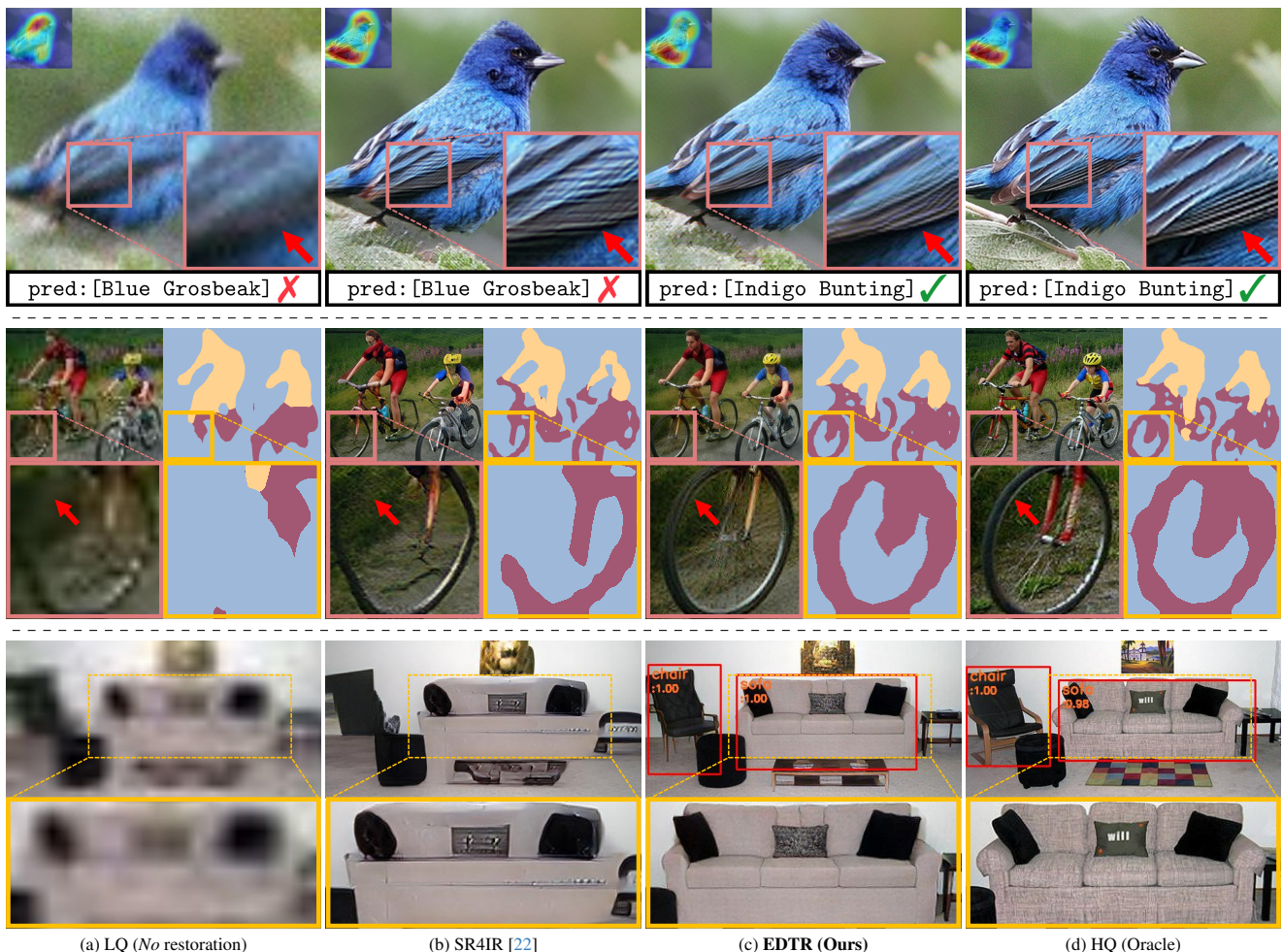
\begin{figure*}[t!]
    \centering
    \captionsetup[subfigure]{labelfont=scriptsize, textfont=scriptsize}
    \renewcommand{\wp}{0.247}
        \subfloat{\includegraphics[width=\wp\linewidth]{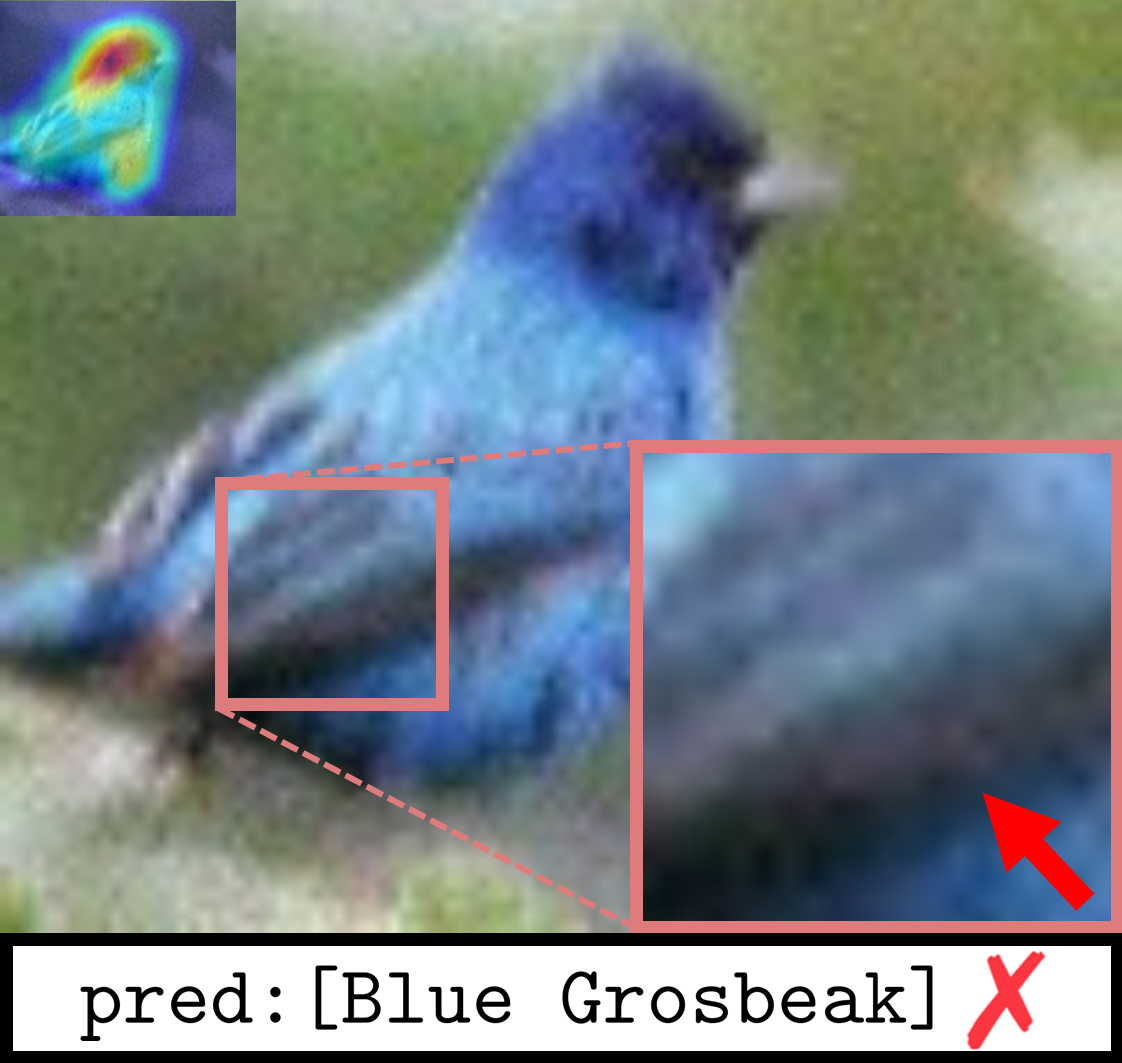}}
        \hfill
        \subfloat{\includegraphics[width=\wp\linewidth]{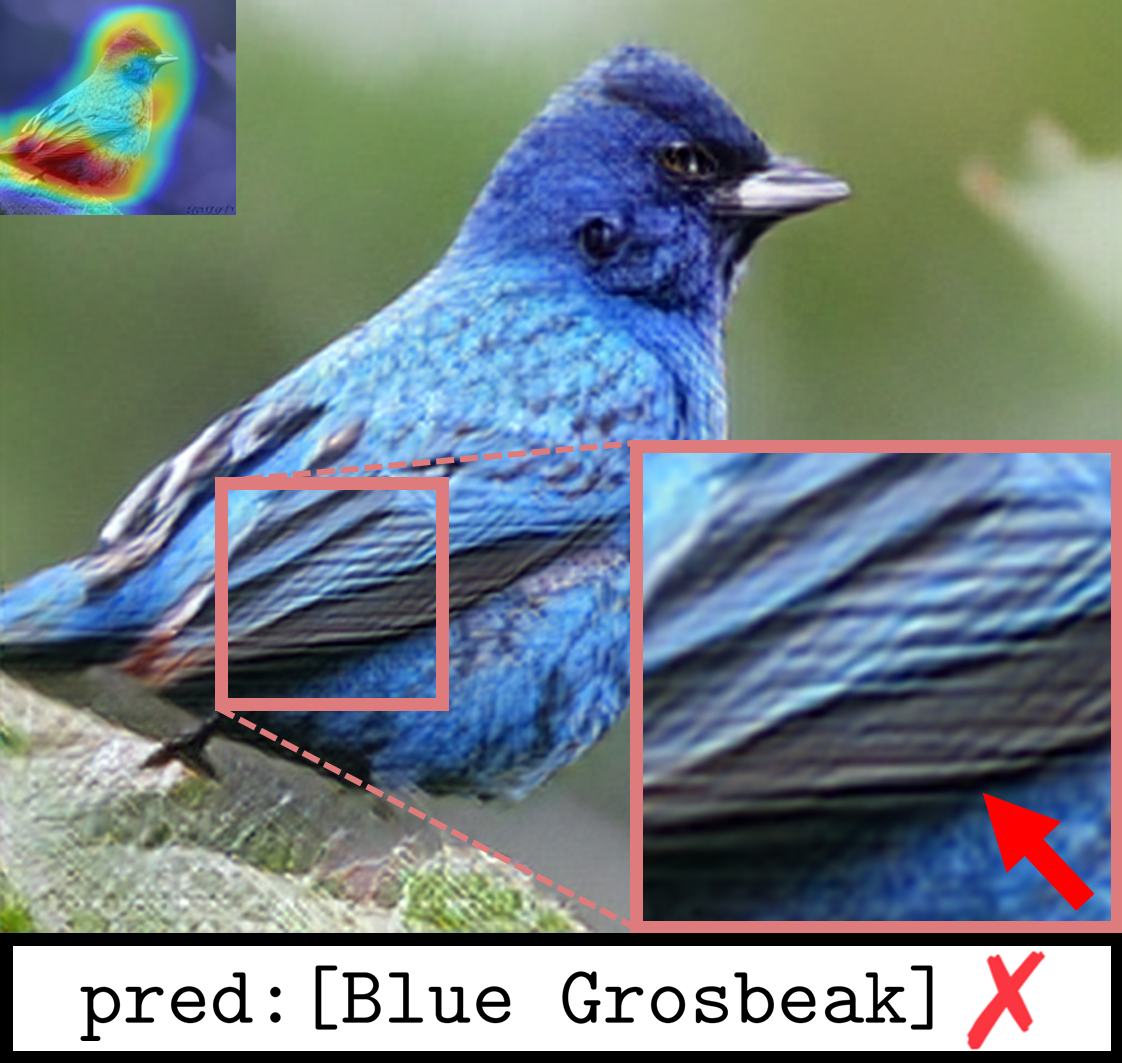}}
        \hfill
        \subfloat{\includegraphics[width=\wp\linewidth]{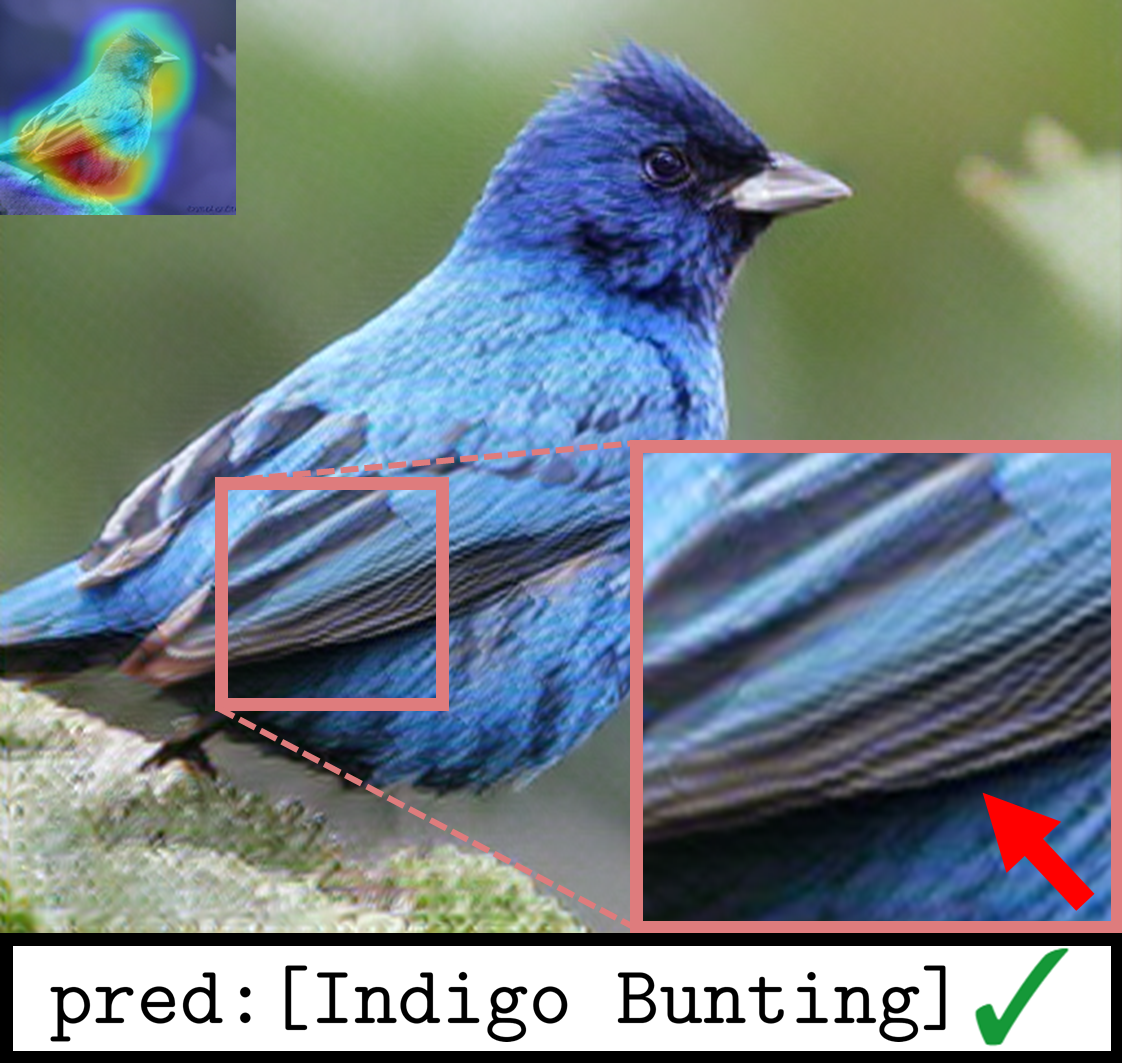}}
        \hfill
        \subfloat{\includegraphics[width=\wp\linewidth]{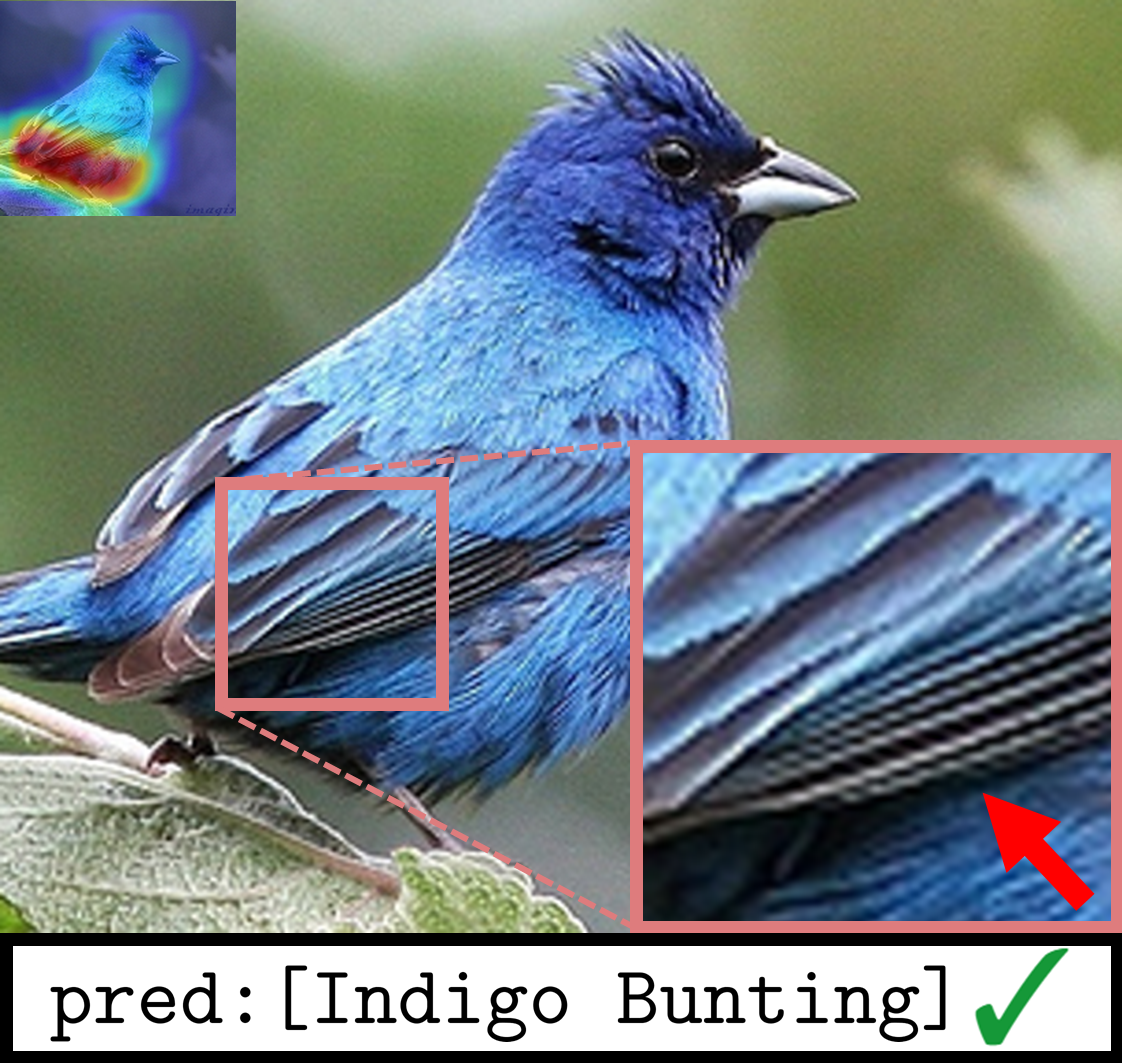}}
        \addtocounter{subfigure}{-4}
        \\
        \vspace{-2.5mm}
        \begin{tikzpicture}
            \draw[dashed] (0,0) -- (17.2,0);
        \end{tikzpicture}
        \vspace{1mm}
        \\
        % \subfloat{\includegraphics[width=\wp\linewidth]{Figures/submission/qualitative/SEG/C07/SEG-LQ-C07.png}}
        % \hfill
        % \subfloat{\includegraphics[width=\wp\linewidth]{Figures/submission/qualitative/SEG/C07/SEG-SR4IR-C07.png}}
        % \hfill
        % \subfloat{\includegraphics[width=\wp\linewidth]{Figures/submission/qualitative/SEG/C07/SEG-EDTR-C07.png}}
        % \hfill
        % \subfloat{\includegraphics[width=\wp\linewidth]{Figures/submission/qualitative/SEG/C07/SEG-Oracle-C07.png}}
        % \addtocounter{subfigure}{-4}
        \subfloat{\includegraphics[width=\wp\linewidth]{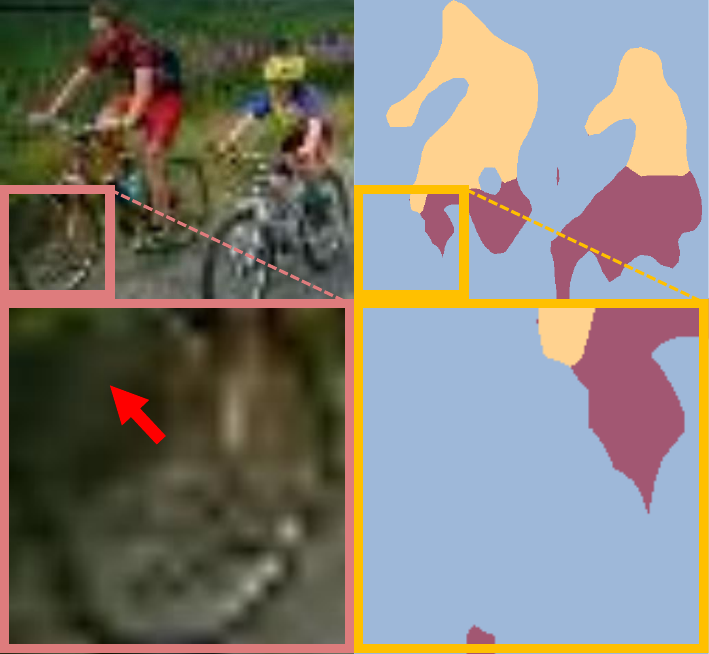}}
        \hfill
        \subfloat{\includegraphics[width=\wp\linewidth]{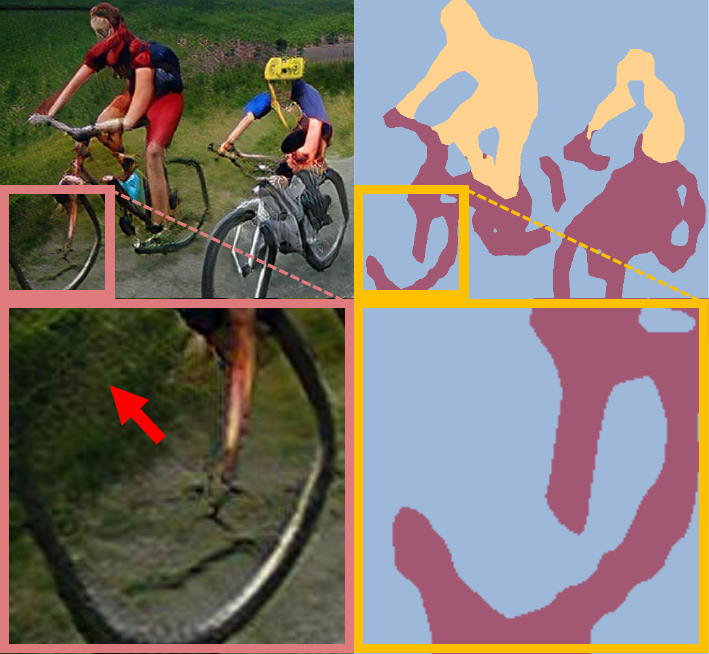}}
        \hfill
        \subfloat{\includegraphics[width=\wp\linewidth]{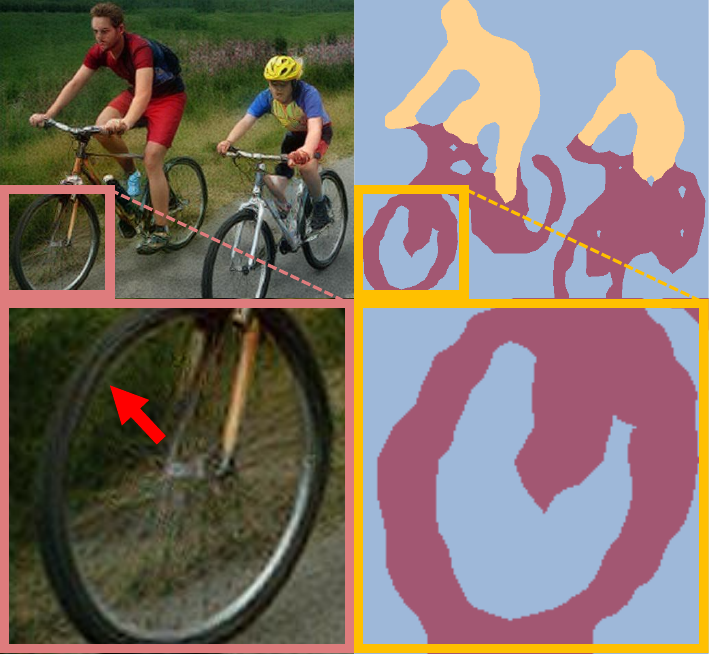}}
        \hfill
        \subfloat{\includegraphics[width=\wp\linewidth]{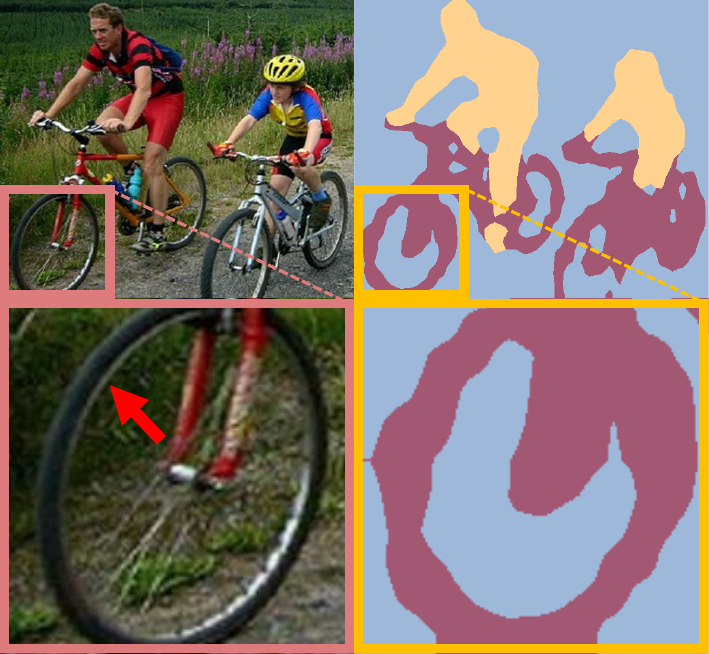}}
        \addtocounter{subfigure}{-4}
        \\
        \vspace{-2.5mm}
        \begin{tikzpicture}
            \draw[dashed] (0,0) -- (17.2,0);
        \end{tikzpicture}
        \vspace{1mm}
        \\
        \subfloat[LQ~(\textit{No} restoration)]{\includegraphics[width=\wp\linewidth]{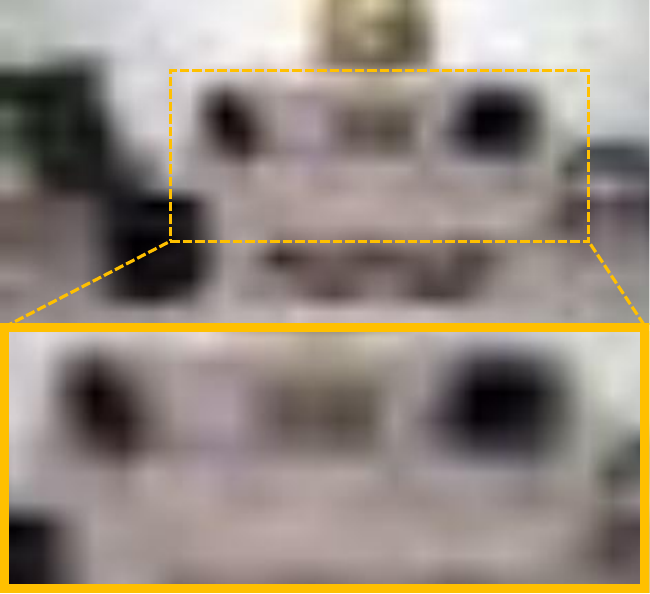}}
        \hfill
        \subfloat[SR4IR~\cite{kim2024beyond}]{\includegraphics[width=\wp\linewidth]{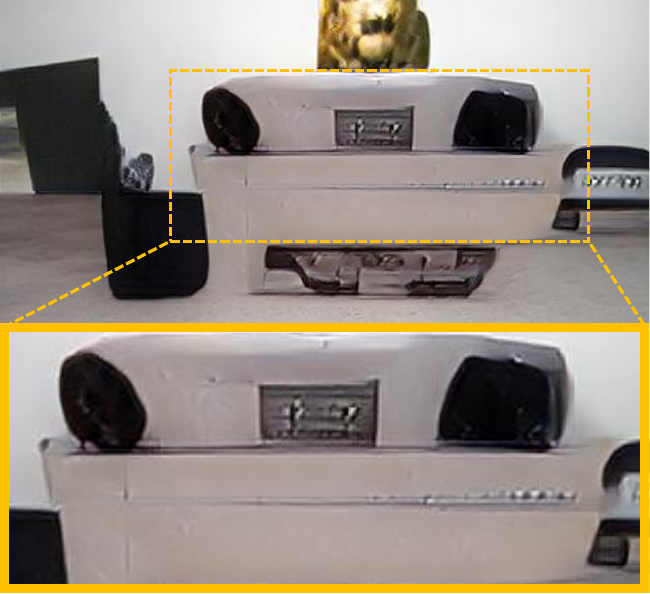}}
        \hfill
        \subfloat[\textbf{EDTR~(Ours)}]{\includegraphics[width=\wp\linewidth]{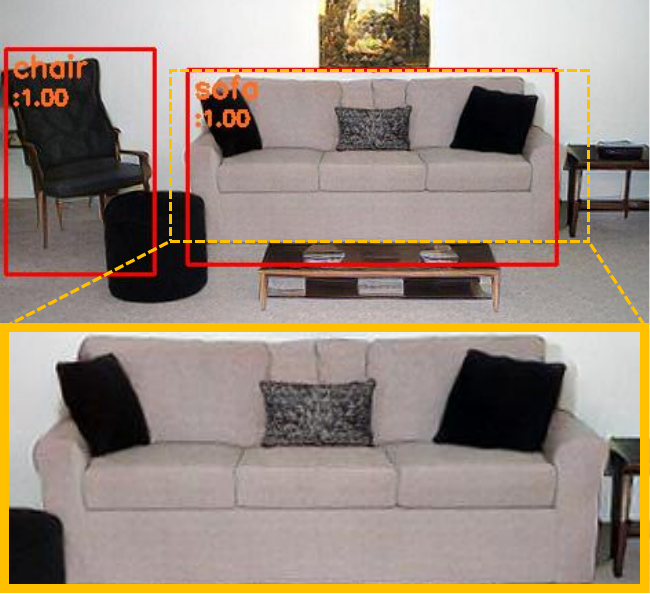}}
        \hfill
        \subfloat[HQ~(Oracle)]{\includegraphics[width=\wp\linewidth]{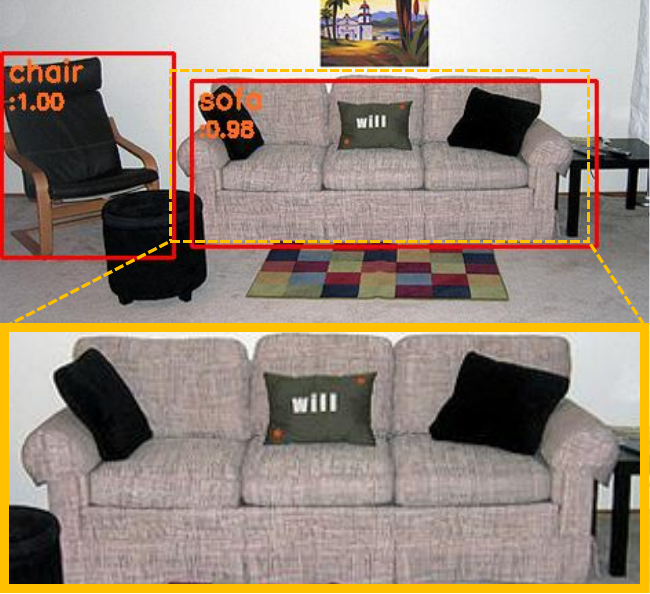}}
    \vspace{-0.3cm}
    \caption{\textbf{Visualization of images and diverse high-level vision task results on degraded LQ (Mixture-\textit{B}) images.}
        The first, second, and third rows represent the results of image classification, semantic segmentation, and object detection, respectively.
        We show the restored images and the corresponding predicted labels.
        For visualization, we use the EDTR-1\,step model for classification and the EDTR-4\,step model for segmentation and detection.
        For more visualization results, please refer to Section~\ref{sec:further-visualization} of our supplementary materials.
    }
    \vspace{-4mm}
    \label{fig:qualitative-results}
\end{figure*}
\vspace{-1mm}

\vspace{-3mm}
\paragraph{Image classification.}
Table~\ref{table:main-table} shows that the proposed EDTR significantly outperforms previous methods under both types of degradation, achieving classification accuracy improvements of +2.9\% and +5.4\% over the previous state-of-the-art task-driven image restoration~(TDIR) method, SR4IR~\cite{kim2024beyond}.
Performance gains increase as the degradation becomes more severe, highlighting our method's efficacy in practical scenarios.
Furthermore, EDTR achieves the highest scores in image quality assessments such as NIQE and Q-Align, indicating its effectiveness in producing visually pleasing results.
However, despite the superior task performance, EDTR does not achieve the highest PSNR score, suggesting that PSNR is a \textit{less relevant metric in TDIR}.
As presented in the first row of Figure~\ref{fig:qualitative-results}, our EDTR, leveraging a strong natural image prior, successfully restores detailed wing patterns—an important area for classification as shown by the Grad-CAM~\cite{selvaraju2017grad} in the upper-left—and correctly predicts the label, whereas the previous method fails to restore these details and misclassifies the image.

\begin{table*}[t!]
\footnotesize
\centering
\setlength\tabcolsep{1.0pt}
\def\arraystretch{1.1}
\resizebox{1.0\linewidth}{!}{
    \begin{tabular}{C{1.2cm}|C{1.2cm}|C{1.3cm}C{1.3cm}|C{2.2cm}C{2.0cm}C{2.4cm}|C{1.4cm}C{1.4cm}C{1.4cm}}
    \toprule
    \multirow{2}{*}{Exp} & \multirow{2}{*}{SD} & \multicolumn{2}{c|}{Training loss} & \multicolumn{3}{c|}{Diffusion process\;\;\;\;} & \multirow{2}{*}{Acc$_\uparrow$ (\%)} & \multirow{2}{*}{$f_d$$_\downarrow$} & \multirow{2}{*}{Q-Align$_\uparrow$} \\
     & & \multirow{1}{*}{HLF} & \multirow{1}{*}{FM} & Pre-restoration & \multirow{1}{*}{Partial diffusion} & \multirow{1}{*}{\# of denoising steps} & & & \\
    \hline
    (1) & \cmark & \xmark & \cmark & \xmark & \xmark & 50 & 64.3 & 0.641 & \textbf{3.87} \\
    \hline
    (2) & \cmark & \cmark & \cmark & \xmark & \xmark & 50 & 62.9 & 0.754 & 2.37 \\
    (3) & \cmark & \cmark & \cmark & \xmark & \xmark & 1 & 63.5 & 0.579 & 3.18 \\
    (4) & \cmark & \cmark & \cmark & \xmark & \cmark & 1 & 66.6 & 0.530 & 3.38 \\
    \hline
    (5) & \cmark & \xmark & \cmark & \cmark & \cmark & 1 & 64.5 & 0.719 & 2.18 \\
    (6) & \cmark & \cmark & \xmark & \cmark & \cmark & 1 & 67.6 & 0.507 & 3.48 \\
    \hline
    (7) & \xmark & \cmark & \cmark & \cmark & \cmark & 1 & 64.2 & 0.541 & 3.20 \\
    \hline
    \cellcolor{cyan!\cyanP}{\textbf{Ours}} & \cellcolor{cyan!\cyanP}{\cmark} & \cellcolor{cyan!\cyanP}{\cmark} & \cellcolor{cyan!\cyanP}{\cmark} & \cellcolor{cyan!\cyanP}{\cmark} & \cellcolor{cyan!\cyanP}{\cmark} & \cellcolor{cyan!\cyanP}{1} & \cellcolor{cyan!\cyanP}{\textbf{68.8}} & \cellcolor{cyan!\cyanP}{\textbf{0.507}} & \cellcolor{cyan!\cyanP}{3.48} \\
    \bottomrule
    \end{tabular}
}
\vspace{-3mm}
\caption{
    \textbf{Ablation study on the impact of each component on EDTR classification performance.}
}
\vspace{-4mm}
\label{table:ablation-study}
\end{table*}

\vspace{-4mm}
\paragraph{Semantic segmentation.}
The performance improvement in segmentation is more notable than in classification, with mIoU gains of +9.1\% and +11.9\% over the previous state-of-the-art.
Although our EDTR achieves the second-best score in NIQE, it consistently achieves the highest Q-Align score, demonstrating its capability to generate perceptually realistic images.
In addition, unlike classification, increasing the number of denoising steps from 1 to 4 leads to a task performance improvement.
In the second row of Figure~\ref{fig:qualitative-results}, the LQ image is severely degraded, causing the details in the upper-left part of the bike wheel to be almost lost.
While the previous TDIR method fails to restore the missing details, our method effectively restores them by leveraging a strong natural image prior, correctly recognizing that they belong to the wheel.
As a result, EDTR produces a segmentation prediction that correctly maps the upper-left part of the bike wheel, demonstrating its effectiveness.

\vspace{-4mm}
\paragraph{Object detection.}
%  
% Our EDTR shows a surprisingly significant performance gain over the previous state-of-the-art in the Mixture-\textit{A} setting, increasing mAP from 22.3\% to 33.4\% (an improvement of approximately \textit{50\%}).
Our EDTR demonstrates a surprisingly significant performance gain over the previous state-of-the-art in the Mixture-\textit{A} setting, increasing mAP from 22.3\% to 33.4\%—\textit{an improvement of approximately 50\%}.
As shown in the third row of Figure~\ref{fig:qualitative-results}, EDTR correctly restores the shape of the sofa by leveraging the powerful diffusion prior, successfully detecting the sofa in the room, whereas the previous method fails.
These results demonstrate that our method consistently achieves superior task performance and visual quality under severe degradation.

\subsection{Ablation studies} 
% \vspace{-1mm}
% 
All experiments use the Mixture-\textit{B} degradation setting.
\vspace{-1mm}
\subsubsection{Effect of proposed components.}
Table~\ref{table:ablation-study} shows the classification performance of our EDTR according to the use of each component.
Specifically, when a proposed component is not used, \ie, \xmark, we follow the conventional SD-based IR method setting, as described in Section~\ref{ssec:preliminary}.
Except for Exp-(7), we adopt the FM loss as a default, since it relates to task network training rather than the IR method.
In addition, to assess how well task-relevant details are restored, we measure the distance in the feature space of the task network $\mathcal{H}_\text{HQ}$, which is pre-trained on HQ images.
We denote this fidelity metric as feature distance, defined as $f_d = \normone{\mathcal{H}_\text{HQ}^f(I_\text{EDTR}) - \mathcal{H}_\text{HQ}^f(I_\text{HQ})}$.

Exp-(1) shows performance without any of the proposed components, \ie, conventional SD-based IR method~\cite{lin2023diffbir}.
Compared to EDTR performance (last row), although it achieves a more perceptually realistic restoration, as indicated by the higher Q-Align score, its large $f_d$ value suggests that task-relevant details are not well restored, resulting in a 4.5\% worse classification accuracy.

\begin{figure}[t!]
    \centering
    \subfloat[LQ]{\includegraphics[width=0.24\linewidth]{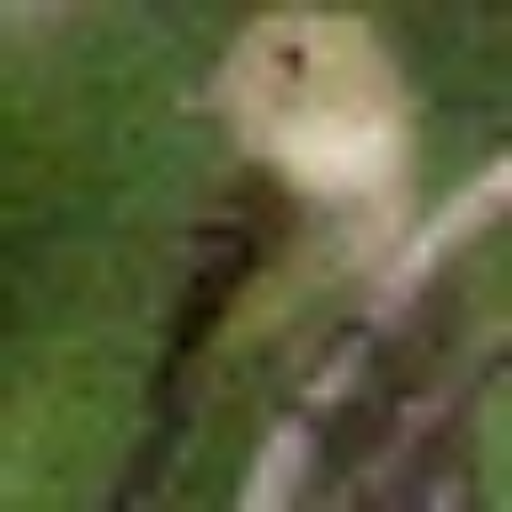}}
    \hfill
    \subfloat[Exp-(2)]{\includegraphics[width=0.24\linewidth]{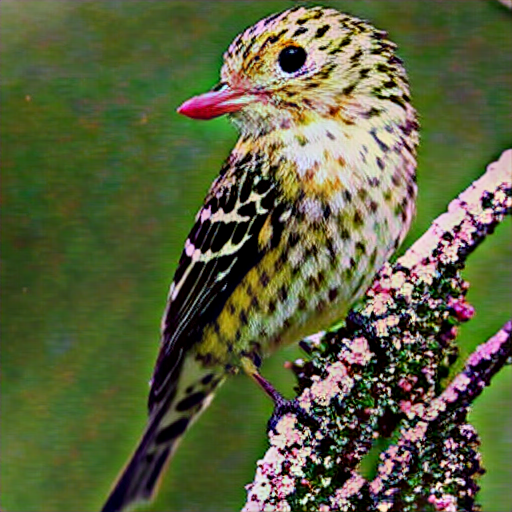}}
    \hfill
    \subfloat[\textbf{EDTR (Ours)}]{\includegraphics[width=0.24\linewidth]{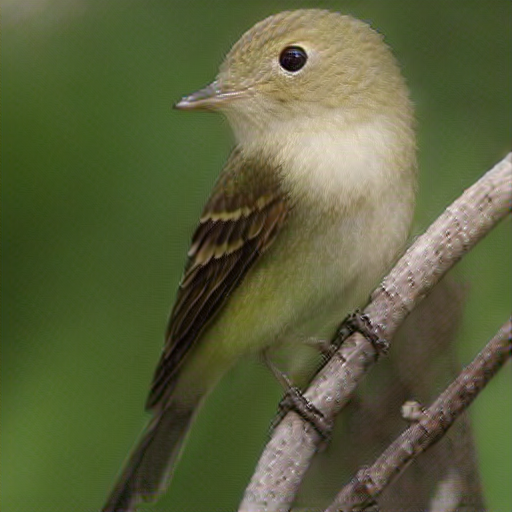}}
    \hfill
    \subfloat[HQ]{\includegraphics[width=0.24\linewidth]{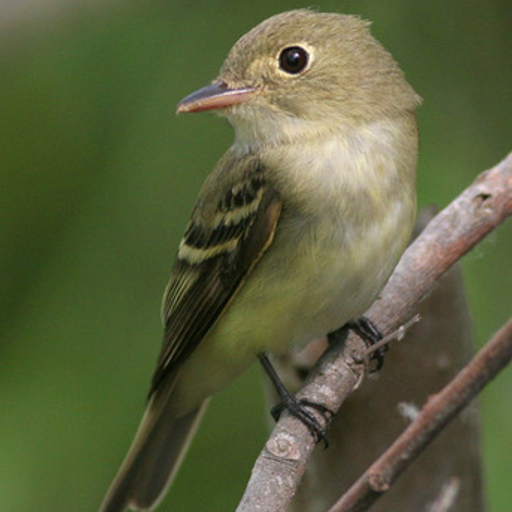}}
    \\
    \figspace
    \vspace{-1mm}
    \caption{
        \textbf{Visualization comparison in classification.}
    }
    \label{fig:vis-comp-cls}
    \figxspace
    \vspace{-3mm}
\end{figure}

\vspace{-4mm}
\paragraph{Effect of pre-restoration \& partial diffusion.}
Exp-(2) shows performance without our design choice for the diffusion process, \ie, SD is directly applied to TDIR.
Although it incorporates a loss for restoring task-relevant details, its task performance is even worse than that of Exp-(1).
Figure~\ref{fig:vis-comp-cls} illustrates that Exp-(2) leads to excessive generation and produces non-existent content, in contrast to the results of EDTR, ultimately failing to improve task performance.
This highlights the necessity of a tailored diffusion process design for TDIR.
Exp-(3) shows that even with short-step denoising, EDTR without partial diffusion does not perform well for TDIR, demonstrating the importance of partial diffusion.
Exp-(4) further shows that the pre-restoration process is also essential for achieving both strong task performance and good visual quality.
For performance regarding the pre-restoration network, see Supplementary Section~\ref{sec:impact-pixel-error}.

\vspace{-4mm}
\paragraph{Effect of HLF \& FM losses.}
Exp-(5) and Exp-(6) show EDTR performance without HLF and FM losses, respectively.
In both cases, task performance drops significantly, highlighting the importance of each training loss term.
Interestingly, Exp-(5) results in a notably lower Q-Align score, indicating that the \textit{HLF loss is crucial for achieving visually pleasing results}, and it enables the generation of such results even in the challenging one-step denoising scenario.
This finding is consistent with previous research~\cite{song2023consistency}, as the HLF loss shares similarities with LPIPS~\cite{zhang2018unreasonable} in measuring distance in feature space.
In addition, incorporating feature spaces from two networks in the HLF loss (as in Equation~\eqref{eq:HLF-loss}) yields a +1.0\% gain compared to using only one (see Supplementary Section~\ref{sec:complementary-HLF} for details).

\vspace{-4mm}
\paragraph{Effect of SD.}
Exp-(7) shows the performance of EDTR without pre-trained SD weights.
Specifically, we remove the frozen denoising network (purple module in Figure~\ref{fig:edtr-overview}); thus, the total number of trainable parameters remains nearly the same.
Compared to this setting, EDTR largely reduces the $f_d$, resulting in a significant 4.6\% increase in classification accuracy.
These results highlight the crucial role of the diffusion prior in restoring task-relevant details.

\subsubsection{Effect of the number of denoising steps.}
\label{abl:number-diffusion-step}
Table~\ref{table:number-denoising-EDTR} shows the performance of EDTR with varying numbers of denoising steps during inference.
As shown in the table, a larger number of denoising steps~(\eg, 50) leads to decreased task performance.
This supports our claim that repeatedly introducing details is not beneficial for TDIR and may even dilute crucial task-relevant information.
In addition, we observe that short-step denoising ($n=4$) is sufficient to produce visually pleasing results, as indicated by its favorable Q-Align score—further confirming the effectiveness of the HLF loss in guiding perceptual quality.

\subsection{Generalizability of EDTR}
% 
% \vspace{-1mm}
\paragraph{Performance of EDTR across various tasks.}
In Table~\ref{table:performance-EDTR-various-tasks}, we present the performance of EDTR-1\,step models, each trained separately on a specific task.
Note that the diagonal (\textcolor{cyan!60}{\textbf{blue}}) values represent the performance of the original EDTR on its respective task.
Since the original EDTR is independently trained for a single task, it performs well on its target task but less effectively on others.
To demonstrate that our EDTR can be extended for broader applicability, we additionally train a single model on all three high-level vision tasks simultaneously, referring to this model as $\text{EDTR}\_\textit{All-in-One}$.
The last row in the table presents the results.
We observe that this model achieves performance comparable to, or even better than, the original EDTR across all three tasks, highlighting its potential for generalization to a variety of applications.

\vspace{-4mm}
\paragraph{Generalization toward unseen datasets.}
\label{abl:unseen}
Table~\ref{table:unseen} shows the classification performance on unseen datasets during training.
Specifically, we train the IR models using Tiny-ImageNet\textsuperscript{\textdagger}\footnote{Since the original Tiny-ImageNet~\cite{le2015tiny} has a resolution of 64x64, which is too small for IR problem, we resampled from ImageNet~\cite{data_imagenet} with the same classes and number of images to create Tiny-ImageNet\textsuperscript{\textdagger}.}~\cite{le2015tiny} and evaluate them on CUB200~\cite{WahCUB_200_2011} and StanfordCars~\cite{krause20133d} datasets, which were not seen during training.
When evaluating on these unseen datasets, the task networks trained on HQ images corresponding to each dataset are used.
The results show that our EDTR significantly outperforms other methods on these unseen datasets, demonstrating the robustness of our approach.

\begin{figure}[t!]
    \centering
    \subfloat[LQ~(\textit{No} restoration)\label{fig:real-world-no-restoration}]{\includegraphics[width=0.48\linewidth]{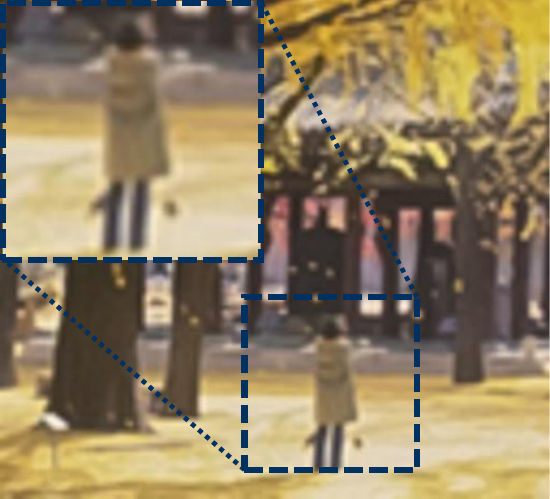}}
    \hfill
    \subfloat[\textbf{EDTR (Ours)}\label{fig:real-world-edtr}]{\includegraphics[width=0.48\linewidth]{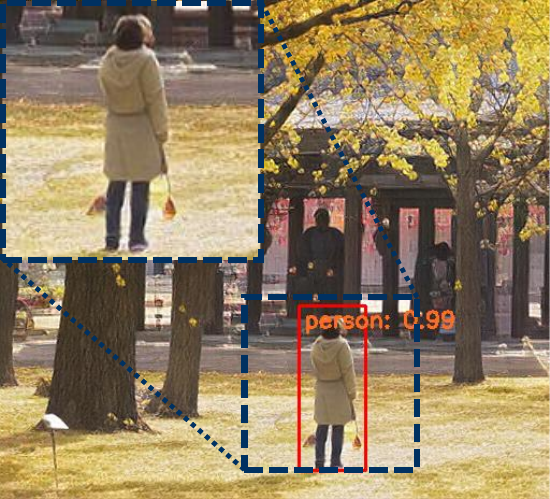}}
    \\
    \figspace
    \vspace{-1mm}
    \caption{
        \textbf{Real-world object detection results and visualization of the restored image.}
        The image is from $\texttt{"56.png"}$ in the real-world image set RealPhoto60~\cite{yu2024scaling}.
    }
    \label{fig:real-world}
    \figxspace
    \vspace{-2mm}
\end{figure}

\vspace{-4mm}
\paragraph{Results on real-world images.}
\label{abl:real-world}
Figure~\ref{fig:real-world} presents real-world object detection results and restored image visualizations.
We compare (a) object detection without restoration and (b) EDTR-4\,step trained on Mixture-\textit{B} degradation.
Due to complex real-world degradation, detection without restoration fails to identify the person in the image.
In contrast, EDTR effectively restores key high-frequency details, such as clothing wrinkles on a person, enabling successful detection.
These results demonstrate that EDTR can handle complex real-world degradations and detect objects in degraded images by effectively leveraging the powerful diffusion prior.
For more visualization results, please refer to Section~\ref{sec:further-visualization} in our supplementary materials.
\begin{table}[t!]
\footnotesize
\centering
\setlength\tabcolsep{1.0pt}
\def\arraystretch{1.1}
\resizebox{1.0\linewidth}{!}{
    \begin{tabular}{L{4.1cm}|C{1.4cm}|C{1.4cm}|C{1.4cm}|C{1.4cm}}
    \toprule
    \;~\# of denoising steps & 4 & 30 & 40 & 50 \\
    \hline
    \;~Classification ($\text{Acc$_\uparrow$}$~/~Q-Align$_\uparrow$) & \textbf{68.4} / \textbf{3.49}  & 67.7 / 3.46 & 67.1 / 3.46 & 66.8 / 3.45 \\
    \;~Segmentation ($\text{mIoU$_\uparrow$}$~/~Q-Align$_\uparrow$) & \textbf{62.9} / \textbf{3.74} & 62.4 / 3.54 & 62.4 / 3.46 & 61.8 / 3.43 \\
    \;~Detection ($\text{mAP$_\uparrow$}$~/~Q-Align$_\uparrow$) & \textbf{31.9} / \textbf{4.02}  & 31.8 / 3.97 & 31.8 / 3.99 & 31.6 / 3.94 \\
    \bottomrule
    \end{tabular}
}
\vspace{-3mm}
\caption{
    \textbf{Effect of the number of denoising steps in EDTR.}
    \vspace{-1mm}
}
\vspace{-1mm}
\label{table:number-denoising-EDTR}
\end{table}
\begin{table}[t!]
\footnotesize
\centering
\setlength\tabcolsep{1.0pt}
\def\arraystretch{1.1}
\resizebox{1.0\linewidth}{!}{
    \begin{tabular}{L{2.8cm}|C{2.4cm}C{2.4cm}C{2.4cm}}
    \toprule
    \multirow{2}{*}{\diagbox[width=2.87cm]{\;~Used model}{Eval task\;\;}} & Classification & Segmentation & Detection \\
    & (Acc$_\uparrow$) & (mIoU$_\uparrow$) & (mAP$_\uparrow$) \\
    \hline
    \;~$\text{EDTR}$\_\textit{Classification} & \cellcolor{cyan!\cyanP}\textbf{68.8} & 48.7 & 14.3 \\
    \;~$\text{EDTR}$\_\textit{Segmentation} & 56.8 & \cellcolor{cyan!\cyanP}\textbf{60.4} & 24.8 \\
    \;~$\text{EDTR}$\_\textit{Detection} & 59.3 & 58.9 & \cellcolor{cyan!\cyanP}\textbf{30.6}  \\
    \hline
    \;~$\text{EDTR}$\_\textit{All-in-One} & {68.4} & {61.0} & {30.4}  \\
    \bottomrule
    \end{tabular}
}
\vspace{-3mm}
\caption{
    \textbf{Performance of EDTR models across various tasks.}
    $\text{EDTR}$\_\textit{Task} represents the EDTR model optimized for each task.
    \vspace{-1mm}
    % The diagonal (\textcolor{gray}{gray}) values represent the performance of the original EDTR on its respective task.
}
\vspace{-1mm}
\label{table:performance-EDTR-various-tasks}
\end{table}
\begin{table}[!t]
\footnotesize
\centering
\setlength\tabcolsep{1.0pt}
\def\arraystretch{1.1}
\resizebox{1.0\linewidth}{!}{
    \begin{tabular}{L{3.2cm}|C{2.6cm}|C{2.2cm}C{2.2cm}}
    \toprule
    \multirow{2}{*}{\diagbox[width=3.27cm]{\;~IR methods}{Dataset\;\;}}  &  \textbf{\textcolor{teal}{Seen}} & \multicolumn{2}{c}{\textbf{\textcolor{purple}{Unseen}}} \\
    & Tiny-ImageNet\textsuperscript{\textdagger}~\cite{le2015tiny} & CUB200~\cite{WahCUB_200_2011} & StanfordCars~\cite{krause20133d} \\
    \hline
    \;~Oracle & 83.7 & 82.5 & 90.4 \\
    \hline
    \;~\textit{No} restoration & 61.2 & 10.1 & 12.1 \\
    % \;~DiffBIR~\fakecite{27} & 62.6 & 41.0 & 57.7 \\
    % \;~URIE~\cite{son2020urie} & 40.1 & 20.1 & 19.4 \\
    \;~SwinIR~\cite{sr_swinir} & 68.0 & 21.8 & 38.7 \\
    \;~SR4IR~\cite{kim2024beyond} & 68.9 & 43.3 & 46.1 \\
    \;\cellcolor{cyan!\cyanP}~\textbf{EDTR}-1\,step \textbf{(Ours)} & \cellcolor{cyan!\cyanP}\underline{75.1} & \cellcolor{cyan!\cyanP}\underline{52.5} & \cellcolor{cyan!\cyanP}\textbf{67.0} \\
    \;~\cellcolor{cyan!\cyanP}\textbf{EDTR}-4\,step \textbf{(Ours)} & \cellcolor{cyan!\cyanP}\textbf{76.1} & \cellcolor{cyan!\cyanP}\textbf{53.0} & \cellcolor{cyan!\cyanP}\underline{66.4} \\
    \bottomrule
    \end{tabular}
}
\vspace{-3.0mm}
\caption{
    \textbf{Classification on seen and unseen datasets.}
    \vspace{-2mm}
}
\label{table:unseen}
\end{table}
\vspace{-1mm}

%% file: Sections/submission_arXiv/6_conclusion.tex
\section{Conclusion}
\vspace{-1mm}
This paper proposes EDTR, a novel task-driven image restoration~(TDIR) method that directly leverages a pre-trained diffusion prior to enhance high-level vision task performance.
We observe that while incorporating the diffusion prior produces visually pleasing results, restoring task-relevant details remains challenging.
To address this, we propose tailored designs for exploiting the diffusion prior in TDIR.
Our method significantly improves the performance of various tasks in complex image degradation scenarios while achieving visually appealing results, demonstrating its broad applicability and effectiveness.
We anticipate that our approach will provide new insights into leveraging diffusion priors to enhance task performance.

\noindent\textbf{Acknowledgments.}
This work was supported in part by the IITP grants [No.2021-0-01343, Artificial Intelligence Graduate School Program (Seoul National University), No. 2021-0-02068, and No.2023-0-00156], and the Industrial Technology Alchemist Project [No. RS-2024- 00432410] funded by MOTIE, Korea.

\iffalse
\noindent\textbf{Acknowledgments.}
This work was supported in part by the IITP grants [No. 2021-0-01343, Artificial Intelligence Graduate School Program (Seoul National University), No.2021-0-02068, and No.2023-0-00156], the NRF grant [No.2021M3A9E4080782] funded by the Korean governm-
\vspace{-5.5mm}
\flushleft\noindent ent (MSIT).
\endflushleft
\fi

%% file: Sections/supple_arXiv/Degradation_details.tex
\vspace{-1mm}
\section{Degradation details}
\label{sec:degradation}
\vspace{-1mm}
We provide the detailed image degradation settings used in our main experiments~(Table~\ref{table:main-table} of our main manuscript).
\iffalse
\begin{compactitem}[$\bullet$]
    % \item \textbf{Single:} Bilinear downsampling is applied with a scale factor of $\times8$.
    % 
    \item \textbf{Mixture-\textit{A}:} 
    % 
    Bilinear downsampling with a scale factor of $\times8$ is applied, followed by JPEG compression with a quality factor of 75.
    % 
    \item \textbf{Mixture-\textit{B}:}
    % 
    The Gaussian blur, bilinear downsampling, Gaussian noise, and JPEG compression are sequentially applied.
    % 
    Gaussian blur kernel size, downsampling ratio, Gaussian noise standard deviation, and JPEG quality factor are randomly selected within the ranges $[0.1, 8.0]$, $[1, 16]$, $[0, 10]$, and $[50, 100]$, respectively.
    % 
\end{compactitem}
\fi
\def\myvspace{0mm}
\begin{itemize}
    \item \textbf{Mixture-\textit{A}:} 
    Bilinear downsampling with a scale factor of $\times8$ is applied, followed by JPEG compression with a quality factor of 75.
    \vspace{\myvspace}
    \item \textbf{Mixture-\textit{B}:}
    The Gaussian blur, bilinear downsampling, Gaussian noise, and JPEG compression are sequentially applied.
    Gaussian blur kernel size, downsampling ratio, Gaussian noise standard deviation, and JPEG quality factor are randomly selected within the ranges $[0.1, 8.0]$, $[1, 16]$, $[0, 10]$, and $[50, 100]$, respectively.
    \vspace{\myvspace}
\end{itemize}
Note that after applying degradation, all degraded images are resized back to their original resolution using bilinear interpolation to match the size of high-quality images, following the CodeFormer~\cite{zhou2022towards}.

%% file: Sections/supple_arXiv/Training_details.tex
\vspace{-1mm}
\section{Training details}
\label{sec:training-details}
\vspace{-1mm}
The EDTR and high-level vision task networks are optimized over 10k iterations using the AdamW~\cite{misc_adamw} and SGD optimizers, respectively.
The learning rates are set to $10^{-4}$ for EDTR and $5\times10^{-3}$ for the task network, with a cosine annealing~\cite{misc_cosine_annealing} schedule.
For each degraded dataset, the pre-restoration network (\ie, SwinIR~\cite{sr_swinir}) is trained with the same number of iterations using the AdamW optimizer and a learning rate of $10^{-4}$.
The training batch size is 32 for image classification and 16 for other tasks.
The image resolution is fixed at $512\times512$ for all tasks.
If an image is smaller than $512\times512$, we pad it to meet the resolution and then crop it back to its original size.
We calculate the HLF loss using the output of the feature extractor for each task network, \eg, for image classification, the feature directly preceding the final fully connected layer in ResNet~\cite{cls_resnet}.
When computing the training loss for the task network in Equation~\eqref{eq:training-loss}, the balancing hyper-parameter $\alpha$ is set to 1 for classification, 0.5 for segmentation, and 0.2 for detection.
The task loss is defined as cross-entropy loss for classification and segmentation, and as a combination of classification, Region-of-Interest regression, objectness, and Region Proposal Network box regression losses for detection.

%% file: Sections/supple_arXiv/Impact_pre-restoration.tex
\vspace{-1mm}
\section{Impact of pre-restoration network}
\label{sec:impact-pixel-error}
\vspace{-1mm}
Table~\ref{table:impact-pre-restoration} presents the performance of EDTR-1\,step along with various pre-restoration~(\ie, pixel-error) networks.
We observe that a more powerful pre-restoration network, which achieves a higher pre-restoration PSNR, leads to better high-level vision task performance.
These results support our claim that removing degradation artifacts as much as possible before applying the diffusion prior is crucial for effectively harnessing the power of the diffusion prior in the TDIR, as discussed in Section~\ref{ssec:arch-edtr}.
In addition, this result also indicates that developing a classical IR model aimed at achieving higher PSNR is also relevant to the TDIR.
\vspace{-2mm}
\begin{table}[!h]
\footnotesize
\centering
\setlength\tabcolsep{1.0pt}
\def\arraystretch{1.1}
% \vspace{-3.0mm}
\resizebox{1.0\linewidth}{!}{
    \begin{tabular}{L{3.6cm}|C{1.8cm}|C{1.8cm}|C{1.8cm}}
    \toprule
    \,~Pre-restoration network & EDSR~\cite{sr_edsr} & RRDBNet~\cite{sr_esrgan} & SwinIR \textbf{(Used)} \\
    \hline
    \;~Acc$_\uparrow$ / Pre-restoration PSNR$_\uparrow$ & 66.8 / 24.45 & 67.3 / 24.59 & 68.8 / 25.04 \\
    \bottomrule
    \end{tabular}
}
\vspace{-3mm}
\caption{
    \textbf{Impact of pre-restoration network in classification.}
    % TODO.
    % 
}
\vspace{-3mm}
\label{table:impact-pre-restoration}
\end{table}

%% file: Sections/supple_arXiv/Training_algorithm.tex
\vspace{-1mm}
\section{Training algorithm}
\label{sec:training-algorithm}
\vspace{-1mm}
Algorithm~\ref{alg:edtr} presents the detailed training procedure for jointly training EDTR and task network~$\mathcal{H}$, which are introduced in Section~\ref{ssec:arch-edtr} and \ref{ssec:training-edtr-HT}.
Note that EDTR and task network are trained alternately, \ie, in one training iteration, EDTR is trained first, followed by an update to the $\mathcal{H}$.
\begin{algorithm*}[h!]
    \caption{Training algorithm for jointly training EDTR and the task network~$\mathcal{H}$}
    \label{alg:edtr}
    \renewcommand{\algorithmiccomment}[1]{\bgroup\hfill//~#1\egroup}
    \begin{algorithmic}[1]
    \REQUIRE
        Trainable parameter $\theta_\text{EDTR}$ for EDTR,
        $\theta_\mathcal{H}$ for task network $\mathcal{H}$,
        pre-restoration network $\mathcal{R}_\text{pix}$,
        denoising network combined with ControlNet $\epsilon_\theta$,
        VAE encoder $\mathcal{E}$ and decoder $\mathcal{D}$,
        set of HQ patches $\mathcal{I}_\text{HQ}$,
        image degradation model $\mathit{Deg}$,
        timestep for partial diffusion $t_p$,
        number of denoising steps $n$,
        variance schedule of Gaussian noise $\beta_t \in (0,1)$,
        high- and low-frequency wavelet components $\mathbf{H}$ and $\mathbf{L}$,
        total training iterations $N$,
        learning rates $\eta_\text{\,EDTR}$ and $\eta_\mathcal{\,H}$
    % \ENSURE
        % EDTR parameters $\theta_\text{EDTR}$ and task network parameters $\theta_\mathcal{H}$. \\
    \vspace{1mm}
    \STATE $q(z_t | z_{t-1}) \coloneqq\,\mathcal{N}(\sqrt{1-\beta_t}\,z_{t-1}, \beta_t\,\mathbf{I})$, $\alpha_t = 1 - \beta_t$, $\bar{\alpha}_t = \prod_{i=1}^{t} \alpha_{i}$, $\epsilon \sim \mathcal{N}(0,\mathbf{I})$  \\
    \STATE $\mathcal{T} = [t_p, \lfloor \frac{t_p \cdot (n-1)}{n} \rfloor,..., \lfloor \frac{t_p}{n} \rfloor]$ \COMMENT{Used $n$ timesteps for EDTR}\\
    \FOR {$i = 1:N$}
        \STATE $I_\text{HQ} \sim \mathcal{I}_\text{HQ}$, $I_\text{LQ} = \mathit{Deg} (I_\text{HQ})$ \COMMENT{Sample HQ images and generate LQ images} \\
        \STATE $z_\text{pre-res} = \mathcal{E}(\mathcal{R}_\text{pix}(I_\text{LQ}))$ \COMMENT{Pre-restoration and encoding} \\
        \STATE \textcolor{ForestGreen}{\texttt{\# Training EDTR}} \\
        \STATE $t \sim \text{Uniform}(\mathcal{T})$  \COMMENT{Sample timestep $t$}\\
        \STATE $z_{t, \text{partial}} =\, q(z_{t} | z_0 = z_\text{pre-res})$ \COMMENT{Partial diffusion with timestep $t$, Equation~\eqref{eq:partial-diffusion}}\\
        \STATE $z_{\text{diff-res}} = ({z_{t,\text{partial}} - \sqrt{1-\bar{\alpha}_{t}}{\epsilon_\theta}(z_{t,\text{partial}}, t, z_\text{pre-res})}) / {\sqrt{\bar{\alpha}_{t}}}$  \COMMENT{One-step denoising}\\
        \STATE $I_{\text{EDTR},\text{train}} = \mathbf{H}(\mathcal{D}(z_\text{diff-res})) + \mathbf{L}(\mathcal{R}_\text{pix}(I_\text{LQ}))$ \COMMENT{RGB image decoding with color correction, Equation~\eqref{eq:decoding}}\\
        \STATE $\theta_\text{EDTR} \leftarrow \theta_\text{EDTR} - \eta_{\,\text{EDTR}} \nabla_{\theta_\text{EDTR}} \mathcal{L}_\text{HLF}$ \COMMENT{Update $\theta_\text{EDTR}$ using HLF loss, Equation~\eqref{eq:HLF-loss}} \\
        \STATE \textcolor{ForestGreen}{\texttt{\# Training task network}} \\
        \STATE $z_{t_p, \text{partial}} = \, q(z_{t_p} | z_0 = z_\text{pre-res})$ \COMMENT{Partial diffusion with 
        timestep $t_p$, Equation~\eqref{eq:partial-diffusion}}\\
        % \STATE $z_{t_p, \text{partial}} = z_{\text{partial}}$ \COMMENT{Defined consistently with line~\ref{line1}}\\
        \STATE \texttt{\# $n$-step denoising, we set $n$ to a small value for short-step (\eg, 1, 4)} \\
        \FOR {$j = 0:(n-1)$}
            \STATE ${z}_{\text{diff-res}} = ({z_{\mathcal{T}[j], \text{partial}} - \sqrt{1-\bar{\alpha}_{\mathcal{T}[j]
            }}{\epsilon_\theta}(z_{\mathcal{T}[j], \text{partial}}, \mathcal{T}[j], z_\text{pre-res})}) / {\sqrt{\bar{\alpha}_{\mathcal{T}[j]}}}$ \COMMENT{One-step denoising} \\
            \IF {$j \neq (n-1) $}
                \STATE $z_{\mathcal{T}[j+1],\text{partial}} = q(z_{\mathcal{T}[j+1]}|z_{\mathcal{T}[j]}, z_0 = {z}_{\text{diff-res}})$ \COMMENT{Adding noise} \\
            \ENDIF
        \ENDFOR
        \STATE $I_{\text{EDTR}} = \mathbf{H}(\mathcal{D}(z_\text{diff-res})) + \mathbf{L}(\mathcal{R}_\text{pix}(I_\text{LQ}))$ \COMMENT{RGB image decoding with color correction, Equation~\eqref{eq:decoding}}\\
        \STATE $\theta_\mathcal{H} \leftarrow \theta_\mathcal{H} - \eta_{\,\mathcal{H}} \nabla_{\theta_\mathcal{H}} (\mathcal{L}_\text{task} + \alpha\,\mathcal{L}_\text{FM})$ \COMMENT{Update $\theta_\mathcal{H}$ using task loss and FM loss, Equations~\eqref{eq:task-loss} and \eqref{eq:feature-mathcing-loss}} \\
    \ENDFOR
    \end{algorithmic}
\end{algorithm*}

%% file: Sections/supple_arXiv/Complementary_HLF.tex
\section{Benefit of using two feature spaces in HLF}
\label{sec:complementary-HLF}
\vspace{-1mm}
We propose the HLF loss, which calculates the distance in the feature space of two task networks, effectively guiding the diffusion prior to restore task-relevant details.
To evaluate the effectiveness of using the feature space from both task networks, we compare the performance of EDTR trained with the loss for each single task network.
Specifically, we compare two cases: using only $\mathcal{H}^f$ and using only $\mathcal{H}_\text{HQ}^f$, in the original HLF loss (Equation~\eqref{eq:HLF-loss}).
Note that using only $\mathcal{H}^f$ is the same configuration (\ie, TDP loss) proposed in the previous TDIR method SR4IR~\cite{kim2024beyond}.

Table~\ref{table:ablation-study-HLF} presents the performance of EDTR under different training loss settings.
The EDTR model trained with our HLF loss achieves the best performance in both task accuracy and visual quality.
We claim that this is because the HLF loss extracts complementary information from both task networks, leading to improved performance compared to using a single task network.
\begin{table}[t!]
\footnotesize
\centering
\setlength\tabcolsep{1.0pt}
\def\arraystretch{1.1}
\resizebox{1.0\linewidth}{!}{
    \begin{tabular}{C{2.0cm}|C{1.2cm}|C{1.2cm}|C{1.4cm}C{1.4cm}}
    \toprule
    \,~\multirow{1}{*}{Training loss}\vspace{0.5mm} & \multirow{1}{*}{$\mathcal{H}^f$}\vspace{0.5mm} & \multirow{1}{*}{$\mathcal{H}_\text{HQ}^f$}\vspace{1mm} &  \multirow{1}{*}{Acc$_\uparrow$ (\%)}\vspace{1mm} & \multirow{1}{*}{Q-Align$_\uparrow$}\vspace{1mm} \\
    \hline
    \;~Only $\mathcal{H}^f$ & \cmark & \xmark & 67.8 & 3.36 \\
    \;~Only $\mathcal{H}_\text{HQ}^f$ & \xmark & \cmark & 68.0 & 3.26 \\
    \;~HLF~\textbf{(Ours)} & \cmark & \cmark & \textbf{68.8} & \textbf{3.48} \\
    \bottomrule
    \end{tabular}
}
\vspace{-2mm}
\caption{
    \textbf{EDTR performance on classification with different training losses.}
    The EDTR-1\,step model is used.
}
\vspace{-4mm}
\label{table:ablation-study-HLF}
\end{table}
\begin{figure}[t!]
    % \vspace{-3mm}
    \centering
    \subfloat[$\mathcal{H}^f$]{\includegraphics[width=0.48\linewidth]{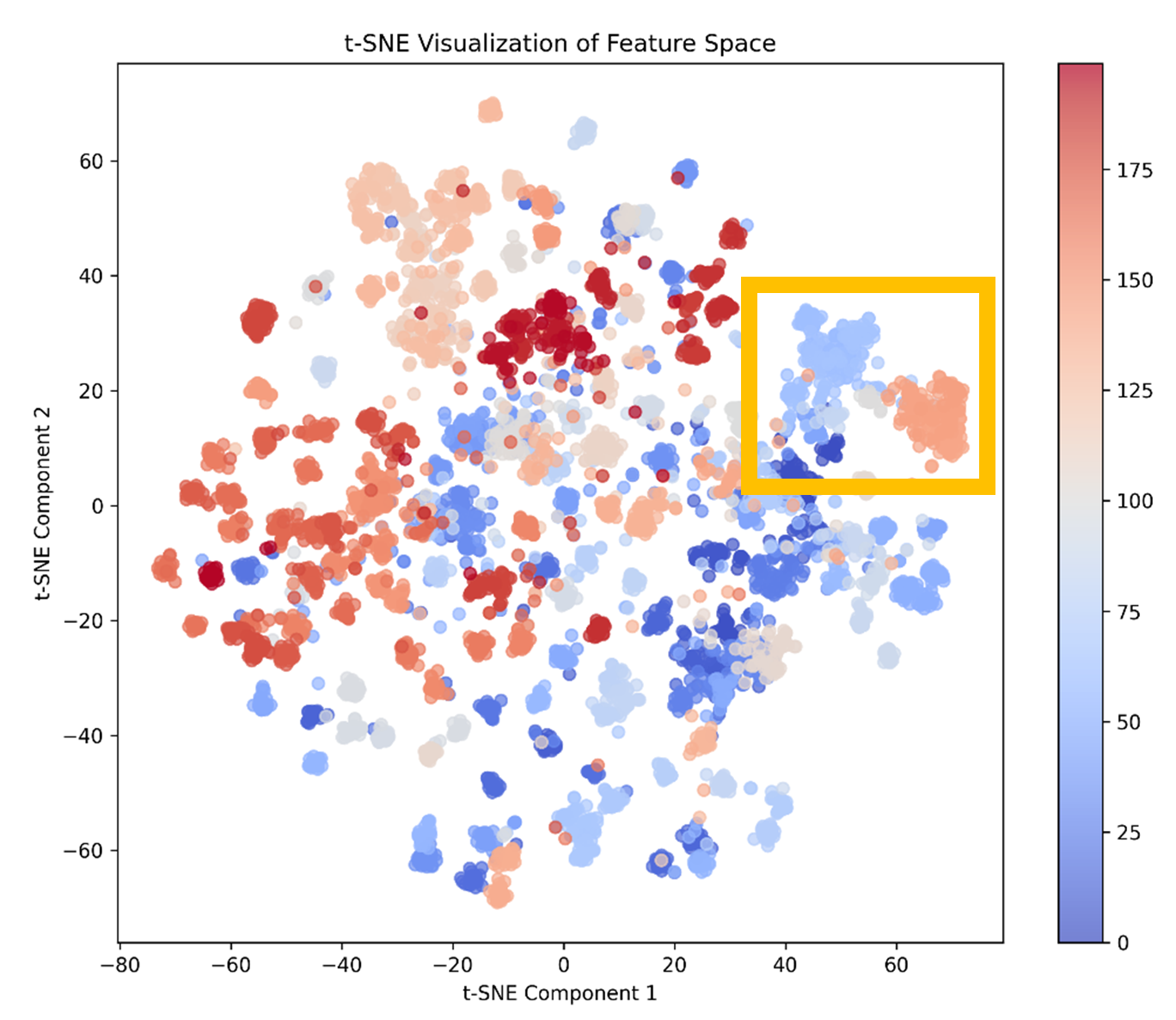}}
    \hfill
    \subfloat[$\mathcal{H}_\text{HQ}^f$]{\includegraphics[width=0.48\linewidth]{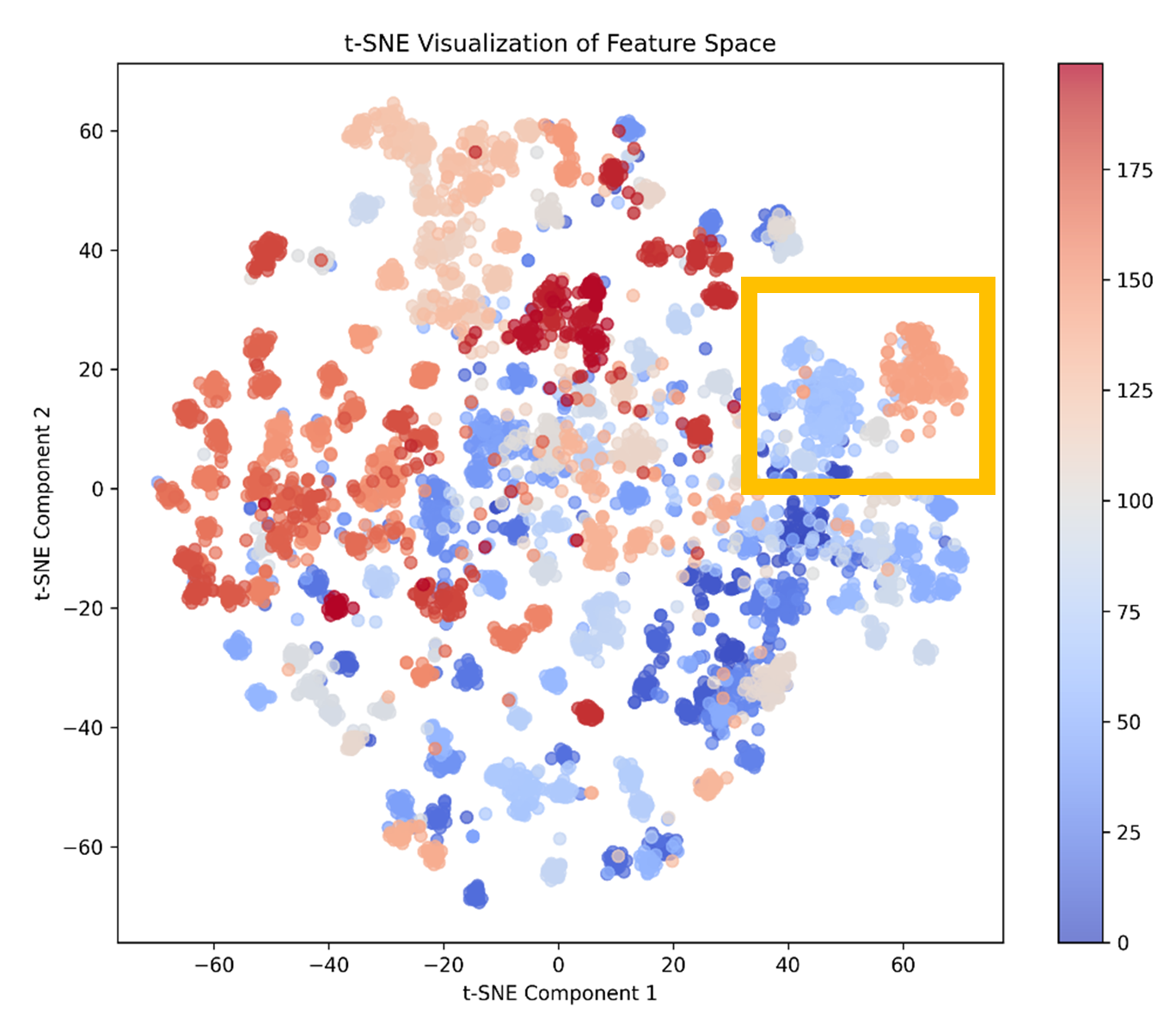}}
    \\
    \figspace
    \vspace{-1mm}
    \caption{
        \textbf{t-SNE visualizations of the feature space.}
        The feature spaces are from ResNet for classification.
        Each color of the point represents a classification label.
        % 
        % We highlight one of the different areas between the two feature spaces using a yellow box.
    }
    \label{fig:tsne}
    \figxspace
    \vspace{-2mm}
\end{figure}
Figure~\ref{fig:tsne} further validates our claim by visualizing the feature spaces of two task networks, $\mathcal{H}^f$ and $\mathcal{H}_\text{HQ}^f$.
Specifically, we use the t-SNE~\cite{JMLR:v9:vandermaaten08a} method for feature space visualization.
Although the two feature spaces exhibit notable similarity due to the feature matching loss, apparent differences exist between the two feature spaces.
For example, as highlighted in the yellow box in Figure~\ref{fig:tsne}, the relationship between the red and blue point groups differs in $\mathcal{H}^f$ and $\mathcal{H}_\text{HQ}^f$.
This suggests that incorporating two feature spaces in HLF loss can provide complementary information for the TDIR, leading to improved performance.

%% file: Sections/supple_arXiv/Computational_cost.tex
\vspace{-2mm}
\section{Computational cost of the EDTR}
\label{sec:computational-cost}
\vspace{-1mm}
Table~\ref{table:computational-cost} presents the computational cost of our EDTR.
The throughput and VRAM usage are measured on 512$\times$512 resolution images using a single NVIDIA RTX A6000 GPU and an Intel Xeon Gold 6226R CPU.
Notably, despite incorporating the large SD, EDTR achieves reasonable throughputs of 3.79 and 2.23 img/s for 1-step and 4-step settings, respectively, owing to its efficient short-step denoising.
\begin{table}[h!]
\footnotesize
\centering
\setlength\tabcolsep{1.0pt}
\def\arraystretch{1.1}
\vspace{-1mm}
\resizebox{1.0\linewidth}{!}{
    \begin{tabular}{L{2.8cm}|C{1.8cm}|C{3.0cm}|C{2.0cm}}
    \toprule
    \multirow{2}{*}{\;~IR methods} & Throughput & \# of parameters & VRAM usage \\
    & (img/s) & (B) & (MB) \\
    \hline
    \;~DiffBIR~\cite{wang2024exploiting} (50\,step) & 0.31 & 1.683 {(0.363 + \textcolor{gray}{1.320})} & 9310 \\
    \;~\textbf{EDTR}-1\,step \textbf{(Ours)} & 3.79 & 1.683 {(0.413 + \textcolor{gray}{1.270})} & 9310 \\
    \;~\textbf{EDTR}-4\,step \textbf{(Ours)} & 2.23 & 1.683 {(0.413 + \textcolor{gray}{1.270})} & 9310 \\
    \bottomrule
    \end{tabular}
}
\vspace{-2mm}
\caption{
    \textbf{Computational cost of the EDTR.}
    The \textcolor{gray}{gray} number represents the frozen parameters in SwinIR and SD.
    EDTR has more trainable parameters due to its trainable VAE decoder.
}
\label{table:computational-cost}
\end{table}
\vspace{-5mm}

%% file: Sections/supple_arXiv/Stochasticity.tex
\vspace{-2mm}
\section{Output stochasticity of the EDTR}
\label{sec:output-stochasticity}
\vspace{-1mm}
The outputs of EDTR exhibit stochasticity during inference due to the random noise~$\epsilon$ in the partial diffusion process, as described in Equation~\eqref{eq:partial-diffusion}.
Table~\ref{table:output-stochasticity} presents the average and standard deviation of the EDTR output metrics.
While the performance of high-level vision tasks shows some variability, image quality metrics remain highly consistent, with a standard deviation of less than 0.002.
Therefore, we report image quality metrics from a single inference while averaging the results of four inferences to ensure reliability when evaluating the task performance.
\vspace{-1mm}
\begin{table}[h!]
\footnotesize
\centering
\setlength\tabcolsep{1.0pt}
\def\arraystretch{1.1}
\resizebox{1.0\linewidth}{!}{
    \begin{tabular}{L{2.7cm}|C{1.5cm}|C{1.5cm}|C{1.5cm}|C{1.5cm}}
    \toprule
    \,~{Methods} & Acc$_\uparrow$ (\%) & NIQE$_\downarrow$ & Q-Align$_\uparrow$ & PSNR$_\uparrow$ \\
    \hline
    \;~\textbf{EDTR}-1\,step \textbf{(Ours)} & 68.8 / 0.249 & 4.75/ 0.002 & 3.48/ 0.002 & 23.03 / 0.001 \\
    \bottomrule
    \end{tabular}
}
\vspace{-2mm}
\caption{
    \textbf{Output stochasticity of EDTR-1\,step.}
    We report the average\,/\,standard deviation over 10 inference runs for each metric on the image classification task under Mixture-\textit{B} degradation.
}
\vspace{-0.3cm}
\label{table:output-stochasticity}
\end{table}

%% file: Sections/supple_arXiv/Comparison_detection.tex
\vspace{-2mm}
\section{Comparison with DiffBIR for detection}
\label{sec:comparison-detection}
\begin{figure}[t!]
    % \vspace{-2mm}
    \centering
    \subfloat[LQ~(\textit{No} restoration)]{\includegraphics[width=0.48\linewidth]{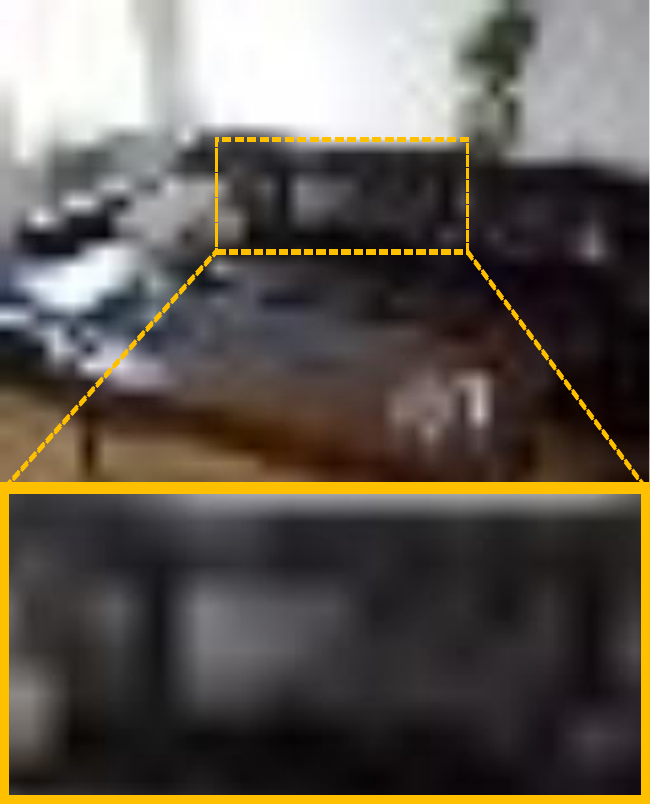}}
    \hfill
    \subfloat[DiffBIR~\cite{wang2024exploiting}]{\includegraphics[width=0.48\linewidth]{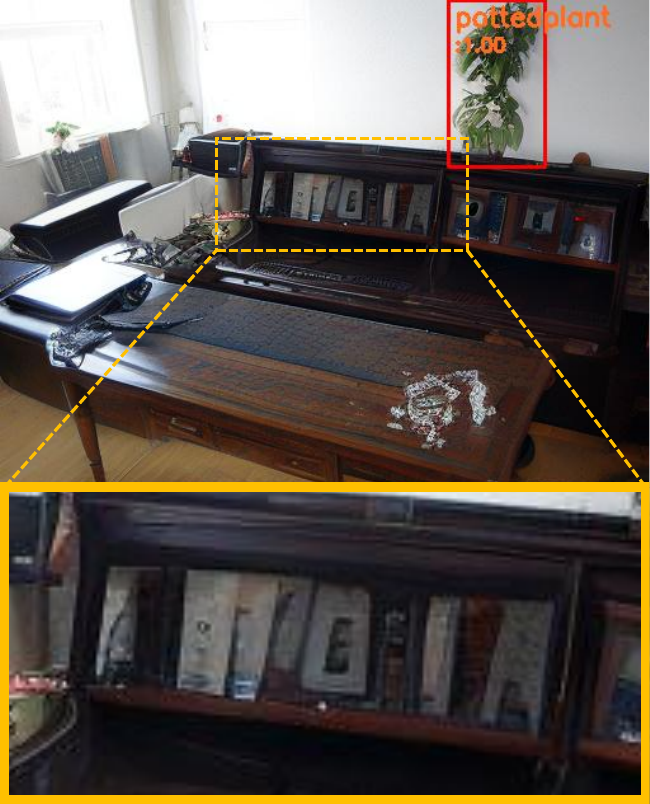}}
    \vspace{2mm}
    \\
    \subfloat[\textbf{EDTR (Ours)}]{\includegraphics[width=0.48\linewidth]{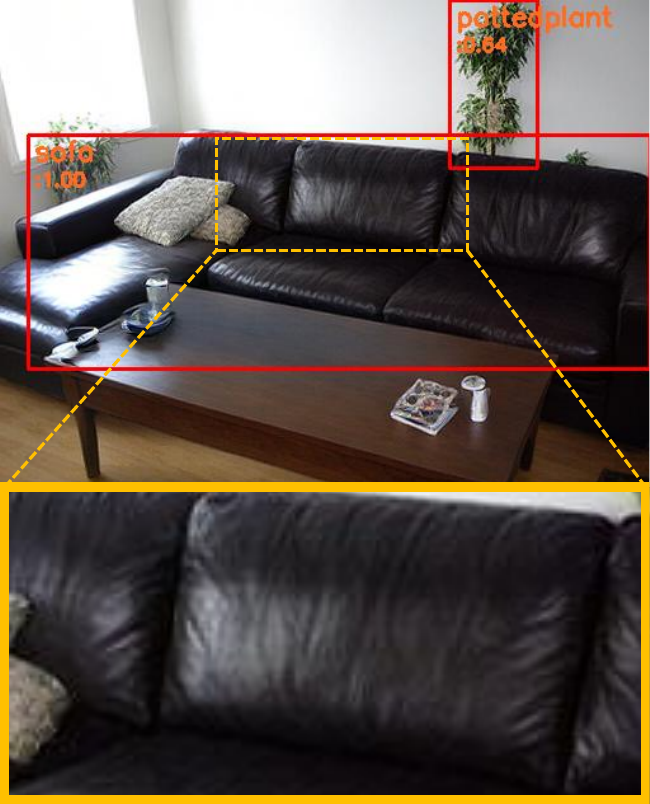}}
    \hfill
    \subfloat[HQ~(Oracle)]{\includegraphics[width=0.48\linewidth]{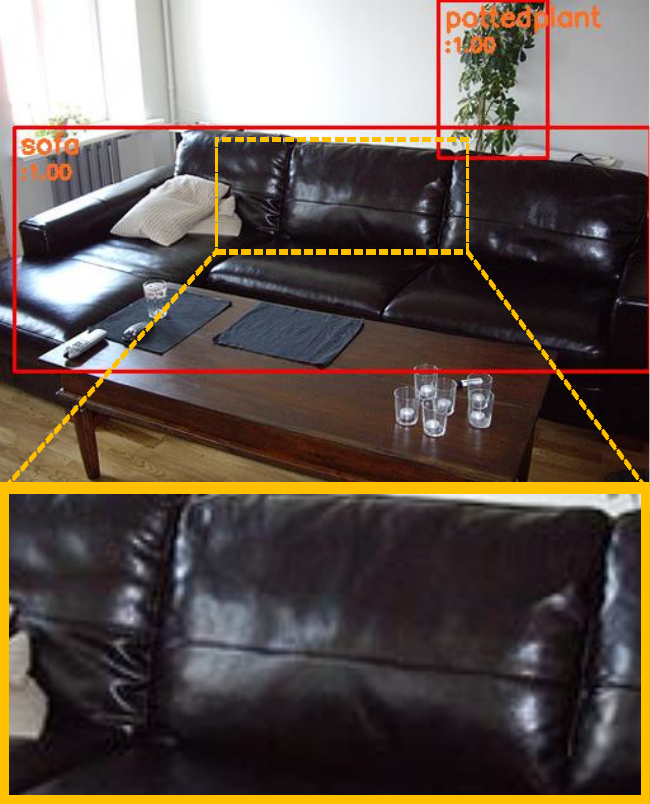}}
    \figspace
    \vspace{-1mm}
    \caption{
        \textbf{Comparison between EDTR-4\,step and conventional SD-based IR method on object detection.}
    }
    \label{fig:egp-detection}
    \figxspace
    \vspace{-2mm}
\end{figure}

\vspace{-1mm}
Table~\ref{table:egp-detection} compares EDTR with the conventional SD-based IR method (\ie, DiffBIR~\cite{wang2024exploiting}, Exp-(1) setting in Table~\ref{table:ablation-study} of our main manuscript) for object detection.
Figure~\ref{fig:egp-detection} illustrates that, although the conventional SD-based IR method restores the image in a visually pleasing manner, its generative prior can produce undesirable patterns, leading to misdetection results.
For example, the area highlighted in the yellow box in Figure~\ref{fig:egp-detection} is part of a sofa, but the conventional SD-based IR method restores it to resemble a bookshelf, resulting in a failure to detect the sofa.
These results further validate that the \textit{diffusion prior must be carefully managed} to effectively restore task-relevant details.
\begin{table}[t!]
\footnotesize
\centering
\setlength\tabcolsep{1.0pt}
\def\arraystretch{1.1}
\resizebox{1.0\linewidth}{!}{
    \begin{tabular}{L{5.2cm}|C{1.4cm}|C{1.2cm}}
    \toprule
    \,~{Methods} & mAP$_\uparrow$ (\%) & Q-Align$_\uparrow$ \\
    \hline
    \;~Conventional SD-based IR method~\cite{wang2024exploiting} & 25.9 & \underline{3.87} \\
    \;~\textbf{EDTR}-1\,step \textbf{{(Ours)}} & \underline{30.6} & 3.64 \\
    \;~\textbf{EDTR}-4\,step \textbf{{(Ours)}} & \textbf{31.9} & \textbf{4.02} \\
    \bottomrule
    \end{tabular}
}
\vspace{-3mm}
\caption{
    \textbf{Performance comparison for detection (Mixture-\textit{B}).}
}
\vspace{-0.2cm}
\label{table:egp-detection}
\end{table}
% 
% \begin{table}[t!]
% \footnotesize
% \centering
% \setlength\tabcolsep{1.0pt}
% \def\arraystretch{1.1}
% \resizebox{1.0\linewidth}{!}{
%     \begin{tabular}{L{4.8cm}|C{1.3cm}|C{1.2cm}}
%     \toprule
%     \,~{Methods} & mAP$_\uparrow$ (\%) & $f_d$$_\downarrow$ & Q-Align$_\uparrow$ \\
%     \hline
%     \;~Conventional SD-based IR method~\fakecite{27} & 25.9 & 0.971 & \textbf{3.87} \\
%     \;~\textbf{EDTR (proposed)} & \textbf{30.6} & \textbf{0.790}  & 3.64 \\
%     \bottomrule
%     \end{tabular}
% }
% \vspace{-3mm}
% \caption{
%     \textbf{Performance comparison with conventional SD-based IR method on object detection.}
%     % 
% }
% \vspace{-0.2cm}
% \label{table:egp-detection}
% \end{table}
% % 

%% file: Sections/supple_arXiv/SR4IR-combined-with-SD.tex
% \vspace{2mm}
\section{SR4IR combined with SD}
\label{sec:sr4ir-combined-with-sd}
Table~\ref{table:sr4ir-combined-with-sd} presents the results of directly incorporating SD into the previous state-of-the-art TDIR method, SR4IR~\cite{kim2024beyond}, with the results visualized in Figure~\ref{fig:main_figure_c}.
Note that this approach differs from Exp-(2) in Table~\ref{table:ablation-study}, including the use of FM loss and other techniques (\eg, TDP and CQMix, which were introduced in SR4IR).
Despite the incorporation of a strong diffusion prior, it performs even worse than the original SR4IR.
%
% In contrast, our EDTR successfully harnesses the diffusion prior to enhance task performance.
%
These results further highlight that even with a strong diffusion prior, effectively handling it to restore task-relevant details is a crucial challenge.
\vspace{-2mm}
\begin{table}[h!]
\footnotesize
\centering
\setlength\tabcolsep{1.0pt}
\def\arraystretch{1.1}
\resizebox{1.0\linewidth}{!}{
    \begin{tabular}{L{3.8cm}|C{1.1cm}|C{1.1cm}|C{1.1cm}|C{1.1cm}}
    \toprule
    \,~{Methods} & Acc$_\uparrow$ (\%) & NIQE$_\downarrow$ & Q-Align$_\uparrow$ & PSNR$_\uparrow$ \\
    \hline
    \;~SR4IR~\cite{kim2024beyond} & 63.4 & 6.08 & 3.11 & 23.62 \\
    \;~SR4IR~\cite{kim2024beyond} + SD~\cite{rombach2022high} & 57.6 & 6.01 & 2.35 & 16.92 \\
    \;~\textbf{EDTR}-1\,step \textbf{(Ours)} & 68.8 & 4.75 & 3.48 & 23.03 \\
    \bottomrule
    \end{tabular}
}
\vspace{-2mm}
\caption{
    \textbf{Performance of the SR4IR combined with SD.}
}
\vspace{-5mm}
\label{table:sr4ir-combined-with-sd}
\end{table}

%% file: Sections/supple_arXiv/Details_for_previous.tex
\vspace{1mm}
\section{Details for the previous works}
\label{sec:details-previous}
% 
% \vspace{-2mm}
\paragraph{TDSR.} 
Since the official code for TDSR~\cite{sr_tdsr} is not available, we re-implemented it ourselves.
We use TDSR-0.01, which employs a pixel loss ratio of 1.0 and a task loss ratio of 0.01.
Unlike the original TDSR, which is limited to object detection, we extend it to image classification and semantic segmentation, utilizing the respective high-level vision task losses, \eg, cross-entropy loss for image classification.
We adopt SwinIR~\cite{sr_swinir} as the restoration model.

\vspace{-4mm}
\paragraph{RSRSSN.} 
Since the official code for RSRSSN~\cite{zhao2018residual} is not available, we re-implemented it ourselves.
Following the paper, we train the restoration model and task network in an end-to-end manner using feature map multi-box loss and task loss, with respective weight ratios of 0.1 and 1.0.
As with our re-implementation of TDSR, we extend RSRSSN to include image classification and semantic segmentation.
We additionally incorporate pixel loss during training, as we observed significantly degraded image quality without it.
We use SwinIR as the restoration model.

\vspace{-4mm}
\paragraph{SR4IR.} 
We use the official code from SR4IR~\cite{kim2024beyond} to obtain the results.
SwinIR is used as the restoration model.

%% file: Sections/supple_arXiv/Additional_ablations.tex
\section{Additional ablation studies}
\label{sec:additional-ablation}
\vspace{-2mm}
\begin{table}[h!]
\footnotesize
\centering
\setlength\tabcolsep{1.0pt}
\def\arraystretch{1.1}
\resizebox{1.0\linewidth}{!}{
    \begin{tabular}{C{0.9cm}|L{3.8cm}|C{1.1cm}|C{1.1cm}|C{1.1cm}|C{1.1cm}}
    \toprule
    {Exp} & \,~{Methods} & Acc$_\uparrow$ (\%) & NIQE$_\downarrow$ & Q-Align$_\uparrow$ & PSNR$_\uparrow$ \\
    \hline
    (1) & \;~\textbf{EDTR} \textbf{(Ours)} & 68.8 & 4.75 & 3.48 & 23.03 \\
    % \;~(2) & \;~Using only $\mathcal{H}_\text{p}$ for $\mathcal{L}_\text{HLF}$ & 68.0 & 5.06 & 3.26 & 23.00 \\
    (2) & \;~\textit{without} trainable $\mathcal{D}$ & 67.7 & 6.03 & 3.55 & 23.61 \\
    (3) & \;~\textit{without} color correction & 68.4 & 4.89 & 3.56 & 22.84 \\
    % \;~(5) & \;~\textit{without} using $I_\text{HQ}$ in $\mathcal{L}_\text{task}$,$\mathcal{L}_\text{FM}$ & 68.5 & 5.10 & 3.41 & 23.02 \\
    % \;~(4) & \;~Using a text prompt & 68.8 & 4.73 & 3.43 & 23.03 \\
    \bottomrule
    \end{tabular}
}
\vspace{-2mm}
\caption{
    \textbf{Performance of EDTR-1\,step under various settings.}
}
\vspace{-5mm}
\label{table:performance-under-various}
\end{table}

\vspace{-2mm}
\paragraph{Trainable decoder.}
\label{abl:text-prompt}
The Exp-(2) result in Table~\ref{table:performance-under-various} shows the EDTR performance without a trainable VAE decoder.
Although EDTR without a trainable decoder obtains a higher PSNR amount of 0.58 dB, it scores 1.1\% lower in classification accuracy and shows a worse NIQE score; therefore, we opt for the VAE decoder to be trainable.

\vspace{-2mm}
\paragraph{Wavelet color correction.}
\label{abl:wavelet-corrrection}
The Exp-(3) result in Table~\ref{table:performance-under-various} shows the EDTR performance without Wavelet color correction, as introduced in Equation~\eqref{eq:decoding} of the main manuscript.
EDTR without Wavelet color correction achieves a slightly lower classification accuracy ($-$0.4\%) and a PSNR drop of 0.19 dB.
Therefore, we opt to include Wavelet color correction in our EDTR.
% 

% \vspace{-2mm}
% \paragraph{Using $I_\text{HQ}$ for training the task network.}
% \label{abl:using-HQ-image-task}
% % 
% The Exp-(5) result from Table~\ref{table:performance-under-various} shows the EDTR performance without using HQ images to train the task network, as described in Equations~\fakeref{6} and~\fakeref{7} of our main manuscript.
% EDTR without HQ images scores 0.3\% lower in classification accuracy and exhibits instability early in training.
% % 
% Therefore, we include $I_\text{HQ}$ when training the task network.
% % 
% 

% \vspace{-2mm}
% \paragraph{Effect of text prompt.}
% \label{abl:text-prompt}
% % 
% The Exp-(4) result from Table~\ref{table:performance-under-various} presents the EDTR performance with a text prompt.
% % 
% Specifically, we use a text description extracted from DAPE~\fakecite{54} as additional conditional input to the denoising network of StableDiffusion, in contrast to the original EDTR, which uses no text prompt.
% % 
% The results indicate that employing text descriptions does not improve classification accuracy or image quality; therefore, we opt not to use them.
% % 

%% file: Sections/supple_arXiv/Further_visualization.tex
\section{Further visualization results}
\label{sec:further-visualization}
\begin{figure*}[h!]
    \centering
    \captionsetup[subfigure]{labelfont=scriptsize, textfont=scriptsize}
    \renewcommand{\wp}{0.247}
        \subfloat[LQ~(\textit{No} restoration)]{\includegraphics[width=\wp\linewidth]{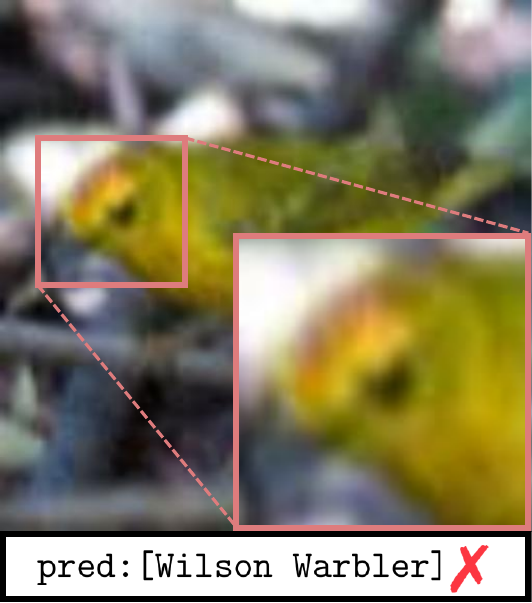}}
        \hfill
        \subfloat[SwinIR~\cite{sr_swinir}]{\includegraphics[width=\wp\linewidth]{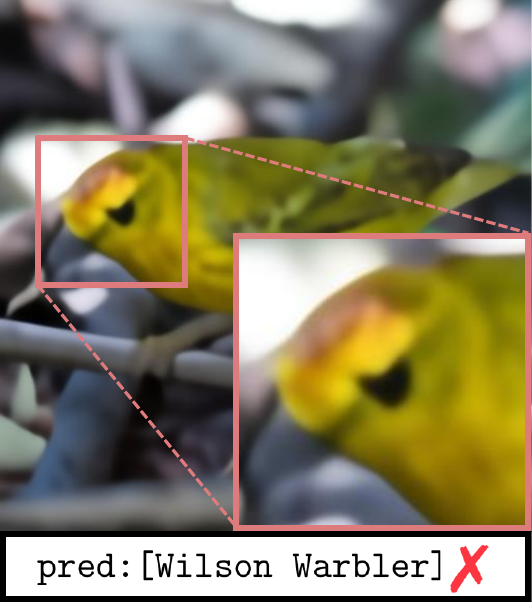}}
        \hfill
        \subfloat[TDSR~\cite{sr_tdsr}]{\includegraphics[width=\wp\linewidth]{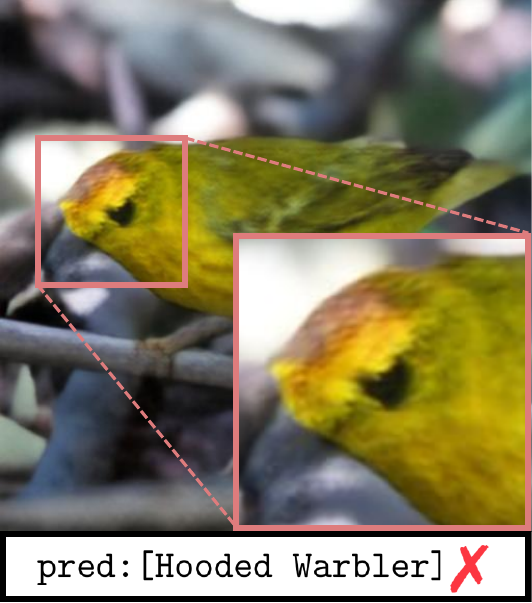}}
        \hfill
        \subfloat[RSRSSN~\cite{zhao2018residual}]{\includegraphics[width=\wp\linewidth]{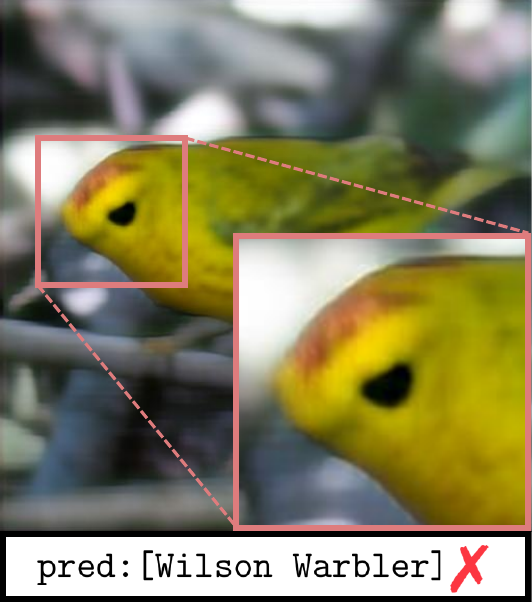}}
        \vspace{1mm}
        \\
        \subfloat[SR4IR~\cite{kim2024beyond}]{\includegraphics[width=\wp\linewidth]{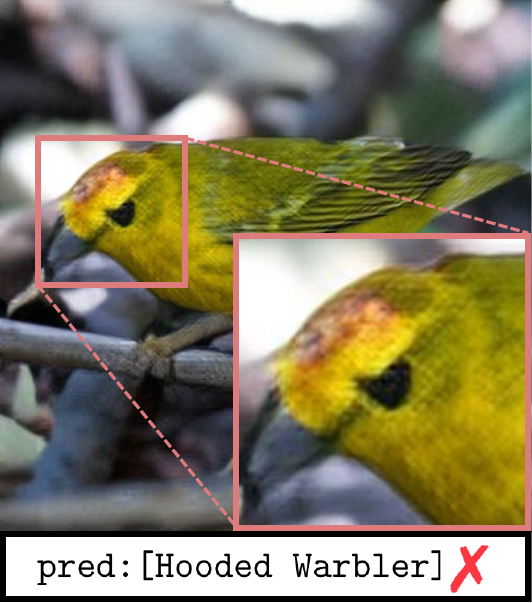}}
        \hfill
        \subfloat[\textbf{EDTR~(Ours)}]{\includegraphics[width=\wp\linewidth]{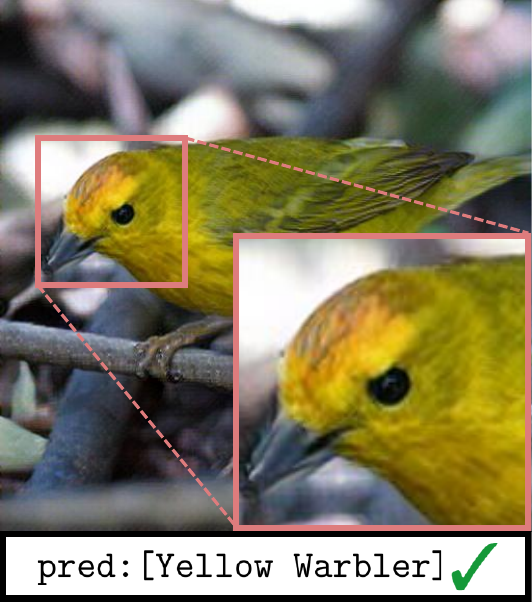}}
        \hfill
        \subfloat[HQ~(Oracle)]{\includegraphics[width=\wp\linewidth]{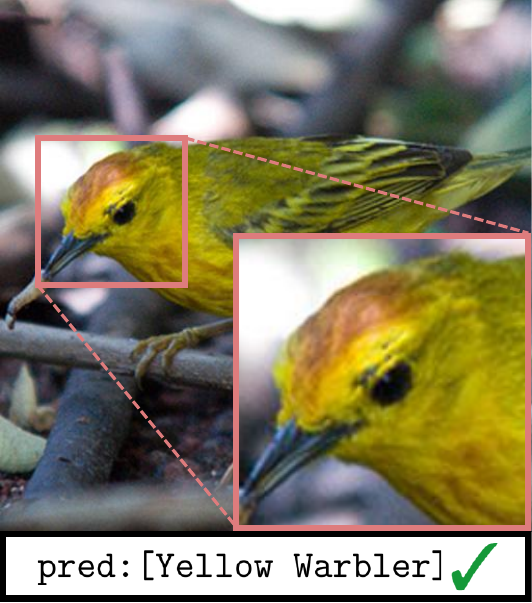}}
        \hfill
        \subfloat[HQ~(Ground-truth)]{\includegraphics[width=\wp\linewidth]{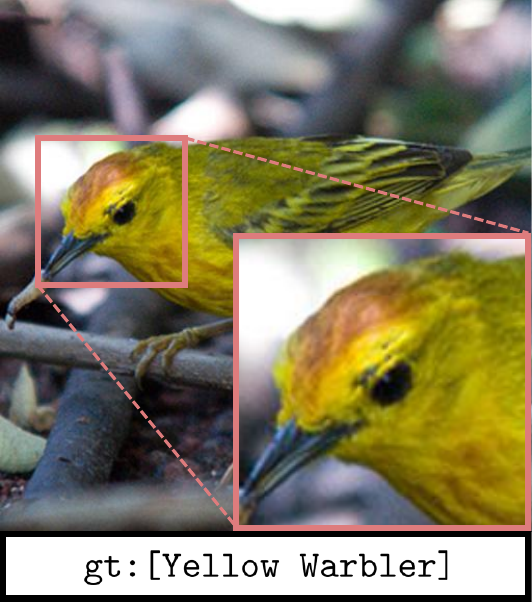}}
        \addtocounter{subfigure}{-8}
        \\
        \vspace{3mm}
        \begin{tikzpicture}
            \draw[dashed] (0,0) -- (17.2,0);
        \end{tikzpicture}
        \vspace{4mm}
        \\
        \subfloat[LQ~(\textit{No} restoration)]{\includegraphics[width=\wp\linewidth]{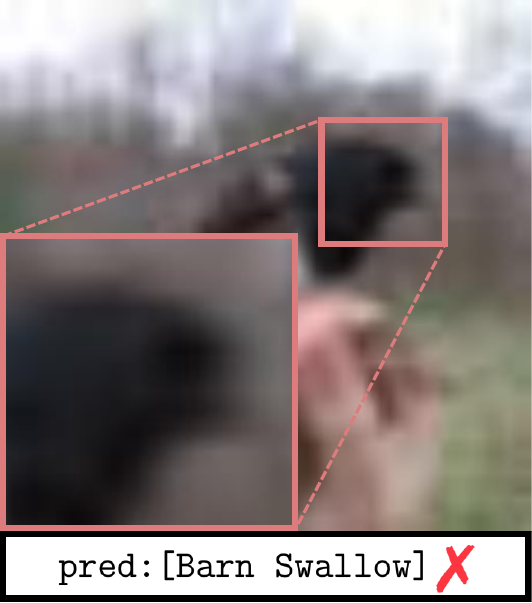}}
        \hfill
        \subfloat[SwinIR~\cite{sr_swinir}]{\includegraphics[width=\wp\linewidth]{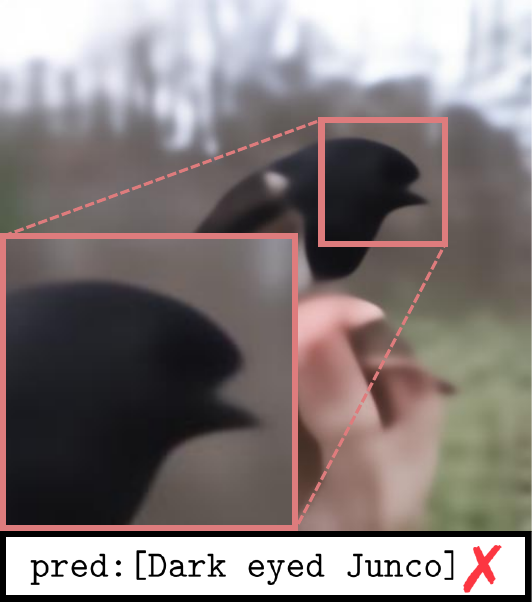}}
        \hfill
        \subfloat[TDSR~\cite{sr_tdsr}]{\includegraphics[width=\wp\linewidth]{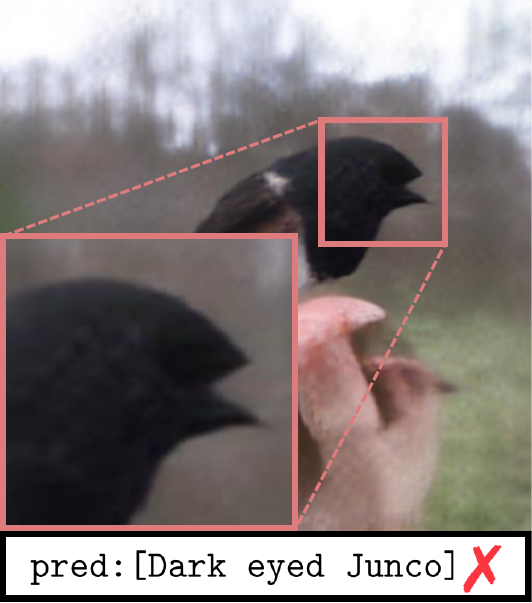}}
        \hfill
        \subfloat[RSRSSN~\cite{zhao2018residual}]{\includegraphics[width=\wp\linewidth]{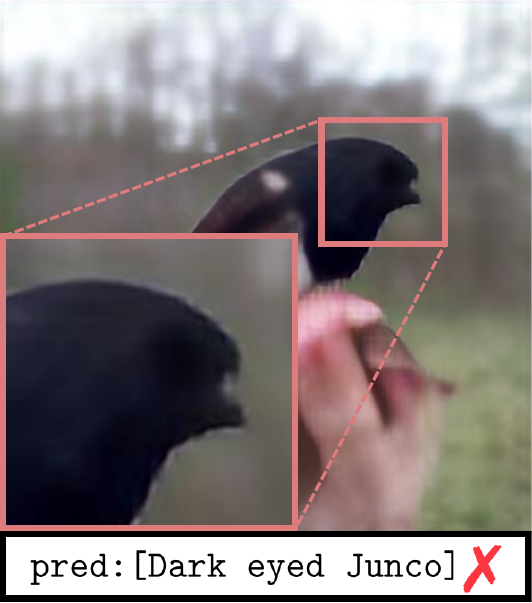}}
        \vspace{1mm}
        \\
        \subfloat[SR4IR~\cite{kim2024beyond}]{\includegraphics[width=\wp\linewidth]{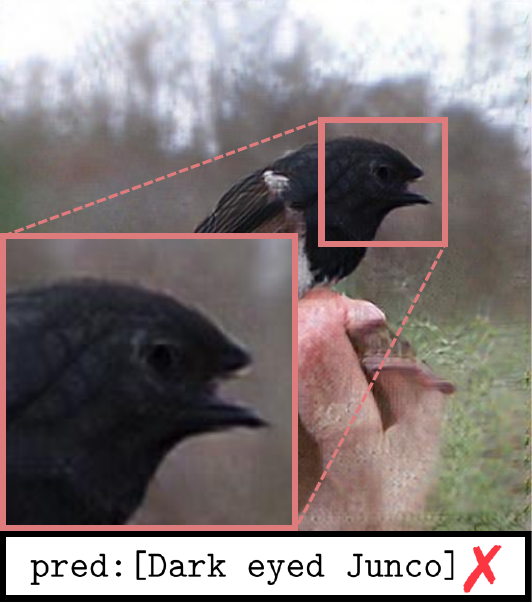}}
        \hfill
        \subfloat[\textbf{EDTR~(Ours)}]{\includegraphics[width=\wp\linewidth]{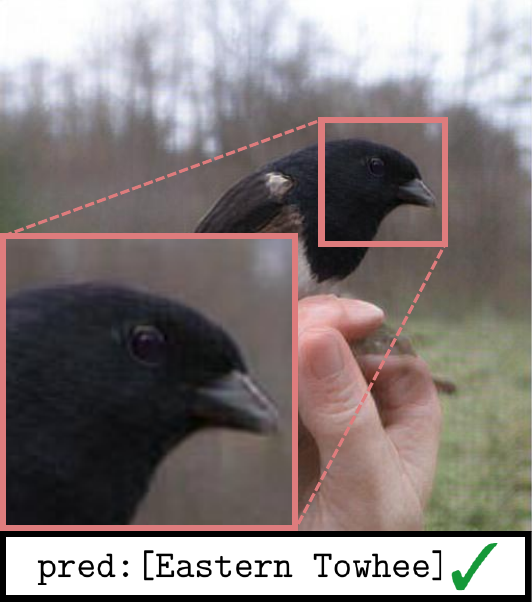}}
        \hfill
        \subfloat[HQ~(Oracle)]{\includegraphics[width=\wp\linewidth]{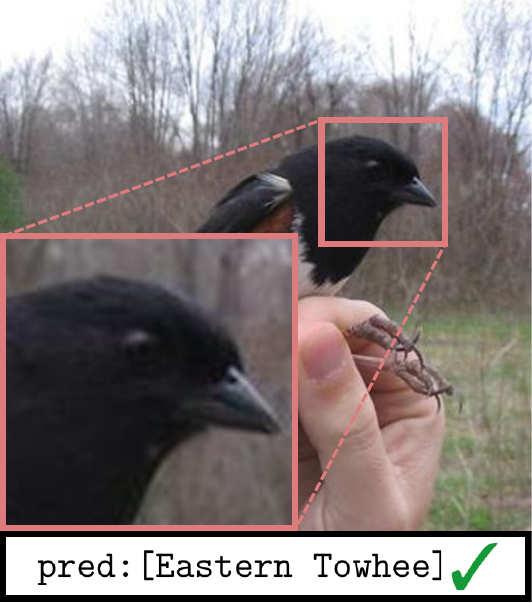}}
        \hfill
        \subfloat[HQ~(Ground-truth)]{\includegraphics[width=\wp\linewidth]{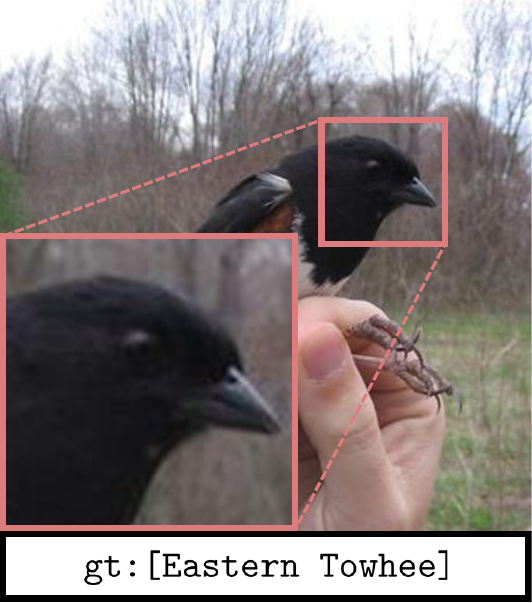}}
    \vspace{-0.1cm}
    \caption{\textbf{Further visualization of images and image classification results on degraded LQ (Mixture-\textit{B}) images.}
        We show the restored images and the corresponding predicted or ground-truth labels.
        The EDTR-1\,step model is used for visualization.
    }
    \label{fig:further-vis-cls}
\end{figure*}

\begin{figure*}[h!]
    \centering
    \captionsetup[subfigure]{labelfont=scriptsize, textfont=scriptsize}
    \renewcommand{\wp}{0.247}
        \subfloat[LQ~(\textit{No} restoration)]{\includegraphics[width=\wp\linewidth]{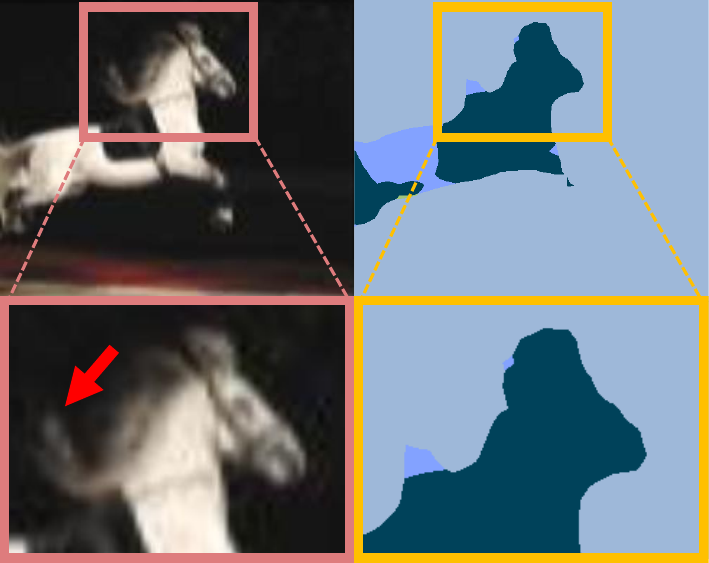}}
        \hfill
        \subfloat[SwinIR~\cite{sr_swinir}]{\includegraphics[width=\wp\linewidth]{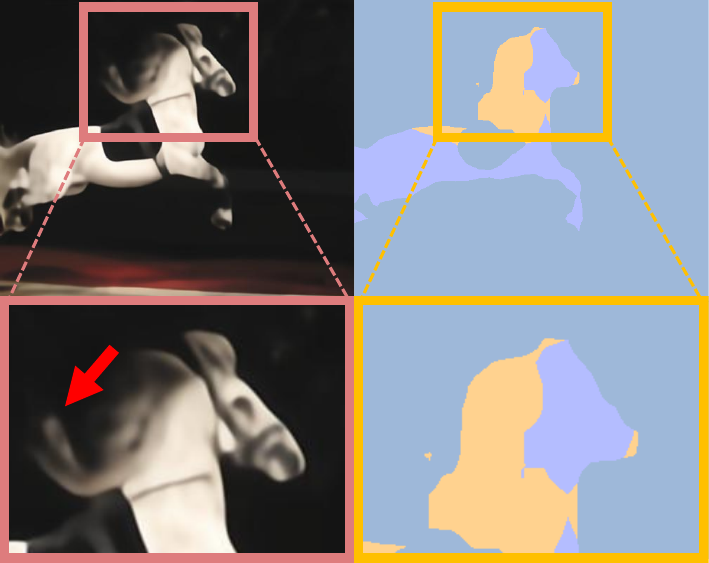}}
        \hfill
        \subfloat[TDSR~\cite{sr_tdsr}]{\includegraphics[width=\wp\linewidth]{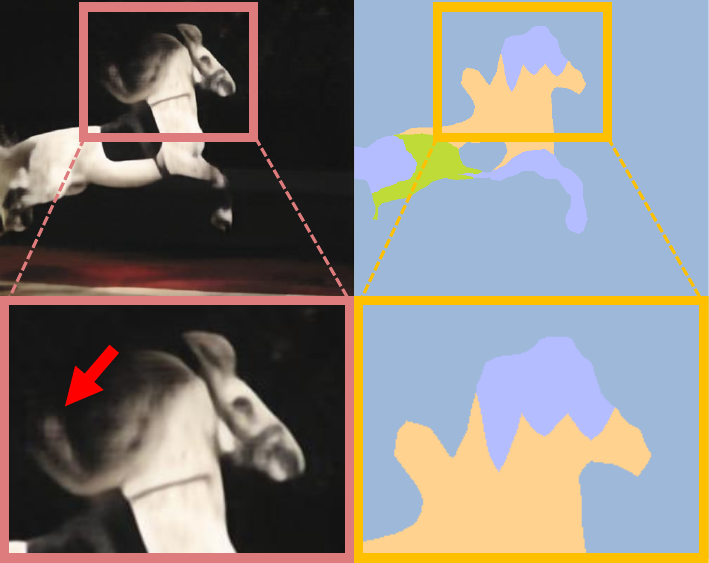}}
        \hfill
        \subfloat[RSRSSN~\cite{zhao2018residual}]{\includegraphics[width=\wp\linewidth]{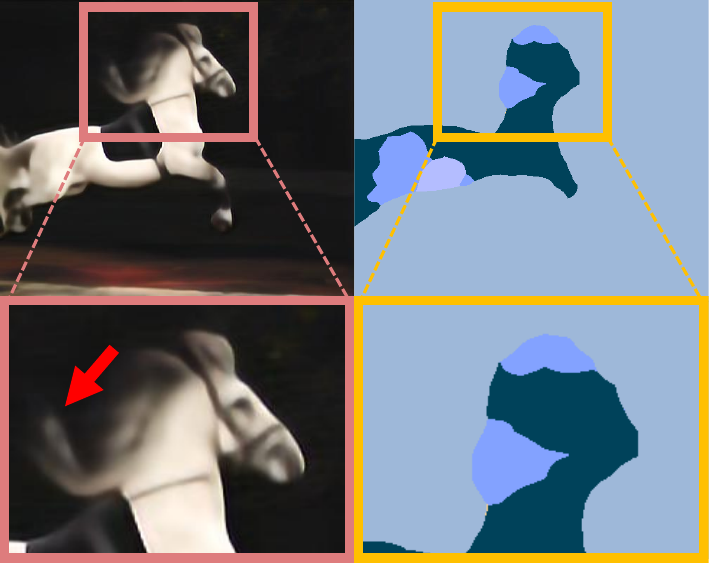}}
        \vspace{2mm}
        \\
        \subfloat[SR4IR~\cite{kim2024beyond}]{\includegraphics[width=\wp\linewidth]{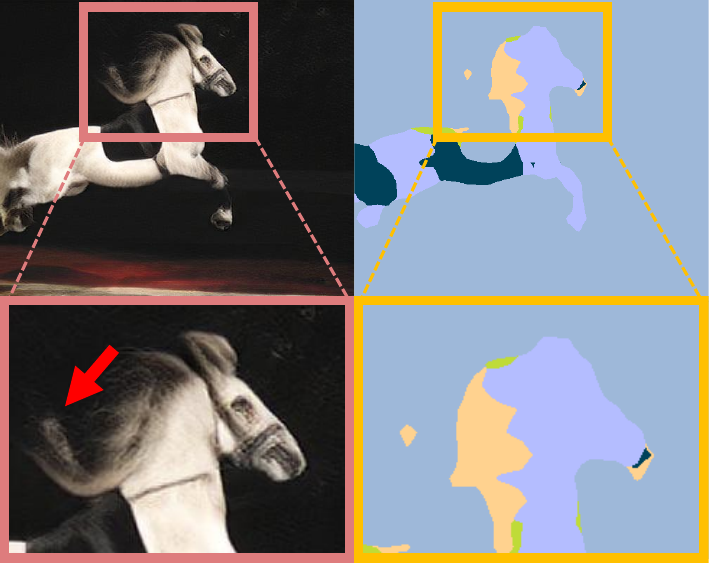}}
        \hfill
        \subfloat[\textbf{EDTR~(Ours)}]{\includegraphics[width=\wp\linewidth]{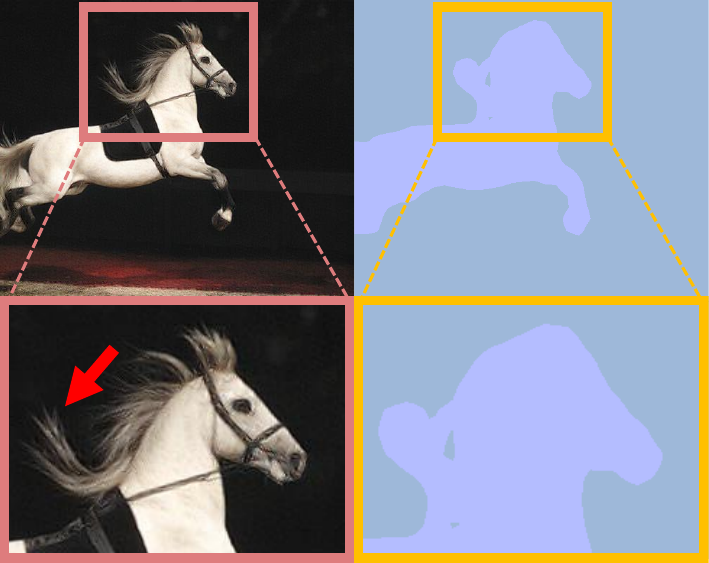}}
        \hfill
        \subfloat[HQ~(Oracle)]{\includegraphics[width=\wp\linewidth]{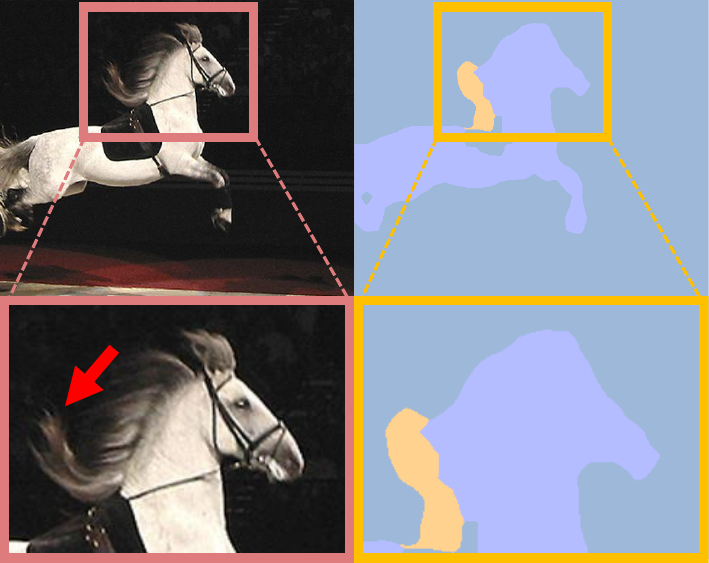}}
        \hfill
        \subfloat[HQ~(Ground-truth)]{\includegraphics[width=\wp\linewidth]{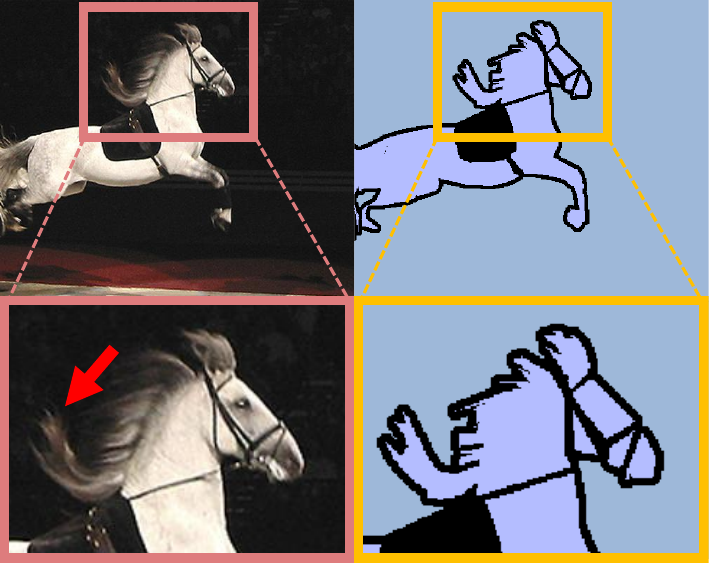}}
        \addtocounter{subfigure}{-8}
        \\
        \vspace{5mm}
        \begin{tikzpicture}
            \draw[dashed] (0,0) -- (17.2,0);
        \end{tikzpicture}
        \vspace{6mm}
        \\
        \subfloat[LQ~(\textit{No} restoration)]{\includegraphics[width=\wp\linewidth]{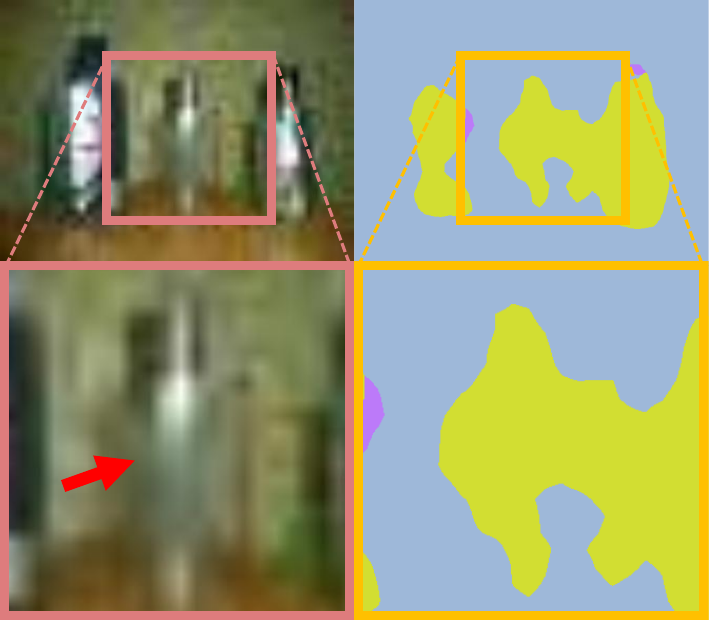}}
        \hfill
        \subfloat[SwinIR~\cite{sr_swinir}]{\includegraphics[width=\wp\linewidth]{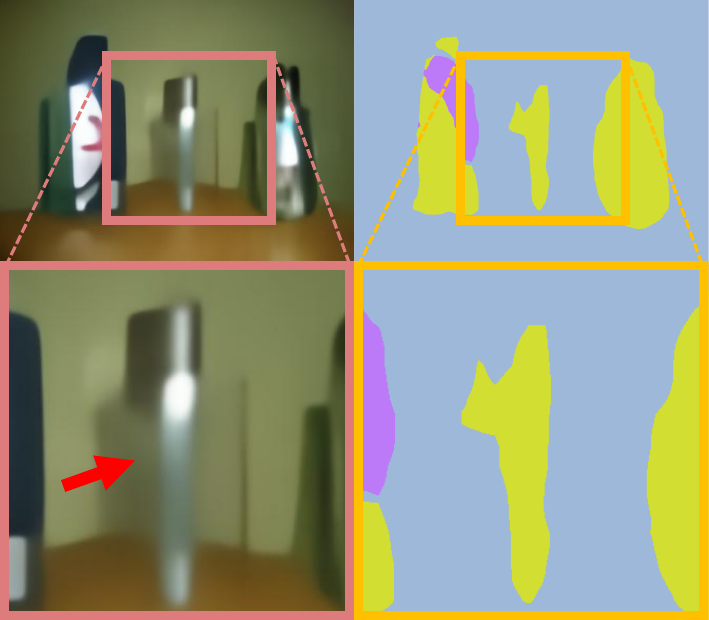}}
        \hfill
        \subfloat[TDSR~\cite{sr_tdsr}]{\includegraphics[width=\wp\linewidth]{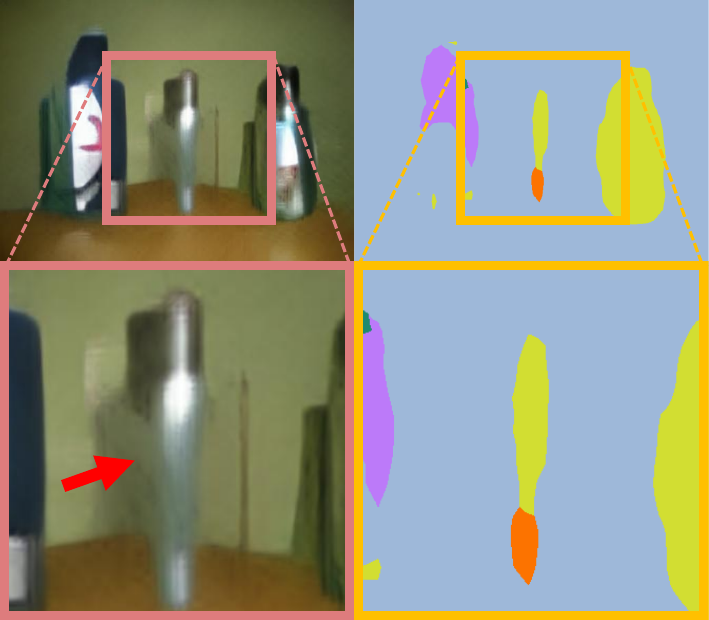}}
        \hfill
        \subfloat[RSRSSN~\cite{zhao2018residual}]{\includegraphics[width=\wp\linewidth]{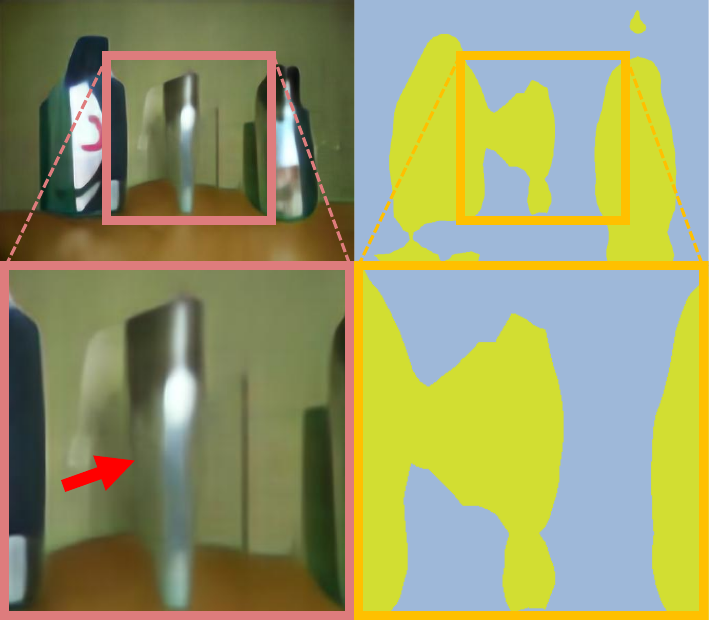}}
        \vspace{2mm}
        \\
        \subfloat[SR4IR~\cite{kim2024beyond}]{\includegraphics[width=\wp\linewidth]{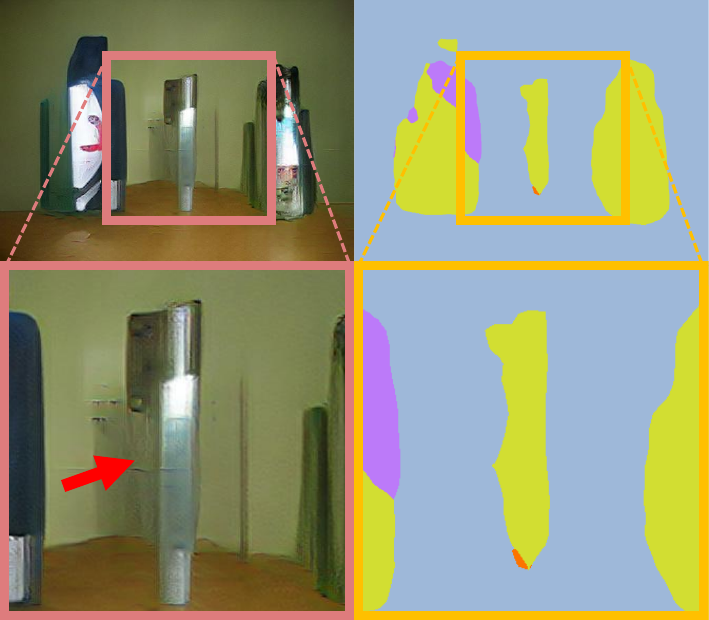}}
        \hfill
        \subfloat[\textbf{EDTR~(Ours)}]{\includegraphics[width=\wp\linewidth]{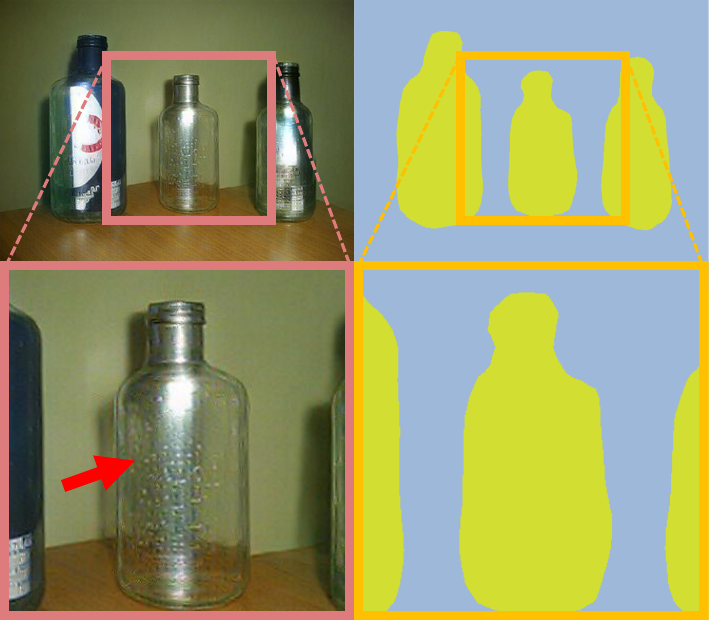}}
        \hfill
        \subfloat[HQ~(Oracle)]{\includegraphics[width=\wp\linewidth]{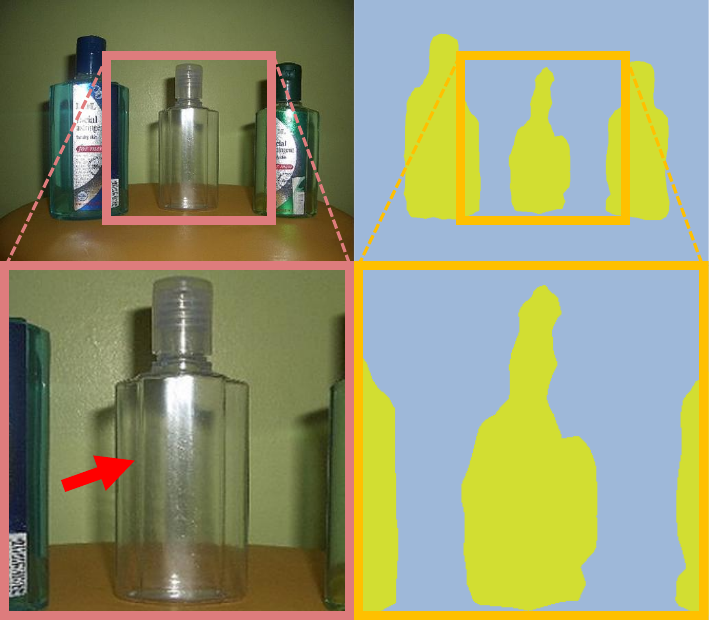}}
        \hfill
        \subfloat[HQ~(Ground-truth)]{\includegraphics[width=\wp\linewidth]{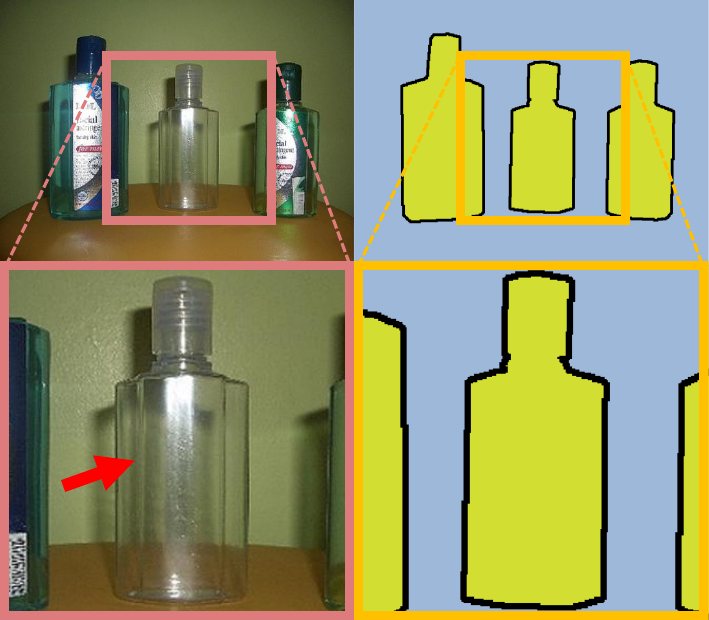}}

    \vspace{-0.1cm}
    \caption{\textbf{Further visualization of images and semantic segmentation results on degraded LQ (Mixture-\textit{B}) images.}
        We show the restored images and the corresponding predicted or ground-truth labels.
        The black line in the ground-truth segmentation map indicates "don't care" regions.
        The EDTR-4\,step model is used for visualization.
    }
    \label{fig:further-vis-seg}
\end{figure*}

\begin{figure*}[h!]
    \centering
    \captionsetup[subfigure]{labelfont=scriptsize, textfont=scriptsize}
    \renewcommand{\wp}{0.247}
        \subfloat[LQ~(\textit{No} restoration)]{\includegraphics[width=\wp\linewidth]{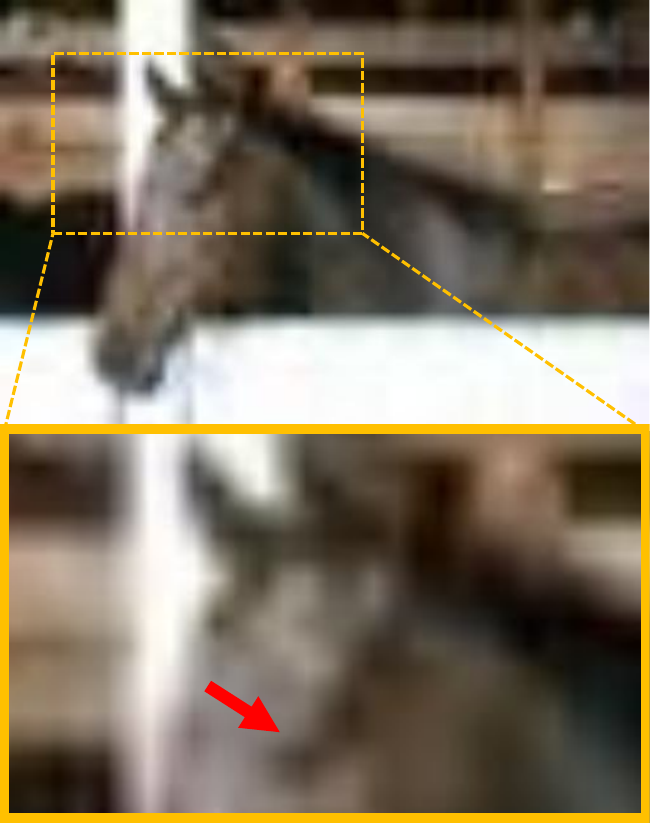}}
        \hfill
        \subfloat[SwinIR~\cite{sr_swinir}]{\includegraphics[width=\wp\linewidth]{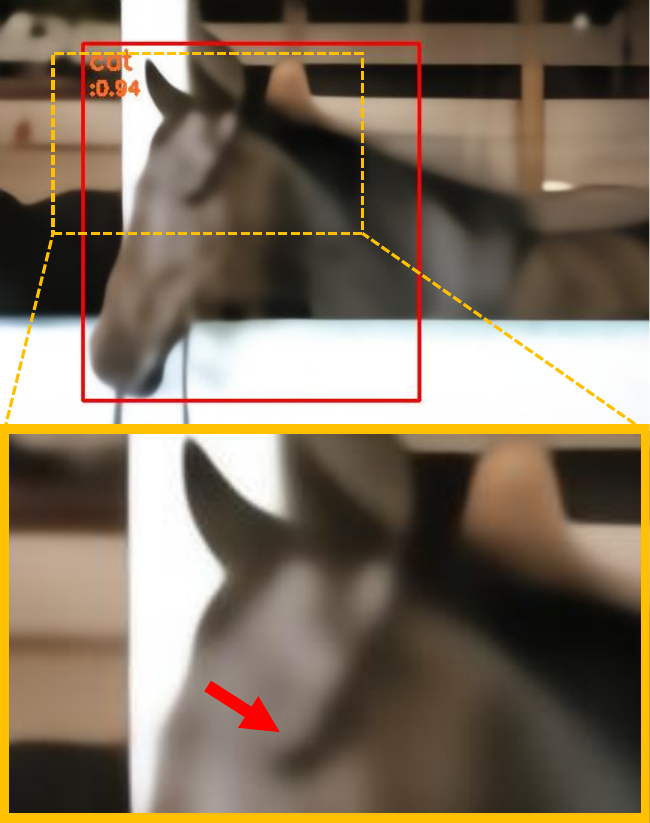}}
        \hfill
        \subfloat[TDSR~\cite{sr_tdsr}]{\includegraphics[width=\wp\linewidth]{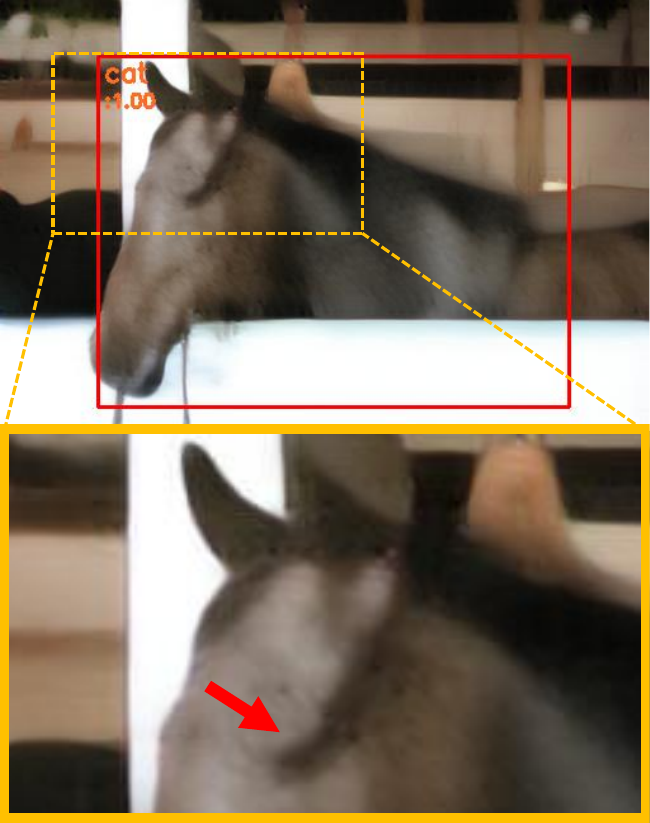}}
        \hfill
        \subfloat[RSRSSN~\cite{zhao2018residual}]{\includegraphics[width=\wp\linewidth]{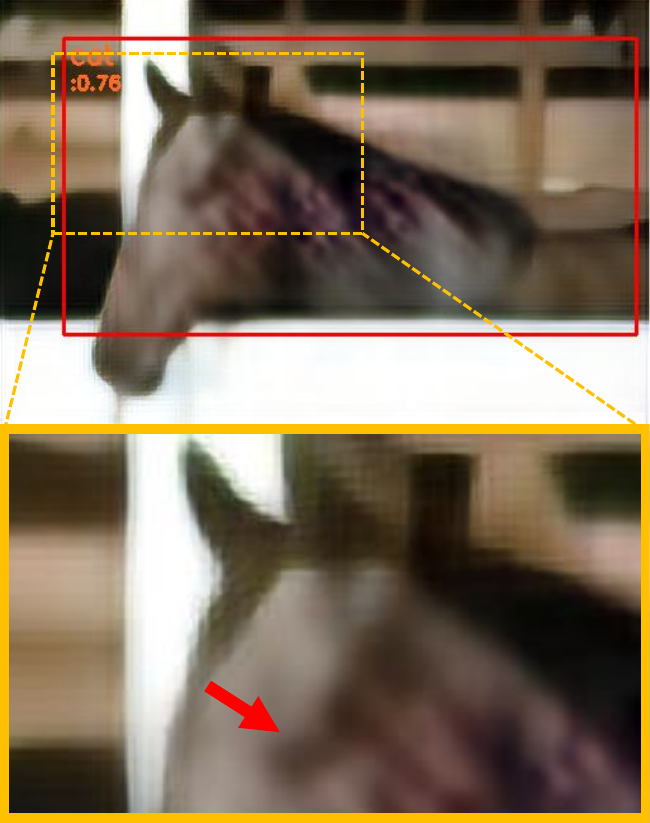}}
        \vspace{1mm}
        \\
        \subfloat[SR4IR~\cite{kim2024beyond}]{\includegraphics[width=\wp\linewidth]{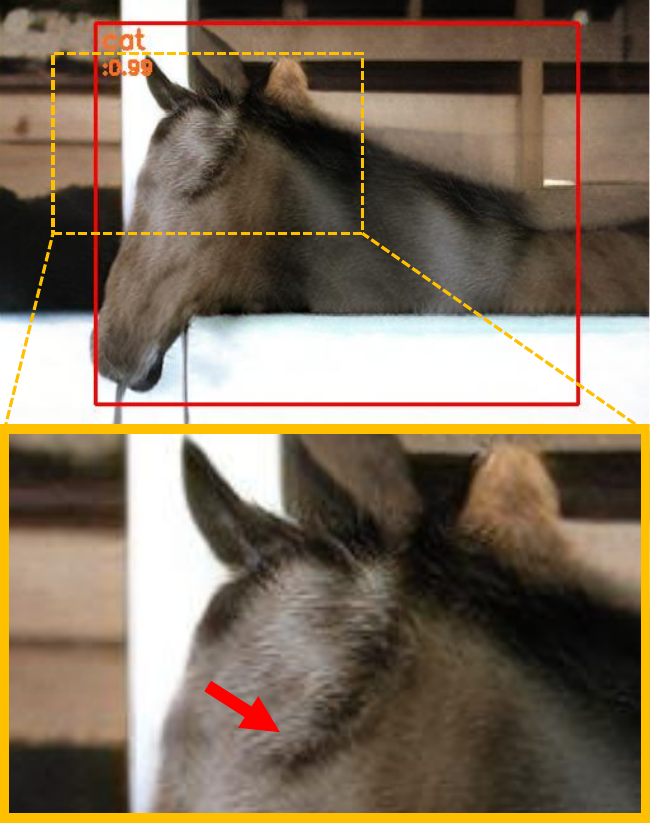}}
        \hfill
        \subfloat[\textbf{EDTR~(Ours)}]{\includegraphics[width=\wp\linewidth]{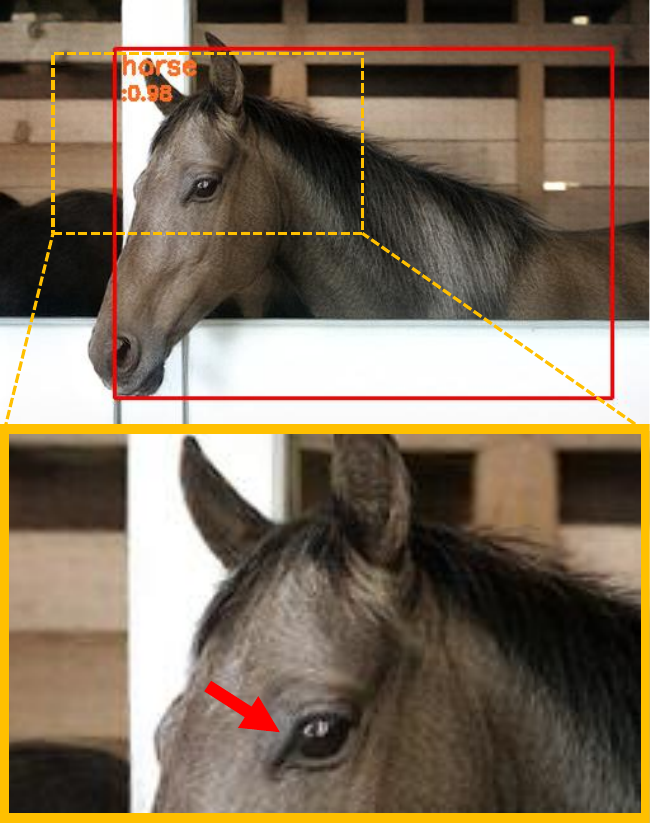}}
        \hfill
        \subfloat[HQ~(Oracle)]{\includegraphics[width=\wp\linewidth]{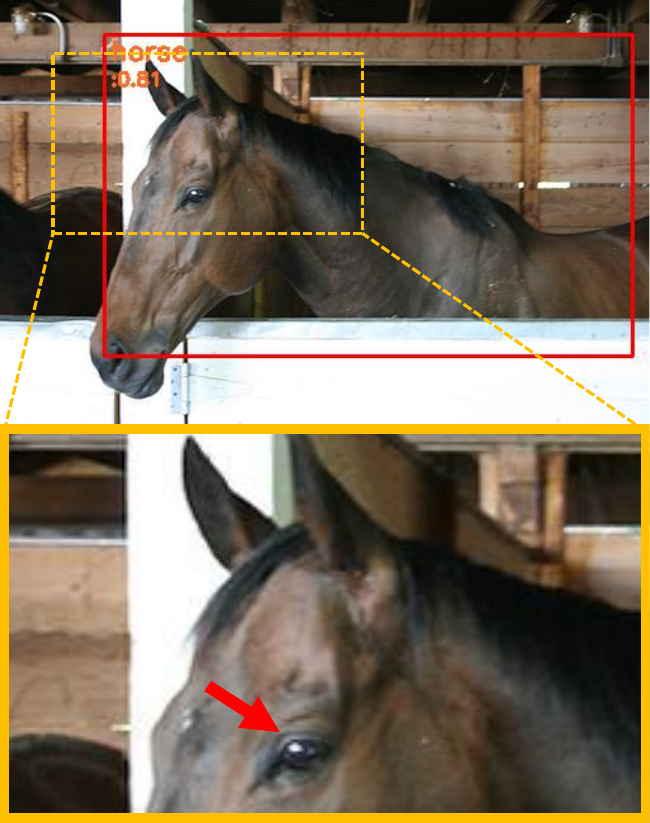}}
        \hfill
        \subfloat[HQ~(Ground-truth)]{\includegraphics[width=\wp\linewidth]{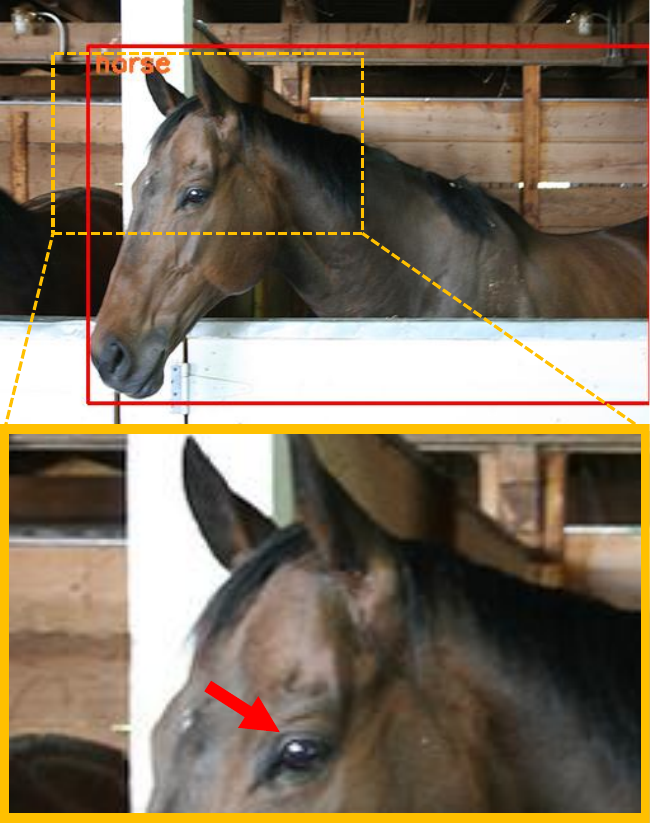}}
        \addtocounter{subfigure}{-8}
        \\
        \vspace{1mm}
        \begin{tikzpicture}
            \draw[dashed] (0,0) -- (17.2,0);
        \end{tikzpicture}
        \vspace{2mm}
        \\
        \subfloat[LQ~(\textit{No} restoration)]{\includegraphics[width=\wp\linewidth]{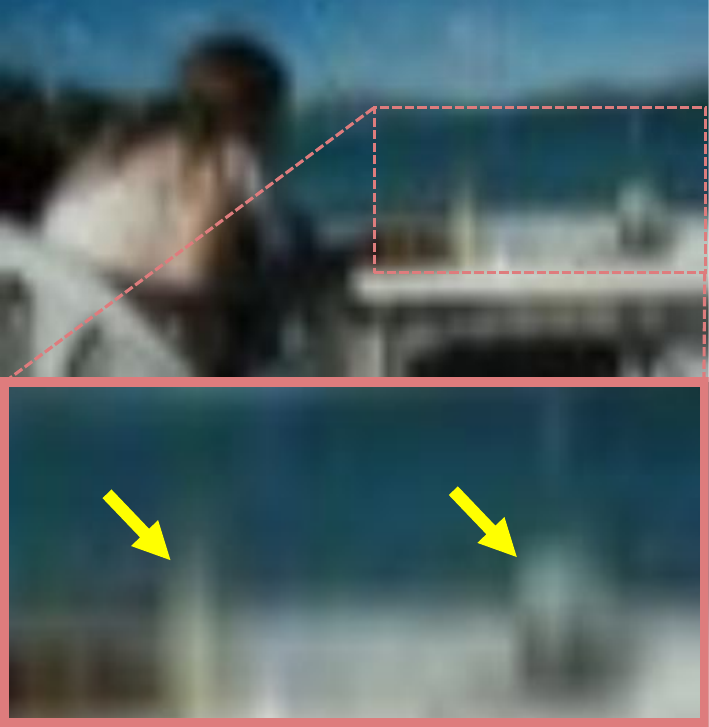}}
        \hfill
        \subfloat[SwinIR~\cite{sr_swinir}]{\includegraphics[width=\wp\linewidth]{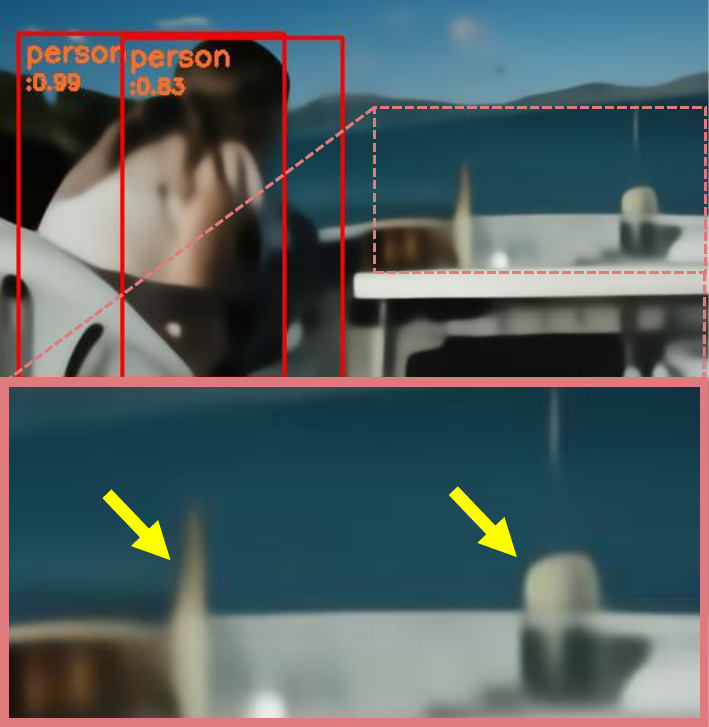}}
        \hfill
        \subfloat[TDSR~\cite{sr_tdsr}]{\includegraphics[width=\wp\linewidth]{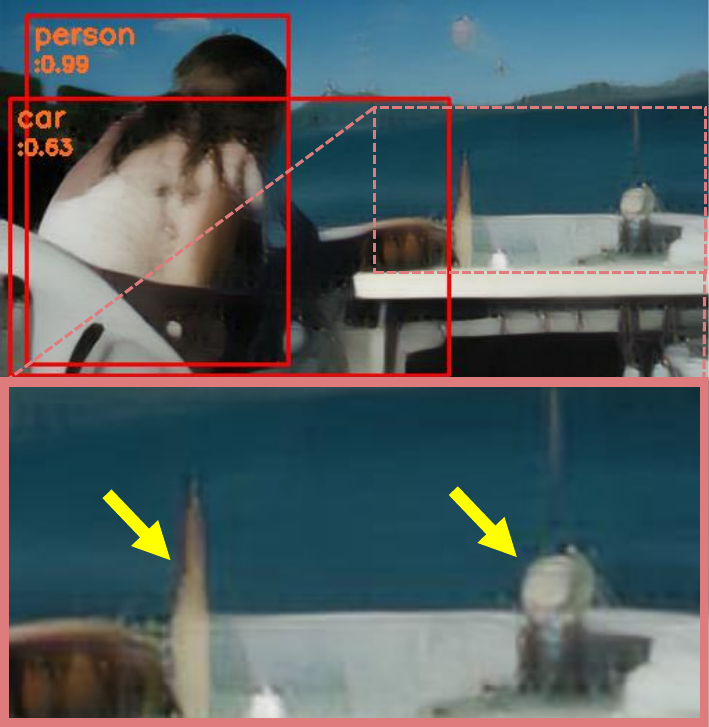}}
        \hfill
        \subfloat[RSRSSN~\cite{zhao2018residual}]{\includegraphics[width=\wp\linewidth]{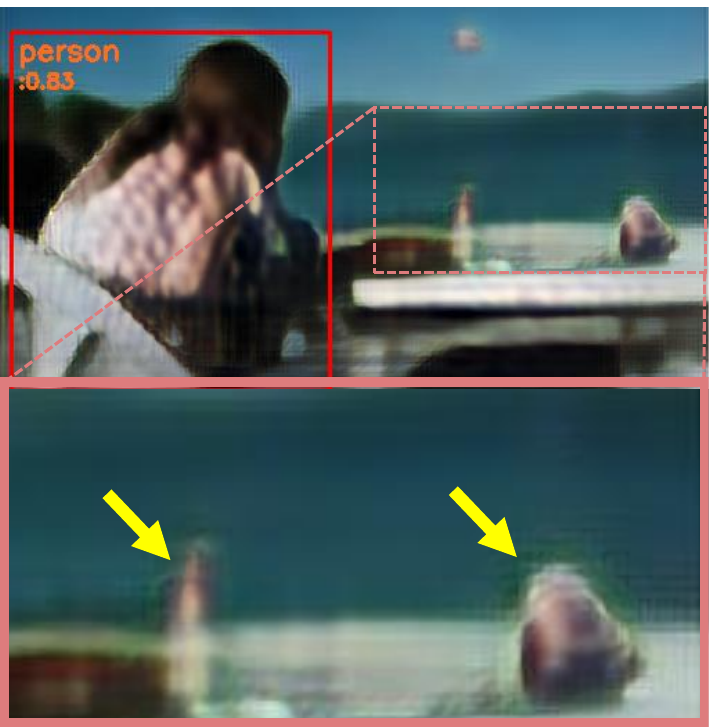}}
        \vspace{1mm}
        \\
        \subfloat[SR4IR~\cite{kim2024beyond}]{\includegraphics[width=\wp\linewidth]{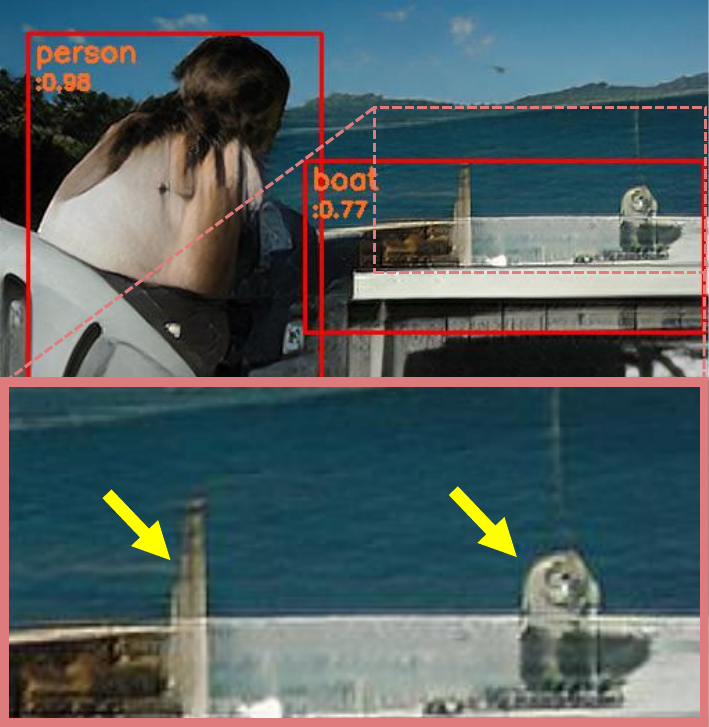}}
        \hfill
        \subfloat[\textbf{EDTR~(Ours)}]{\includegraphics[width=\wp\linewidth]{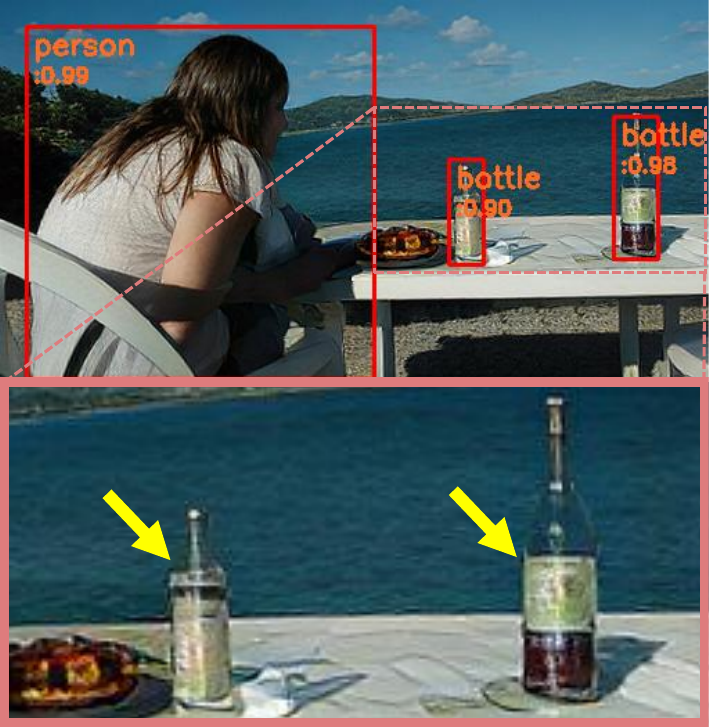}}
        \hfill
        \subfloat[HQ~(Oracle)]{\includegraphics[width=\wp\linewidth]{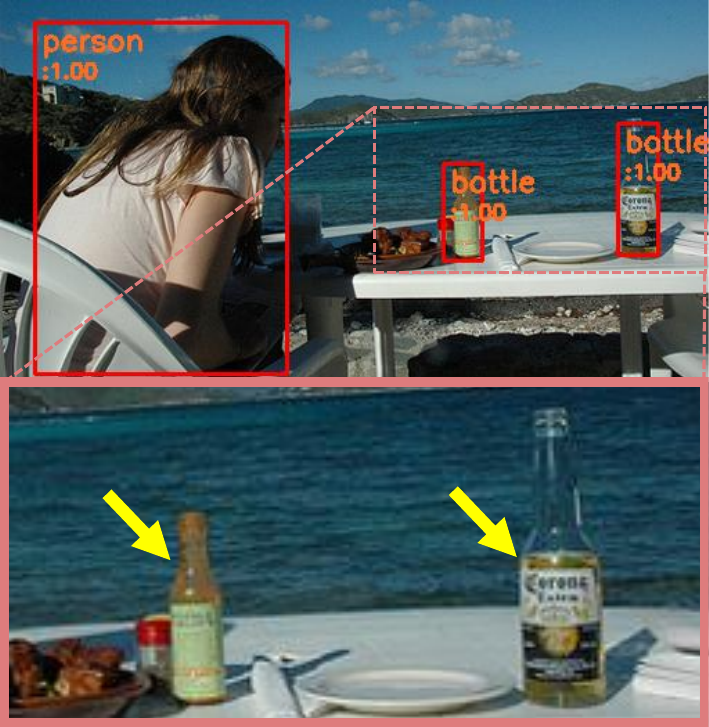}}
        \hfill
        \subfloat[HQ~(Ground-truth)]{\includegraphics[width=\wp\linewidth]{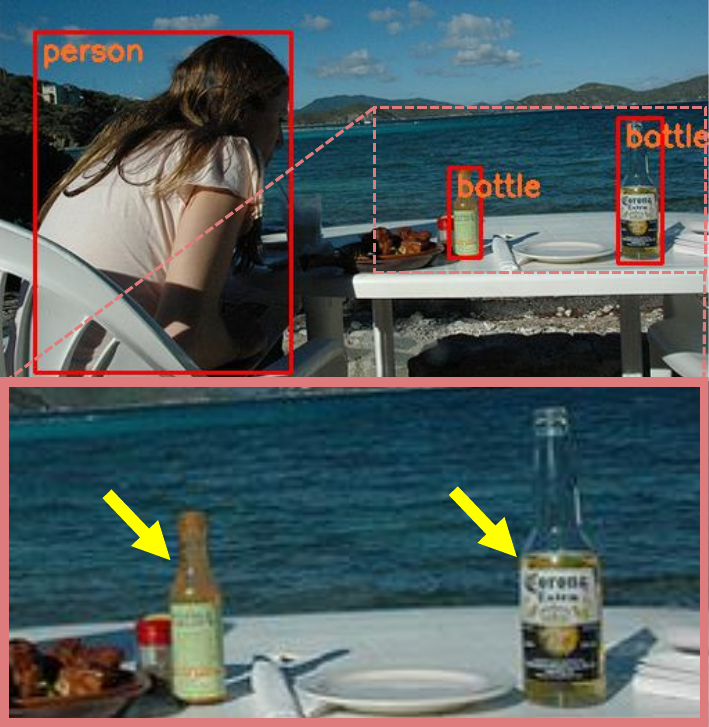}}
    \vspace{-0.1cm}
    \caption{\textbf{Further visualization of images and object detection results on degraded LQ (Mixture-\textit{B}) images.}
        We show the restored images and the corresponding predicted or ground-truth labels.
        The EDTR-4\,step model is used for visualization.
    }
    \label{fig:further-vis-det}
\end{figure*}

% 
% \vspace{-2mm}
\paragraph{Benchmark datasets.}
Figures~\ref{fig:further-vis-cls}, \ref{fig:further-vis-seg}, and \ref{fig:further-vis-det} provide additional visualizations across various high-level vision tasks.
We include further comparisons with SwinIR~\cite{sr_swinir}, TDSR~\cite{sr_tdsr}, RSRSSN~\cite{zhao2018residual}, and ground-truths, expanding on Figure~\ref{fig:qualitative-results} from our main manuscript.
Figure~\ref{fig:further-vis-cls} shows that EDTR successfully restores fine details of the bird, resulting in the only correct classification.
Figure~\ref{fig:further-vis-seg} demonstrates that EDTR restores the horse's mane and the bottle's shape, producing the most accurate segmentation results.
Figure~\ref{fig:further-vis-det} illustrates that EDTR restores the horse's eye and the bottle's shape, achieving the only correct detection.
These results clearly demonstrate that EDTR, which effectively leverages the powerful diffusion prior, is highly valuable for addressing the TDIR problem in challenging degraded scenarios.

\begin{figure}[h!]
    \centering
    \subfloat{\includegraphics[width=0.48\linewidth]{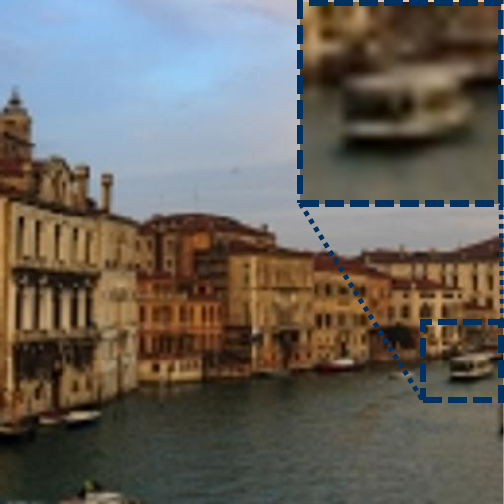}}
    \hfill
    \subfloat{\includegraphics[width=0.48\linewidth]{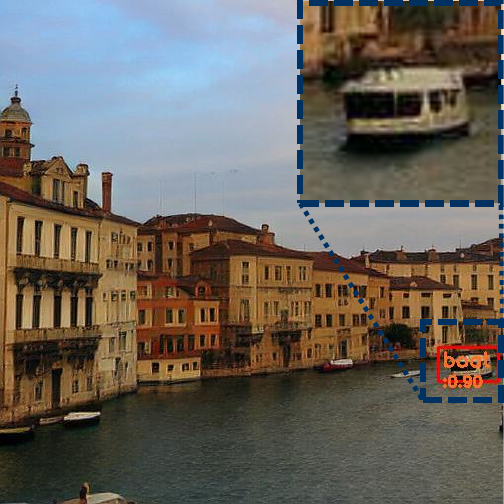}}
    \addtocounter{subfigure}{-2}
    \\
    \vspace{1mm}
    \subfloat{\includegraphics[width=0.48\linewidth]{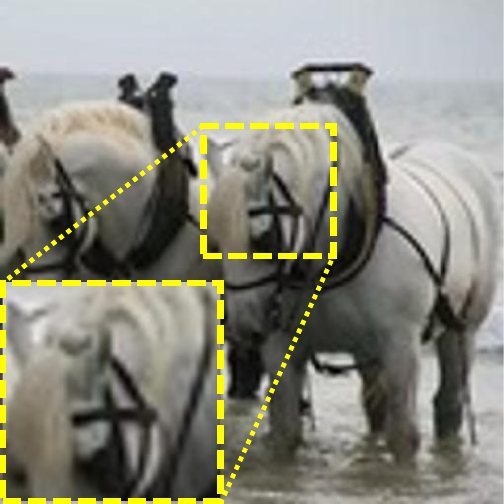}}
    \hfill
    \subfloat{\includegraphics[width=0.48\linewidth]{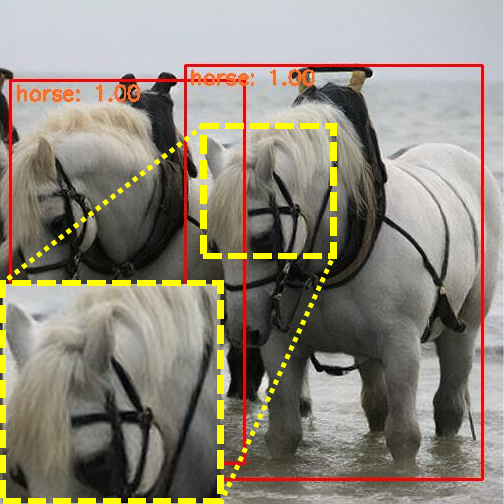}}
    \addtocounter{subfigure}{-2}
    \\
    \vspace{1mm}
    \subfloat[LQ~(\textit{No} restoration)\label{fig:real-world-no-restoration}]{\includegraphics[width=0.48\linewidth]{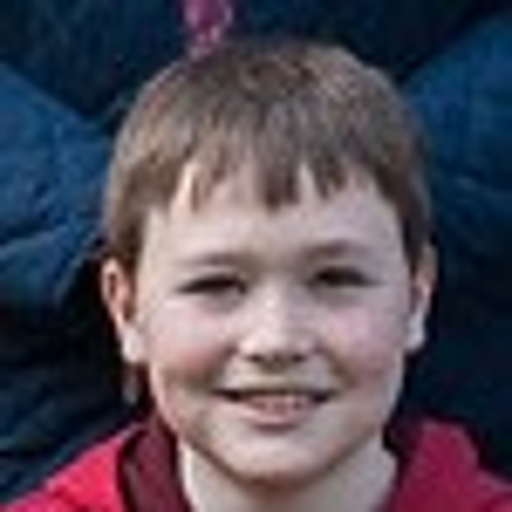}}
    \hfill
    \subfloat[\textbf{EDTR (Ours)}\label{fig:real-world-no-restoration}]{\includegraphics[width=0.48\linewidth]{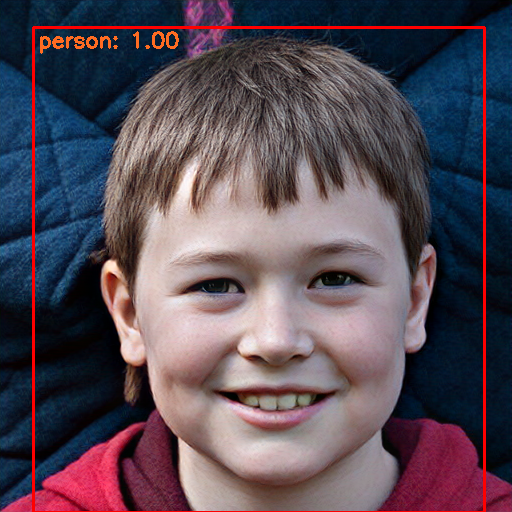}}
    \\
    \figspace
    % \vspace{-1mm}
    \caption{
        \textbf{Additional real-world object detection results and visualization of the restored image.}
        The first, second, and third images are from $\texttt{"14.png"}$, $\texttt{"01.png"}$ and $\texttt{"21.png"}$ in the real-world image set RealPhoto60~\cite{yu2024scaling}.
        The EDTR-4\,step model is used for visualization.
    }
    \label{fig:additional-real-world}
    \figxspace
    \vspace{-2mm}
\end{figure}

\vspace{-2mm}
\paragraph{Real-world images.}
Figures~\ref{fig:additional-real-world} provide additional visualizations of real-world object detection.
Our EDTR successfully restores the shape of the boat, the horse's face, and the boy's face using the powerful diffusion prior, enabling successful detection in real-world degraded images.
These results demonstrate the strong generalizability and practicality of our method.

\FloatBarrier
%%%%% CLASSIFICATION %%%%%

%%%%% SEGMENTATION %%%%%

%%%%% DETECTION %%%%%
\FloatBarrier